\documentclass{article}
\usepackage{arxiv-one_col}

\usepackage[utf8]{inputenc} 
\usepackage[T1]{fontenc}    
\usepackage{hyperref}       
\usepackage{url}            
\usepackage{amsfonts}       
\usepackage{nicefrac}       
\usepackage{graphicx}
\usepackage{multirow}
\usepackage{tabularx}
\usepackage{ragged2e}
\usepackage{enumitem}
\usepackage{booktabs} 
\usepackage{threeparttable} 
\usepackage{xtab}
\usepackage{longtable}
\usepackage{quoting}
\usepackage{subcaption}
\usepackage{array}
\usepackage{xcolor}
\usepackage[dvipsnames]{xcolor} 
\usepackage{ragged2e}
\usepackage{amsmath} 
\usepackage{colortbl} 
\usepackage{orcidlink}
\usepackage{multicol}  
\usepackage{graphicx}  
\usepackage{subcaption} 

\newcommand{\lightrule}{\arrayrulecolor{gray!60}\midrule[0.1pt]\arrayrulecolor{black}}
\usepackage[numbers]{natbib}
\hypersetup{
    colorlinks=true,
    linkcolor=blue,
    citecolor=black,      
    urlcolor=blue,
}

\definecolor{myblue}{RGB}{0, 102, 204}
\definecolor{myorange}{RGB}{204, 102, 0}
\definecolor{mygreen}{RGB}{0, 153, 0}
\definecolor{mypurple}{RGB}{102, 0, 102}

\title{Data-Efficient Adaptation and a Novel Evaluation Method for Aspect-based Sentiment Analysis

}

\author{
  Yan Cathy Hua \orcidlink{0000-0001-9155-9667} \qquad
  Paul Denny \orcidlink{0000-0002-5150-9806} \qquad
  J{\"o}rg Wicker \orcidlink{0000-0003-0533-3368}  \qquad
  Katerina Ta\v{s}kova \orcidlink{0000-0002-3217-7877} \\
  \\
  School of Computer Science, University of Auckland, New Zealand
}

\date{}

\renewcommand{\headeright}{}
\renewcommand{\undertitle}{}
\renewcommand{\shorttitle}{Data-Efficient Adaptation and a Novel Evaluation Method for Aspect-based Sentiment Analysis}

\hypersetup{
pdftitle={Data-Efficient Adaptation and a Novel Evaluation Method for Aspect-based Sentiment Analysis},
pdfauthor={Yan (Cathy) ~Hua, Paul ~Denny, J{\"o}rg ~Wicker, Katerina ~Ta\v{s}kova},
pdfkeywords={ABSA, opinion mining, evaluation, resource-efficient, SLM, model merging},
}

\begin{document}
\setlength{\leftmargini}{1em} 
\maketitle

\begin{abstract} 

Aspect-based Sentiment Analysis (ABSA) is a fine-grained opinion mining approach that identifies and classifies opinions associated with specific entities (aspects) or their categories within a sentence. Despite its rapid growth and broad potential, ABSA research and resources remain concentrated in commercial domains, leaving analytical needs unmet in high-demand yet low-resource areas such as education and healthcare. Domain adaptation challenges and most existing methods' reliance on resource-intensive in-training knowledge injection further hinder progress in these areas. Moreover, traditional evaluation methods based on exact matches are overly rigid for ABSA tasks, penalising any boundary variations which may misrepresent the performance of generative models. This work addresses these gaps through three contributions: 1) We propose a novel evaluation method, Flexible Text Similarity Matching and Optimal Bipartite Pairing (FTS-OBP), which accommodates realistic extraction boundary variations while maintaining strong correlation with traditional metrics and offering fine-grained diagnostics. 2) We present the first ABSA study of small decoder-only generative language models (SLMs; \textless7B parameters), examining resource lower bounds via a case study in education review ABSA. We systematically explore data-free (in-context learning and weight merging) and data-light fine-tuning methods, and propose a multitask fine-tuning strategy that significantly enhances SLM performance, enabling 1.5--3.8~B models to surpass proprietary large models and approach benchmark results with only 200–1000 examples on a single GPU. 3) We release the first public set of education review ABSA resources to support future research in low-resource domains.

\end{abstract}



\section{Introduction}\label{sec1_intro}

Opinionated text through which people express views or attitudes, such as user reviews, comments, and open-ended survey responses, is an important source of insights that influence individual and organisational decisions \citep{liu2012sentimentanalysisbook, batch2_survey_absa}. \textbf{Aspect-based Sentiment Analysis (ABSA)} is a fine-grained opinion mining (i.e. sentiment analysis; \cite{liu2012sentimentanalysisbook}) approach that has seen rapid growth in the past decade \citep{batch2_survey_absa, hua2024absa}. Given a piece of opinionated text, ABSA takes use-case-defined target entities (``\textbf{aspect}'') or their categories (``\textbf{category}''), identifies all opinion expressions (``\textbf{opinion}'') associated with them within each sentence, and classifies each opinion's sentiment polarity (``\textbf{sentiment}'') \citep{hua2024absa, batch2_survey_absa}. ABSA tasks differ in complexity and output richness by the components included, which can be any combination of the associated aspect/opinion expressions and aspect-category/sentiment labels \citep{hua2024absa, batch2_survey_absa} as shown in Table~\ref{table_absa_example}. Compared with document- and sentence-level opinion mining, ABSA has distinct advantages in use cases with complex text data containing mixed opinions on multiple entities, and where the target focus is either latent concepts or entities with highly variable forms \citep{liu2012sentimentanalysisbook, batch2_survey_absa, ref_2025_01}. It has been used in a wide range of domains to distil insight and inform decisions, from products and services, public policy, to health care and education \citep{hua2024absa}. 



\renewcommand{\arraystretch}{1.2}

\begin{table*}[!tbp]
\footnotesize
\caption{Example of ABSA components and subtask outputs}
\label{table_absa_example}

\begin{tabularx}{\textwidth}{
>{\raggedright\arraybackslash}p{0.25\linewidth} 
>{\raggedright\arraybackslash}p{0.125\linewidth} 
>{\raggedright\arraybackslash}p{0.125\linewidth} 
>{\raggedright\arraybackslash}p{0.125\linewidth}  
>{\raggedright\arraybackslash}p{0.37\linewidth}
}
\toprule

\textbf{Example text} & \multicolumn{4}{l}{\textit{It’s loud but the pie is the best.}} \\ \midrule
\textbf{ABSA Component} & \textbf{Aspect ($a$)} & \textbf{Opinion ($o$)} & \textbf{Category ($c$)} & \textbf{Sentiment ($s$)} \\

Relation unit 1 & (implicit) \textsuperscript{\textcolor{myblue}{1}} & \textcolor{myorange}{loud} & \textcolor{mygreen}{ambient} & \textcolor{mypurple}{negative (neg)} \\
Relation unit 2 & \textcolor{myblue}{pie} & \textcolor{myorange}{the best} & \textcolor{mygreen}{food} & \textcolor{mypurple}{positive (pos)} \\
\midrule

\multicolumn{2}{l}{\textbf{ABSA Tasks}} & \multicolumn{1}{l}{\textbf{Output Unit}} & \multicolumn{2}{l}{\textbf{Entry-level Output}} \\
\lightrule

\multicolumn{2}{l}{Aspect Extraction (AE)} & \multicolumn{1}{l}{\textcolor{myblue}{$a_i$}} 
& \multicolumn{2}{l}{{[}\textcolor{myblue}{null}, \textcolor{myblue}{pie}{]}} \\

\multicolumn{2}{l}{Opinion Extraction (OE)} & \multicolumn{1}{l}{\textcolor{myorange}{$o_i$}} 
& \multicolumn{2}{l}{{[}\textcolor{myorange}{loud}, \textcolor{myorange}{the best}{]}} \\

\multicolumn{2}{l}{Aspect Category Detection (ACD)} & \multicolumn{1}{l}{(\textcolor{myblue}{$a_i$}, \textcolor{mygreen}{$c_i$})} 
& \multicolumn{2}{l}{{[}(\textcolor{myblue}{null}, \textcolor{mygreen}{ambient}), (\textcolor{myblue}{pie}, \textcolor{mygreen}{food}){]}} \\

\multicolumn{2}{l}{Aspect Sentiment Classification (ASC)} & \multicolumn{1}{l}{(\textcolor{myblue}{$a_i$}, \textcolor{mypurple}{$s_i$})} 
& \multicolumn{2}{l}{{[}(\textcolor{myblue}{null}, \textcolor{mypurple}{neg}), (\textcolor{myblue}{pie}, \textcolor{mypurple}{pos}){]}} \\

\multicolumn{2}{l}{Aspect-Opinion Pair Extraction (AOPE)} & \multicolumn{1}{l}{(\textcolor{myblue}{$a_i$}, \textcolor{myorange}{$o_i$})} 
& \multicolumn{2}{l}{{[}(\textcolor{myblue}{null}, \textcolor{myorange}{loud}), (\textcolor{myblue}{pie}, \textcolor{myorange}{the best}){]}} \\

\multicolumn{2}{l}{Aspect Sentiment Triplet Extraction (ASTE) \textsuperscript{\textcolor{myblue}{2}}} & \multicolumn{1}{l}{(\textcolor{myblue}{$a_i$}, \textcolor{myorange}{$o_i$}, \textcolor{mypurple}{$s_i$})} 
& \multicolumn{2}{l}{{[}(\textcolor{myblue}{null}, \textcolor{myorange}{loud}, \textcolor{mypurple}{neg}), (\textcolor{myblue}{pie}, \textcolor{myorange}{the best}, \textcolor{mypurple}{pos}){]}} \\

\multicolumn{2}{l}{Aspect-Sentiment Quadruplet Extraction (ASQE) \textsuperscript{\textcolor{myblue}{3}}} & \multicolumn{1}{l}{(\textcolor{myblue}{$a_i$}, \textcolor{myorange}{$o_i$}, \textcolor{mygreen}{$c_i$}, \textcolor{mypurple}{$s_i$})} 
& \multicolumn{2}{l}{{[}(\textcolor{myblue}{null}, \textcolor{myorange}{loud}, \textcolor{mygreen}{ambient}, \textcolor{mypurple}{neg}), (\textcolor{myblue}{pie}, \textcolor{myorange}{the best}, \textcolor{mygreen}{food}, \textcolor{mypurple}{pos}){]}} \\

\bottomrule
\end{tabularx}

\caption*{
\begin{minipage}{\textwidth}
\footnotesize
\setlength{\baselineskip}{1.2\baselineskip}
\hangindent=2.5em \hangafter=1
\textsuperscript{1} This is an implicit aspect that is absent from the text (marked `null') but its category and relations can be inferred from the context. \par
\hangindent=2.5em \hangafter=1
\textsuperscript{2} Despite the task name, each ASTE triplet contains an aspect term, an opinion term, and a sentiment label. \par
\hangindent=0.75em \hangafter=1
\textsuperscript{3} Despite the task name, each ASQE quadruplet contains an aspect term, an opinion term, a category label, and a sentiment label.
\end{minipage}
}

\end{table*}
\renewcommand{\arraystretch}{1}


ABSA relies on capturing relationships that often require additional general-language and domain-knowledge as well as local textual context \citep{batch2_survey_absadl, liu2012sentimentanalysisbook}. This makes domain adaptation a challenge and has attracted much research effort in incorporating domain knowledge and relevant context into model training \citep{hua2024absa, batch2_survey_absa, batch2_survey_absadl}. The resulting representation-learning modules in ABSA models require large domain-specific annotated datasets and/or other lexical resources, which became a primary hurdle for research development in application domains with few public datasets, linguistic resources, and domain-specific models (``low-resource domains'') \citep{hua2024absa}. 

The research and resource gaps across ABSA domains are substantial. A recent systematic review \citep{hua2024absa} of 519 studies (2008–2023) identified 14 application domains, yet 71\% relied on product and service review data, with only two widely used datasets originating elsewhere. This domain also accounted for the most domain-specific studies (N = 126), over ten times more than the next few largest domains: education (N = 12), public policy (N = 8), and healthcare (N = 7). The resulting lack of ABSA resources leaves key analytical needs unmet in non-commercial domains; even in education, large volumes of student feedback remain under-analysed \citep{hua2025edurabsa, edu_1_hmm, edu_5_autoscoring}. Bridging this gap requires methods that can maximise available resources with minimal requirements.

\subsection{Motivation} \label{subsection_motivation}

As detailed in Appendix~\ref{subsec_existing_approaches}, previous ABSA approaches required substantial effort and resources for representation learning, knowledge injection, and relationship modelling \citep{hua2024absa}. Many rely heavily on lexical resources and multi-module architectures that can form learning bottlenecks and poor domain-adaptability \citep{sk2_jointabsa, jointabsa, batch2_survey_absa, implicitOE_7}, while traditional sequence-labelling formulations struggle with implicit aspects and complex aspect–opinion relationships \citep{joint_aste, jointabsa}. We hypothesised that today's pre-trained decoder-only Generative Large Language Models (GLMs)\footnote{We use ``GLM'' to denote decoder-only Large Language Models (LLMs) such as GPTs \citep{refGPT3, gpt4} and the Llama family \citep{llama3}, distinguishing them from earlier encoder-only and encoder-decoder LLMs such as BERT \citep{bert} and T5 \citep{t5}.}, even the small ones with fewer than 7~billion (B) parameters (SLMs), already encode rich general language and semantic relationships through pre-training, reducing the need for complex architectures and extensive task-specific training. Furthermore, we believe that the instruction-following and in-context learning (ICL) capabilities of GLMs make them an ideal candidate for domain-adaptation and use-case recalibration. However, few studies to date have examined the effectiveness of GLMs on ABSA tasks \citep{hua2024absa, bai_llm_2024}, and those that did (e.g. \cite{bai_llm_2024, zhou_llm_2024, llm_czechabsa_2025}) focused only on product and service reviews, with most using the same benchmark dataset family released long before GLM pre-training, risking result contamination. Moreover, the ABSA potential of SLMs that can be hosted locally and fine-tuned on a single GPU for low-resource domains remains underexplored. To our knowledge, the technique of model weight merging, which can enhance model performance without additional training data \citep{mergekit, modelsoup}, has not yet been investigated for ABSA.

In addition, we identified a limitation with the existing ABSA evaluation method that could undermine model performance in multiple ABSA tasks. Most past studies have calculated evaluation metrics using exact matching (e.g., \cite{zhang2021_asqp, cai2021_acos, ref_2025_t5absa, bai_llm_2024}). For example, in Table~\ref{table_absa_example}, each of the ASQE output units is considered a match with ground-truth if and only if all four components in the quadruplet are identical, thus ``(pie, \textbf{the} best, food, pos)'' would be a non-match with ``(pie, best, food, pos)''. We argue that exact matching is not suitable for ABSA tasks with aspect/opinion extraction components (e.g. OE, AOPE, ASTE, and AOPE), as many extraction boundary differences are trivial, subjective, and could stem from different annotation rules. E.g., ``best'' vs. ``the best'' should not invalidate the entire output unit. Instead, a suitable matching criterion should allow realistic extraction boundary variations based on subsequence similarity, analogous to the ROUGE and BLEU metrics widely adopted in text summarisation and translation \citep{rouge, rouge_bleu}. As multi-component tasks such as ASTE and ASQE attract more research and application attention \citep{hua2024absa}, it is crucial to equip the field with an appropriate evaluation method. For domains such as educational review, where aspect and opinion expressions are typically longer and more complex than in product or service reviews \citep{hua2024absa}, an appropriate evaluation criterion could make ABSA solutions more attainable and practically useful.

\subsection{Research Questions} \label{subsection_rqs}

To fill these gaps, we introduce a novel evaluation method tailored for ABSA tasks, systematically examine the capability of pre-trained GLMs and particularly SLMs in ABSA tasks, and explore efficient data-free and data-light approaches to enhance their performance. We use education review (i.e. student reviews of courses, teaching, institutions, and experience) ABSA as a case study for high-demand, low-resource domains \citep{hua2025edurabsa}, and employ a new challenging dataset whose annotation was unseen by the GLMs to ensure result reliability and robustness.  We seek to answer the following Research Questions (RQs):

\vspace{-0.5em}
\begin{quote} 
\setlength{\parskip}{0.75em} 

    \textbf{RQ1.} How well do pre-trained GLMs and SLMs perform across different ABSA tasks, and how are their performances influenced by model size and ICL?  
    
    \textbf{RQ2.} How effective is multitask fine-tuning for SLMs (below 7~B parameters) on ABSA tasks, and what are the minimal data and fine-tuning requirements to achieve noticeable performance gains?  
    
    Further, to maximise the utility and performance of fine-tuned models, we investigate weight-merging as a data-free strategy in RQ3 below.
    
    \textbf{RQ3.} Can weight merging further enhance the performance of fine-tuned SLMs?  

\setlength{\parskip}{1em} 
\end{quote}
\vspace{0.5em}

\subsection{Contributions} \label{contributions}

Our work makes the following contributions:

\begin{itemize}
    
    \item We propose \textbf{Flexible Text Similarity Matching and Optimal Bipartite Pairing (FTS-OBP)}, a novel evaluation method applicable to all ABSA tasks. FTS-OBP is tailored to ABSA tasks in which text extraction and classification are integrated within a single output unit, thereby overcoming the rigidity and coarseness of traditional exact-match-based methods. It provides more realistic and fine-grained performance insights across both components and tasks. Our analyses show that FTS-OBP correlates strongly with traditional metrics whilst providing controlled flexibility for the text extraction components and offering valuable component-specific diagnostic information.
    
    \item To our knowledge, this is the first ABSA study to 1) provide empirical evidence of the performance and resource efficiency of GLMs and SLMs on non-commercial domains, 2) systematically explore the minimal and effective resource requirements for domain adaptation with GLMs and SLMs, and 3) evaluate the effectiveness of model weight merging in ABSA. Our results demonstrate that fine-tuning SLMs is a promising solution for low-resource domains: with only 200 training examples and a rank-4 LoRA adapter, fine-tuned SLMs outperformed much larger GLMs, whilst with 1000 examples, even a 1.5~B-parameter SLM surpassed all tested GLMs, including GPT-4o. Model weight merging further enhanced the performance of fine-tuned SLMs through a fast, data-free process.
    
    \item We fill multiple gaps in the ABSA literature by creating the first set of public education review ABSA resources to support future research and applications, including: 1) multitask models in CUDA and hardware-agnostic ONNX formats that can handle challenging tasks such ASTE, ASQE, and implicit aspect/opinion extraction; 2) implementation details for our approaches and the FTS-OBP method; and 3) results on the first public domain-specific dataset. We share these resources at \url{https://github.com/yhua219/ftsobp_and_edurabsa_slm}.

\end{itemize}


\section{Methods}\label{sec3_methods}

\subsection{ABSA Task Scope}

We propose a multitask solution that can handle all common ABSA subtasks, including OE, ACD (via the ``AOC'' task detailed below), AOPE, ASTE, and ASQE, and can identify implicit aspects and extract implicit opinions. 

Following the SemEval 2015 and 2016 protocol \citep{semeval2015, semeval2016}, we define an aspect as an opinion target rather than any domain entity or attribute to avoid extracting non-opinionated items and better handle implicit aspects. Consequently, we replaced tasks without the opinion component with the opinion-inclusive equivalents: AOPE replacing AE, ASTE replacing ASC, and ACD extended into Aspect-Opinion Categorisation (AOC), where category labels are derived from AOPE output.

\subsection{Problem Formulation} \label{problem_formulation}

We formulate each ABSA task as an instruction-based text generation problem. Using ASQE as an example: Given an array of review text entries $T = \left[ t_1, \ldots, t_n \right]$, sets of category labels $C$ and sentiment labels $S$, and an instruction prompt text $P$, the ASQE task objective is generating the output array $Y = \left[ y_1, \ldots, y_n \right]$ such that for any review text entry $t_i$, the corresponding output $y_i$ is an array containing all $k \geq 0$ units (quadruplets) in $t_i$: 
            $y_i = \left[ (a_1, o_1, c_1, s_1), \ldots, (a_k, o_k, c_k, s_k) \right]$, 
where $a_j$, $o_j$, $c_j$, $s_j$ are the ABSA components: aspect term, associated opinion term, category label ($c_j \in C$), and sentiment label ($s_j \in S$) in the $j^\text{th}$ quadruplet for $t_i$. In particular, for any $j \in [1, k]$: 

\vspace{-1em}
\begin{itemize}
    \setlength{\itemsep}{0pt}
    \setlength{\parskip}{6pt}
    \setlength{\parsep}{0pt}
    \setlength{\parsep}{0pt}
    \setlength{\topsep}{0pt}
    
    \item For implicit aspects, $a_j$ = \text{`}$null$\text{'}.  
    \item $o_j$ includes both explicit and implicit opinion terms. 
    \item $a_j$ (explicit), $o_j$ must be consecutive substrings in $t_i$.
    \item For multi-level category labels such as those in the EduRABSA dataset, $c_j$ = $\text{`} main_j\text { - } sub_j\text{'}$ where $main$ and $sub$ are the labels for the main category (entity) and sub-category (attribute of the entity). 
    \item $s_j \in S = \{positive, neutral, negative\}$.    
\end{itemize}

As ASQE covers all ABSA components and subsumes the other tasks, the prompt instruction $P$ and the output structure of $y_i \in Y$ shown above can be simplified accordingly for other ABSA tasks. Table~\ref{table_absa_example} shows the output structure for OE, AOPE, ASTE, and ASQE; and we define AOC as: $y_i = \left[ (a_1, o_1, c_1), \ldots, (a_k, o_k, c_k) \right]$.

\subsection{Proposed Approaches} \label{subsection_proposed_approaches}

To answer the research questions, we explored three resource-efficient approaches with pre-trained GLMs and SLMs: 1) ICL with 0-shot and few-shot prompts, 2) fine-tuning SLMs with multi-task learning, and 3) model weight merging.

\textbf{Approach 1 - In-context Learning (ICL)}

For ICL, we evaluated pre-trained GLMs and SLMs in chat completion mode with two variables: 1) model size and 2) ICL type (0-shot vs. 4-shot). Each ICL input string included: the ABSA task name, the task instruction (``prompt''), and the review text. In 4-shot settings, the input also contained four example review–output pairs. Appendix~\ref{appendix_b1_icl_input_example} presents the 4-shot ASQE ICL input template, which generalises to 0-shot and the other four tasks, all of which are used in approaches~2 and~3.

\textbf{Approach 2 - SLM Supervised Fine-tuning (SFT)}

Supervised fine-tuning (SFT) is a common method for adapting pre-trained models to specific tasks or domains \citep{hu2022lora, llm_sft}. To target the low-resource constraints, including hardware limitations, we only focused on SLMs. As ABSA builds on general natural language understanding that SLMs are already equipped with, we adopt Low-Rank Adaptation (LoRA) \citep{hu2022lora} as the SFT approach to exploit these pre-trained capabilities while minimising parameter updates compared to full fine-tuning. We explored these key LoRA SFT variables: 1) training dataset size, 2) model size, 3) task-instruction ICL type (0-shot vs. 4-shot), and 4) LoRA rank.

\textbf{Approach 3 - Model Merging}

Another low-resource task/domain-adaptation approach is direct model weight merging, which combines the weights of multiple pre-trained or fine-tuned models with identical architectures to form a new model \citep{mergekit, modelsoup}. This technique can enhance performance and robustness while enabling reuse of the same base models across different merging strategies \citep{mergekit, modelsoup}.

We chose the SLERP merge method\footnote{Our implementation used \texttt{Arcee's MergeKit} \citep{mergekit} library} for its simple implementation, memory efficiency, and superior performance \citep{mergekit}. For any two weight vectors $w_1,w_2$ from different source models, given an interpolation parameter $t\in[0,1]$ that controls the blend ratio, the implementation uses linear interpolation (LERP) if the vectors are nearly collinear after normalisation, and otherwise computes the merged weight vector with SLERP: 

$\displaystyle \mathrm{merge}(w_1,w_2;t) = \frac{\sin((1-t)\theta)}{\sin\theta}\;w_1 \;+\; \frac{\sin(t\theta)}{\sin\theta}\;w_2$, \quad where $\theta$ is the angle between the normalised vectors.



\section{FTS-OBP: A Novel Evaluation Method}
\label{sec4_FTSOBP}



To address the coarseness and rigidity of exact-match-based metrics, we propose \textbf{FTS-OBP}, a novel evaluation method centred on \textbf{Flexible Text Similarity (FTS) Matching} and \textbf{Optimal Bipartite Pairing (OBP)}, which produces rich evaluation statistics both by output units and by ABSA components. 

\begin{figure*}[ht!]
  \centering
  \includegraphics[width=0.98\linewidth]{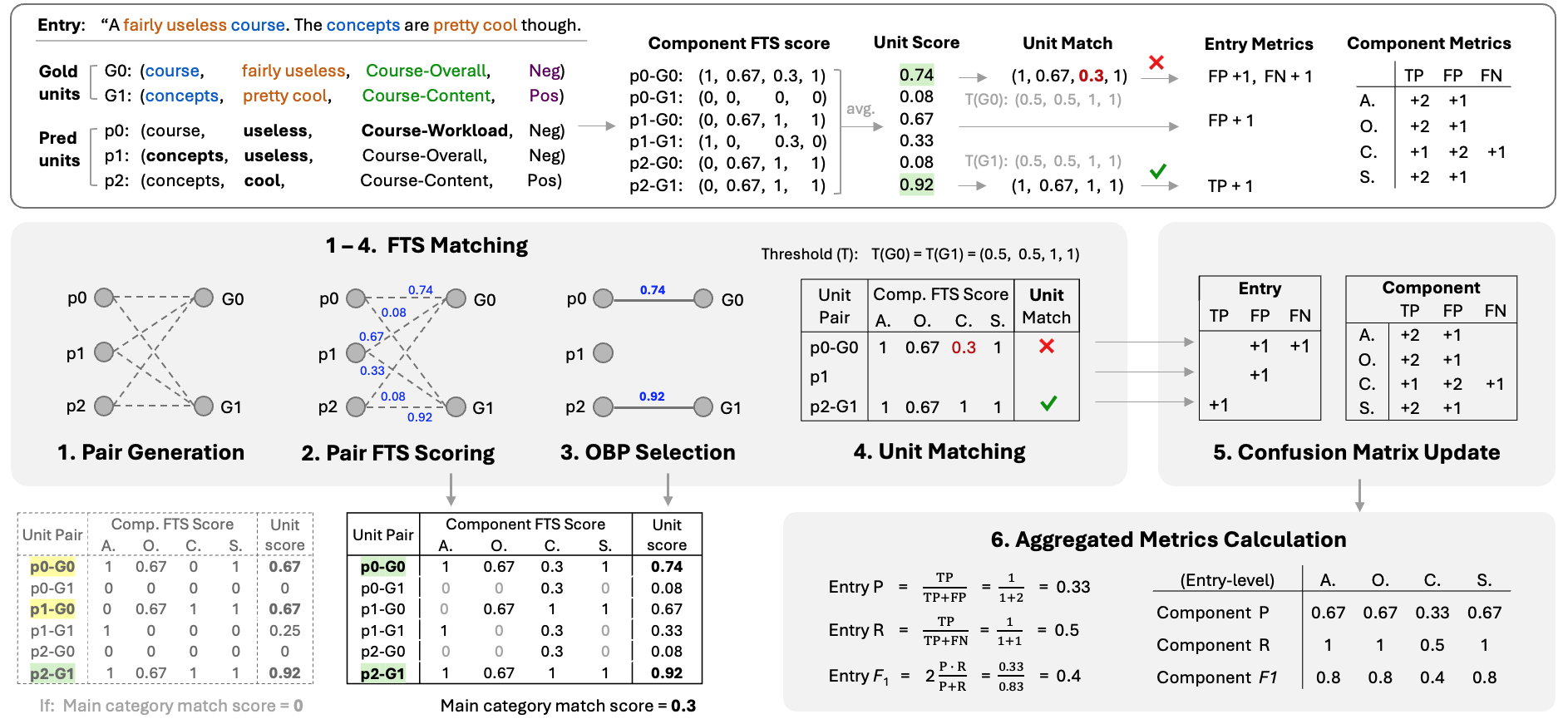}

  \caption{Example of the FTS-OBP evaluation method on the ASQE task, with one review entry and two ground-truth (gold) vs. three model output (pred) quadruplets (units). FTS matching uses exact matches for category (C) and sentiment (S) labels, and Rouge-L $F_1$ scores with a threshold for aspect (A) and opinion (O) extractions. The ``if'' table shows how main-category-match partial scores can assist OBP selection.}
  \label{fig_ftsobp}
\end{figure*}

\subsection{FTS Matching} \label{subsec_fts_matching}

The key difference between FTS-OBP and the traditional exact-match-based approach (``the traditional method'') is in their \textbf{component-level matching criteria}. While the traditional method applies a binary exact-match criterion across all components, FTS-OBP applies \textbf{FTS Matching} that treats the two types of ABSA components differently: 
\begin{itemize}
    \item For the text-extraction components (aspect and opinion), FTS-OBP compares a gold-pred pair's Flexible Text Similarity (FTS) score (detailed in subsection \ref{subsec_eval_method_fts}) against a threshold to accept similar-enough pairs as matches. 
    
    \item For the classification components (category and sentiment labels), FTS-OBP retains exact matching criteria with binary scoring and a threshold of 1. In addition, to provide richer information on multi-level category label matching, FTS-OBP allows an optional user-defined partial score (we chose 0.3 for a small weighting) for main-category label matches. This partial score replaces the binary non-match score 0 and contributes to unit pairing optimisation (introduced in Section~\ref{subsec_obp} below and shown in Figure~\ref{fig_ftsobp}), while exact full matching remains the default component-level matching criterion via the threshold of 1. 
\end{itemize}
Beyond this modification, FTS-OBP remains unchanged from the traditional method, such that a unit-pair match still requires all its components to match, and the unit-level confusion matrix and higher-level metric calculations remain identical to those of the traditional method.

\subsubsection{The Core of FTS Matching -- FTS Scoring with Threshold} \label{subsec_eval_method_fts}

The \textbf{Flexible Text Similarity (FTS) scoring} is the core of the FTS Matching, and is designed to evaluate the similarity of text-extraction component pairs with a score $\in [0, 1]$. It assigns 0 to pred text that is outside the input text (hallucination check) or has no overlap with the gold text, and otherwise returns the Rouge-L $F_1$-score \citep{rouge}. 

Rouge-L finds the longest common subsequence (LCS) between tokenised gold and pred texts, and computes similarity Precision ($P_{RL}$),  Recall ($R_{RL}$), and $F_1$ ($F_{1RL}$) scores from the length ratios between LCS and the source sequences:


\begin{itemize}[parsep=0pt, topsep=6pt]
\centering
\item[]
$\displaystyle \text{P}_{\text{RL}} = \frac{|\text{LCS}(\text{gold}, \text{pred})|}{|\text{pred}|}, \quad \text{R}_{\text{RL}} = \frac{|\text{LCS}(\text{gold}, \text{pred})|}{|\text{gold}|}, \quad F_{1\text{RL}} = \frac{2 \, \text{P}_{\text{RL}} \cdot \text{R}_{\text{RL}}}{\text{P}_{\text{RL}} + \text{R}_{\text{RL}}}$
\end{itemize}

The FTS can optionally take a list of stopwords and remove them from the pred and gold token sequences before computing similarity scores. Once a component-pair receives an FTS score, the match decision is made if the score exceeds a pre-defined \textbf{component score threshold} value $T$.

\subsection{Optimal Bipartite Pairing (OBP)} \label{subsec_obp}

Once non-perfect component matching is allowed, gold-pred unit pairing becomes more complex as multiple feasible pairings may exist. The second key feature of the FTS-OBP method, \textbf{Optimal Bipartite Pairing (OBP)}, addresses this challenge through optimal bipartite matching \citep{linearsum}. For each input entry with $n$ gold units and $p$ pred units, OBP constructs an $n \times p$ similarity matrix $D$, whose element $D_{ij}$ is the unit-similarity score for pair $(i,j)$ computed as the weighted sum of their component-FTS scores (we used equal weights across components). OBP then applies the linear sum assignment algorithm \citep{linearsum} to select the $\min(n,p)$ \textbf{optimal unit-pairs} that maximise the total similarity $\sum_{(i,j) \in \mathcal{M}} D_{ij}$, where $\mathcal{M}$ denotes the selected 1:1 matching\footnote{E.g., in Figure~\ref{fig_ftsobp}, $\mathcal{M}$ = [p0-G0, p2-G1]}, leaving $|n-p|$ units unmatched. The optional partial scores for main-category label matches contribute to $D$, enhancing pairing optimisation and diagnostic capabilities.

\subsection{Metric Calculation} \label{subsec_metric_calculation}

With the unit-pairing and component-matching done, each optimal unit-pair is evaluated for \textbf{unit-level matching}: if all its components pass their respective thresholds, it counts as a true positive (TP); otherwise, it contributes to both false positives (FP) and false negatives (FN). The $|n-p|$ unmatched units are counted as FP (unmatched preds) or FN (unmatched golds). Together, these counts form the \textbf{entry-level confusion matrix}, which is then aggregated across entries to compute task-level P, R, and $F_1$ scores similar to the classical method.

In addition to unit-level metrics, FTS-OBP also computes \textbf{per-component metrics} to provide additional angles of evaluation and diagnostics. Each entry has a separate confusion matrix for each component, and a component-pair counts as TP if it is a match, and otherwise contributes to both FP and FN. The $|n-p|$ unmatched units contribute FP or FN to all components. These per-component confusion matrices are aggregated across entries to compute task-level component-specific precision, recall, and $F_1$ scores. Figure~\ref{fig_component_metrics} illustrates the use of these metrics, which provide another dimension to diagnose model performance alongside the unit-match metrics.

\subsection{FTS-OBP Validation} \label{subsec_fts_validation}
We conducted simulation and empirical validation analyses on FTS-OBP, with full details provided in Appendix~\ref{Appendix_ftsobp}. Overall, FTS-OBP aligns well with the exact-match-based traditional evaluation method, while exhibiting the expected leniency towards minor boundary variations. The simulation results show that FTS-OBP, with our chosen threshold values\footnote{Introduced in Section~\ref{subsection_exp_metric}.}, allows more over-extraction (i.e., pred contains gold) than under-extraction, and strongly penalises boundary shifts (i.e., partial pred–gold overlap). Furthermore, comparison of experimental results evaluated using exact-match-based metrics and FTS-OBP across three datasets reveals a strong positive correlation both within and across tasks (overall Spearman’s $\rho$ = 0.784--0.934, Pearson’s $r$ = 0.760--0.956, all $p$~\textless~0.001), with FTS-OBP consistently scoring higher than the traditional method, particularly with pre-trained GLMs and SLMs. The metric differences diminish with increasing task complexity. Lastly, detailed result analyses presented in Appendices~\ref{appendix_E_detailed_results}~and~\ref{appendix_F_semeval_results} further demonstrate the expected behaviour of FTS-OBP, showing that over 62.04\%--98.84\% of accepted matches were identical across datasets, while fewer than 0.13\% of aspect pairs and 0.23\% of opinion pairs accepted contained boundary shifts.


\section{Experiments}\label{sec5_experiments}

We introduce the experimental setup and present the main results below, with full details provided in Appendices~\ref{appendix_C_experiment_details} and~\ref{appendix_E_detailed_results}--\ref{appendix_F_semeval_results}, respectively. 

\subsection{Experimental Setup}\label{subsec5_1_experiment_setup}

We conducted all experiments on a desktop machine with a single NVIDIA GeForce RTX 3090 GPU with 24~GB of graphics memory, powered by solar energy. All training and local inference were performed on the GPU.

    

The experiment involved six pre-trained models:   \quad  \textbf{Two SLMs} from Huggingface\footnote{\url{https://huggingface.co/}; from the latest checkpoints as of download time (February and July 2025)}: 1) Phi-4-mini-instruct (\cite{phi4mini}; 3.8~B parameters; hereafter, ``Phi4-mini'') and 2) Qwen-2.5-1.5B-Instruct (\cite{qwen25}; hereafter, ``Qwen2.5-1.5B''). These two SLMs served as the base model for LoRA fine-tuning.   \quad  \textbf{Two pairs of large and small GLMs} via API calls in August 2025:  1) GPT-4o, GPT-4o-mini\footnote{Via \href{https://openai.com/api/}{the OpenAI API}.}  \citep{gpt4};  and 2) Llama3-70B and Llama3-8B\footnote{Via \href{https://console.groq.com/docs/deprecations}{the Groq API} `meta-llama/llama3-70b-8192', `meta-llama/llama3-8b-8192' endpoints respectively.}  \citep{llama3}.

\subsubsection{Dataset} 
\label{subsubsec_dataset} 

We used three datasets in total: For the case study and the results introduced in the main text, we used the \textbf{EduRABSA} dataset \citep{hua2025edurabsa}, which contains 6,500 student reviews of courses, teaching staff, and universities, with manually extracted quadruplets for the ASQE task that can also be adapted for other ABSA tasks.  For comparison, we also replicated all three approaches and dataset treatments described below (under single-task settings) on two benchmark ASQE datasets: \textbf{ASQP Rest16} \citep{zhang2021_asqp} on restaurant reviews and \textbf{ACOS Laptop} \citep{cai2021_acos} on laptop reviews. 

For each dataset, we created five training sets with 200, 500, 1000, and 2000 examples per task\footnote{The ASQP Rest16 training set is limited to 1000 examples due to the original dataset's size constraint.}, each paired with an in-training validation set of 200 examples per task for early stopping. For post-training evaluation, all experiments shared a development set (200 examples per task) for hyper-parameter tuning and a test set (300 examples per task) for final evaluation. Each dataset has four variants: with 0-shot or 4-shot prompts, and in either multi-task (MT; all five tasks included) or single-task (ST) form. The MT variants have two further configurations: the ``cascade-order'' type (CC), where each input text yields five adjacent instances in the task order OE, AOPE, ASC, ASTE, and ASQE; and the ``task-order'' type (TT), where instances are grouped by task and correspond to the concatenation of the five single-task sets. Appendix \ref{appendix_exp_dataset} provides dataset statistics and processing details. 



\subsubsection{LoRA supervised fine-tuning (SFT)} 
\label{subsection_exp_sft}

For LoRA SFT of the Phi4-mini and Qwen2.5-1.5B pre-trained models, we used HuggingFace TRL's \texttt{SFTTrainer}\footnote{\url{https://huggingface.co/docs/trl/en/sft_trainer}} with bfloat16 precision, a per-device training and evaluation batch size of 1 (due to GPU memory constraints), and a cosine learning rate scheduler. To reduce gradient collapse and enable more stable learning with higher LoRA ranks, we applied the Rank-stabilised LoRA (rsLoRA) method \citep{rslora} to scale the LoRA adapters by $\alpha / \sqrt{r}$ instead of the conventional $\alpha / r$ \citep{rslora}, and set $\alpha = 2r$ for all training instances. For each training dataset size $T\in$ \{200, 500, 1000, 2000\}, we applied LoRA SFT with different LoRA ranks 4 $\leq R \leq$ 128 and adjusted the hyperparameters accordingly. Unless stated otherwise, all LoRA models were trained using multitask cascade task order data. The full LoRA-SFT setup is detailed in Appendix~\ref{appendix_exp_sft}.



\subsubsection{Model merging} 
\label{subsection_exp_merging}

For weight merging, we selected the top two LoRA-SFT models for each base model on the final holdout set performance. We merged the weight vectors of these two source models across all layers. All four SLERP settings achieved comparable performance on the development set (task macro-$F_1$ $\Delta \leq$ 0.04). The selected \texttt{Merged\_LoRA\_Phi4\_SLERP} and \texttt{Merged\_LoRA\_Qwen2.5\_SLERP} models both used a $t=0.5$ uniform mix. More merging details are provided in Appendix~\ref{appendix_exp_merging}.



\subsubsection{Evaluation metrics} 
\label{subsection_exp_metric}

We used greedy decoding (i.e., \texttt{do\_sample = False}) for all model inference to ensure deterministic comparisons.  The in-training validation metric that guided early stopping was the Rouge-L $F_1$ score\footnote{Introduced in Section \ref{subsec_eval_method_fts}.}.

The performance evaluation metrics for all approaches included macro-precision (P), macro-recall (R), and macro-$F_1$ scores, averaged across the entry-level P, R, and $F_1$ values. All metrics were computed using our FTS-OBP method (Section~\ref{sec4_FTSOBP}), which used \texttt{linear\_sum\_assignment} from \texttt{scipy.optimize} \citep{scipy} and computed Rouge-L $F_1$-scores using the \texttt{rouge-score} library\footnote{\url{https://pypi.org/project/rouge-score/}}.

For FTS-OBP, we used the following stopwords:  [\textit{a, an, the, is, are, was, were, be, to, of, and, in, this, that, have, it, very, really, extremely, super, absolutely, definitely}]. We set the component score threshold value $T$ based on the gold token sequence length $|g|$ as follows:  $T = 0.5$ for $|g| \in [0, 2]$,  \quad $T = 0.6$ for $|g| \in [3, 4]$,  \quad $T = 0.7$ for $|g| \geq 5$. 


\subsection{Results}\label{subsec_results}

This section presents the experimental results for the case study using the EduRABSA dataset to illustrate multitask performance. The supplementary results for the two single-task benchmark datasets are provided in Appendix~\ref{appendix_F_semeval_results} and largely mirror the findings for the three RQs presented below. 



\subsubsection{Results for RQ1 - Pre-trained GLM/SLM performance}
\label{subsec_result_RQ1}

\begin{figure*}[ht!]
  \centering
  \footnotesize
  \includegraphics[width=\linewidth]{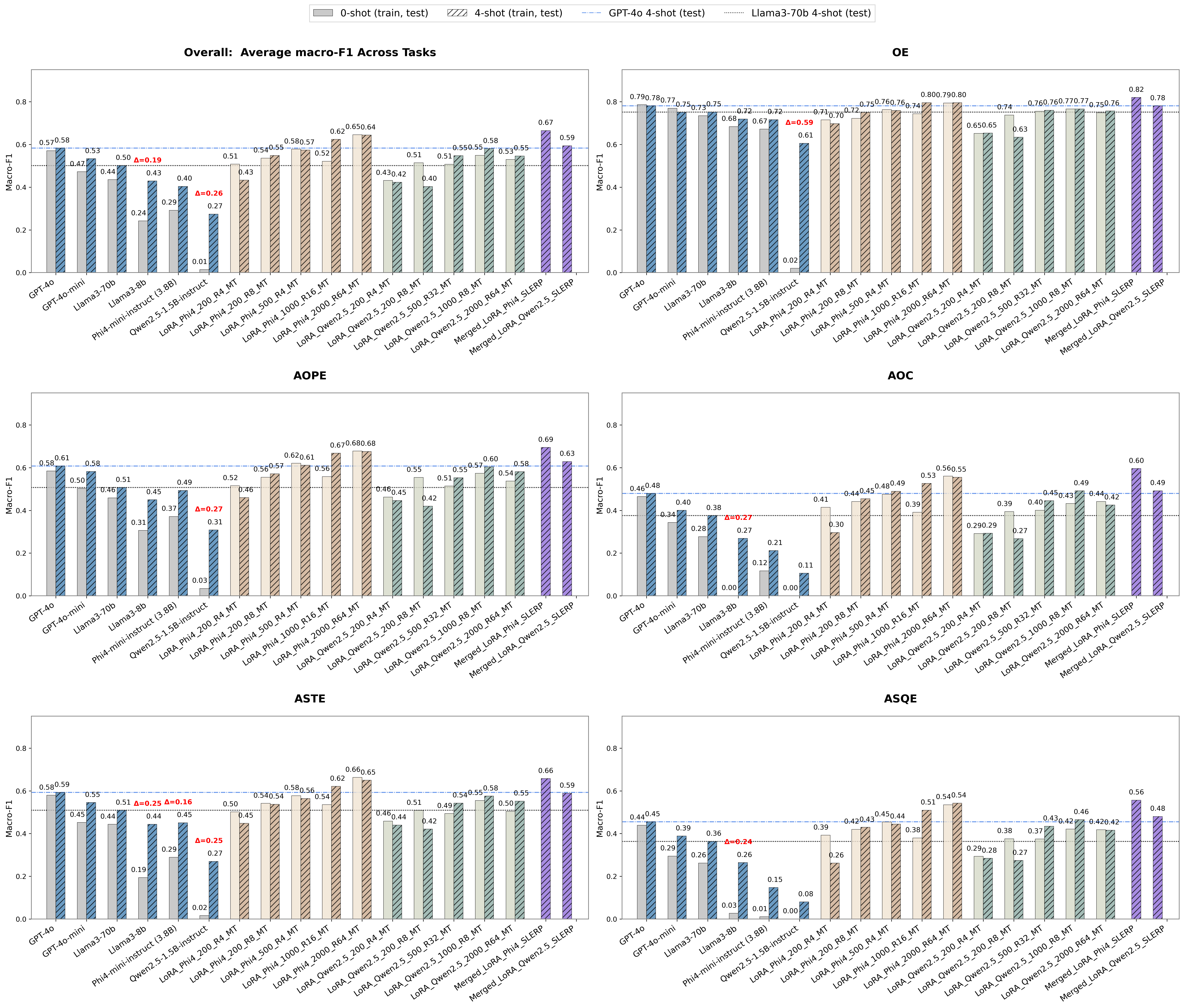}
  \caption{Macro-$F_1$ scores on five ABSA tasks (OE, AOPE, AOC, ASTE, ASQE) for pre-trained GLMs and SLMs, and LoRA-SFT and LoRA weight-merged SLMs on the EduRABSA dataset (300 test examples per task), with 0-shot (0S) and 4-shot (4S) prompt input. $\Delta$ = 4S - 0S score (\textgreater 0.15). The LoRA models were fine-tuned and tested with identical prompts. The merged models were based on 4S LoRA checkpoints.}

  \label{fig_overall_narrow_2bar_allmodels}
\end{figure*}

To address RQ1, we evaluated the four pre-trained GLMs and two SLMs under 0-shot and 4-shot prompting. As shown in Figure~\ref{fig_overall_narrow_2bar_allmodels}, performance (macro-$F_1$) varies substantially across models, primarily due to size. Large GLMs (GPT-4o and Llama3-70B) consistently outperform smaller counterparts (GPT-4o-mini and Llama3-8B) and SLMs across all tasks and ICL settings. Four-shot prompts improve all pre-trained models, particularly small GLMs and SLMs (e.g. from 0.001 to 0.27 for Llama3-8B on AOC, and 0.02 to 0.61 for Qwen2.5-1.5B on OE). Phi4-mini shows strong 4-shot performance relative to its size (3.8~B), matching or surpassing both Llama3-70B 0-shot and Llama3-8B 4-shot on OE, AOPE, and ASTE.

\subsubsection{Results for RQ2 - SLM multitask fine-tuning performance}
\label{subsec_result_RQ2}

For RQ2, we investigated the bounds and capabilities of low-resource fine-tuning by varying the key variables introduced in Section~\ref{subsubsec_dataset}, namely: 1) training set size (200, 500, 1000, or 2000 examples per task), 2) training template ICL version (0-shot vs.\ 4-shot), and 3) task order in the training set (cascade-order (CC) vs. task-type order (TT)). We also compared  multitask (MT) LoRA-SFT models with the single-task (ST) SFT counterparts with identical hyperparameters to assess the effectiveness of multitask fine-tuning. 

Overall, LoRA-SFT consistently improves the performance of both SLMs across all settings and tasks, as shown in Table~\ref{table_best_models}. The top fine-tuned models also consistently outperform (LoRA\_Phi4) or approach (LoRA\_Qwen2.5) the top tested GLM (4-shot GPT-4o) across all tasks. 

As for the role of dataset size, the SFT performance gain is already noticeable with as few as 200 training examples per task and a LoRA rank of 4, particularly with LoRA\_Phi4\_200\_R4 (0-shot) approaching or even surpassing Llama3-70B 4-shot across tasks. Nonetheless, the best-performing LoRA variants are obtained with 1000-example (LoRA\_Qwen2.5, rank~8) and 2000-example (LoRA\_Phi4, rank~64) training sets, highlighting the benefits of greater data diversity.

Figure~\ref{fig_overall_narrow_2bar_allmodels} illustrates the interaction between training set size and ICL template. While 4-shot prompts provide a clear benefit for pre-trained small GLMs and SLMs, they only consistently enhance fine-tuning performance with larger training datasets (1000 and 2000 examples per task). The interaction is less clear with the two benchmark datasets as shown in Figure~\ref{fig_semeval_2bar_allmodels}, likely due to differences in task difficulty and the SLMs’s pre-training exposure.

Finally, regarding the impact of multitask fine-tuning and data arrangement, Table~\ref{table_mtst_ttcc_traintime} shows that multitask SFT achieves performance comparable to its single-task counterparts while requiring only 42\% to 86\% of their combined training time. In terms of multitask data arrangement, the default CC setting generally yields better performance, particularly for Qwen2.5-1.5B on small training sets (200-500 entries/task), occasionally at the cost of longer training time.

Results on the benchmark datasets largely supported the above. As shown in Figure~\ref{fig_semeval_metric_comparison_bar_graph}, the best LoRA-SFT and merged LoRA models outperformed both proprietary GLMs on these datasets, and approached the non-GLM/SLM benchmarks for ACOS Laptop (exact-match macro-$F_1$). This was achieved without the full training set or the benchmark's sophisticated pipeline with contrastive learning and additional knowledge injection, regularisation, and decoding modules \citep{ref_2025_t5absa}.

\subsubsection{Results for RQ3 - Weight merging effectiveness}
\label{subsec_result_RQ3}

Table~\ref{table_best_models} and Figures~\ref{fig_overall_narrow_2bar_allmodels},~\ref{fig_semeval_2bar_allmodels} show the effectiveness of the SLERP weight-merging technique, which further improves the source LoRA-SFT models' performance across datasets and tasks\footnote{With only one exception of merged Qwen2.5 on ACOS Laptop.}. The method is also highly efficient, requiring less than 30~seconds to merge two Qwen2.5-1.5B models and under 60~seconds for the Phi4-mini models.

\renewcommand{\arraystretch}{1.4}

\begin{table}[htbp!]

\captionsetup{skip=0.5em}  
\caption{Performance of pre-trained, LoRA-SFT, and weight-merged-LoRA GLMs and SLMs across five ABSA tasks on the EduRABSA test set (300 examples/task). Metrics: macro-precision (P), macro-recall (R), and macro-$F_1$.}
\label{table_best_models}

\resizebox{\textwidth}{!}{  

\begin{tabular}{l@{\hspace{5pt}}l|
    l@{\hspace{5pt}} l@{\hspace{5pt}} l |
    l@{\hspace{5pt}} l@{\hspace{-5pt}} l |
    l@{\hspace{5pt}} l@{} l |
    l@{\hspace{5pt}} l@{\hspace{-5pt}} l |
    l@{\hspace{5pt}} l@{\hspace{-5pt}} l |
    l@{\hspace{-40pt}} l@{\hspace{5pt}} l
    }

\toprule
\multicolumn{2}{r|}{Task} &  & \textbf{OE}  &  &  & \textbf{AOPE} &  &  & \textbf{AOC} &  &  & \textbf{ASTE} &  &  & \textbf{ASQE} &  & \textbf{Task Averqage} &  &  \\ 
SN & Model & P & R & $F_1$ & P & R & $F_1$ & P & R & $F_1$ & P & R & $F_1$ & P & R & $F_1$ & P & R & $F_1$ \\

\midrule

1 & \textbf{GPT-4o} & 0.79 & 0.79 & 0.78 & 0.60 & 0.63 & 0.61 & 0.48 & 0.49 & 0.48 & 0.60 & 0.61 & 0.59 & 0.46 & 0.47 & 0.45 & \textbf{0.58} & \textbf{0.60} & \textbf{0.58} \\
2 & GPT-4o-mini & 0.77 & 0.75 & 0.75 & 0.59 & 0.60 & 0.58 & 0.40 & 0.41 & 0.40 & 0.54 & 0.57 & 0.55 & 0.39 & 0.40 & 0.39 & 0.54 & 0.55 & 0.53 \\
3 & Llama3-70b & 0.76 & 0.77 & 0.75 & 0.48 & 0.57 & 0.51 & 0.37 & 0.40 & 0.38 & 0.49 & 0.56 & 0.51 & 0.35 & 0.39 & 0.36 & 0.49 & 0.54 & 0.50 \\
4 & Llama3-8b & 0.72 & 0.75 & 0.72 & 0.45 & 0.47 & 0.45 & 0.27 & 0.28 & 0.27 & 0.45 & 0.46 & 0.44 & 0.27 & 0.27 & 0.26 & 0.43 & 0.45 & 0.43 \\
5 & Phi4-mini-instruct (3.8B) & 0.72 & 0.74 & 0.72 & 0.52 & 0.50 & 0.49 & 0.24 & 0.20 & 0.21 & 0.49 & 0.44 & 0.45 & 0.18 & 0.14 & 0.15 & 0.43 & 0.41 & 0.40 \\
6 & Qwen2.5-1.5B-instruct & 0.65 & 0.60 & 0.61 & 0.36 & 0.29 & 0.31 & 0.13 & 0.10 & 0.11 & 0.30 & 0.26 & 0.27 & 0.09 & 0.08 & 0.08 & 0.31 & 0.26 & 0.27 \\
\lightrule
7 & LoRA\_Phi4\_200\_R4\_MT\_0S & 0.76 & 0.72 & 0.71 & 0.57 & 0.52 & 0.52 & 0.46 & 0.41 & 0.41 & 0.53 & 0.51 & 0.50 & 0.43 & 0.40 & 0.39 & 0.55 & 0.51 & 0.51 \\
8 & LoRA\_Phi4\_200\_R8\_MT\_0S & 0.77 & 0.72 & 0.72 & 0.59 & 0.55 & 0.56 & 0.49 & 0.43 & 0.44 & 0.58 & 0.54 & 0.54 & 0.46 & 0.41 & 0.42 & 0.58 & 0.53 & 0.54 \\
9 & LoRA\_Phi4\_500\_R4\_MT\_0S & 0.78 & 0.77 & 0.76 & 0.64 & 0.62 & 0.62 & 0.50 & 0.47 & 0.48 & 0.60 & 0.58 & 0.58 & 0.48 & 0.45 & 0.45 & 0.60 & 0.58 & 0.58 \\
10 & LoRA\_Phi4\_1000\_R16\_MT\_4S & 0.83 & 0.78 & 0.80 & 0.69 & 0.66 & 0.67 & 0.56 & 0.51 & 0.53 & 0.65 & 0.61 & 0.62 & 0.54 & 0.49 & 0.51 & 0.65 & 0.61 & 0.62 \\
11 & \textbf{LoRA\_Phi4\_2000\_R64\_MT\_4S} & 0.81 & 0.80 & 0.80 & 0.68 & 0.68 & 0.68 & 0.56 & 0.56 & 0.55 & 0.66 & 0.66 & 0.65 & 0.55 & 0.55 & 0.54 & \textbf{0.65} & \textbf{0.65} & \textbf{0.64} \\
\lightrule
12 & LoRA\_Qwen2.5\_200\_R4\_MT\_0S & 0.71 & 0.63 & 0.65 & 0.51 & 0.44 & 0.46 & 0.33 & 0.28 & 0.29 & 0.49 & 0.45 & 0.46 & 0.32 & 0.28 & 0.29 & 0.47 & 0.42 & 0.43 \\
13 & LoRA\_Qwen2.5\_200\_R8\_MT\_0S & 0.76 & 0.74 & 0.74 & 0.58 & 0.55 & 0.55 & 0.42 & 0.39 & 0.39 & 0.53 & 0.51 & 0.51 & 0.39 & 0.38 & 0.38 & 0.53 & 0.51 & 0.51 \\
14 & LoRA\_Qwen2.5\_500\_R32\_MT\_4S & 0.79 & 0.77 & 0.76 & 0.56 & 0.57 & 0.55 & 0.47 & 0.45 & 0.45 & 0.56 & 0.55 & 0.54 & 0.45 & 0.44 & 0.43 & 0.57 & 0.56 & 0.55 \\
15 & \textbf{LoRA\_Qwen2.5\_1000\_R8\_MT\_4S} & 0.77 & 0.79 & 0.77 & 0.61 & 0.62 & 0.60 & 0.50 & 0.50 & 0.49 & 0.59 & 0.59 & 0.58 & 0.48 & 0.47 & 0.46 & \textbf{0.59} & \textbf{0.59} & \textbf{0.58} \\
16 & LoRA\_Qwen2.5\_2000\_R64\_MT\_4S & 0.77 & 0.77 & 0.76 & 0.60 & 0.58 & 0.58 & 0.44 & 0.42 & 0.42 & 0.57 & 0.55 & 0.55 & 0.43 & 0.41 & 0.42 & 0.56 & 0.55 & 0.55 \\
\lightrule
17 & \textbf{merged\_Phi4\_SLERP} & 0.84 & 0.82 & 0.82 & 0.71 & 0.70 & 0.69 & 0.61 & 0.59 & 0.60 & 0.67 & 0.66 & 0.66 & 0.57 & 0.55 & 0.56 & \textbf{0.68} & \textbf{0.66} & \textbf{0.67} \\
18 & \textbf{merged\_Qwen2.5\_SLERP} & 0.79 & 0.80 & 0.78 & 0.64 & 0.64 & 0.63 & 0.51 & 0.49 & 0.49 & 0.60 & 0.60 & 0.59 & 0.49 & 0.48 & 0.48 & \textbf{0.60} & \textbf{0.60} & \textbf{0.59} \\ 

\bottomrule
\end{tabular}
} 

\caption*{
\begin{minipage}{\textwidth}
\footnotesize

{
    \setlength{\baselineskip}{1.15\baselineskip}
    \setlength{\parskip}{0.25em}
    
    \hangindent=4.1em \hangafter=1
    \textit{* Note 1.} All pre-trained models (\#1–6) were evaluated with 4-shot prompts per task. The LoRA (\#7–16) and merged LoRA (\#17–18) models were trained and tested using the same 0-shot (0S) or 4-shot (4S) prompts, as indicated in their names. \par
    \hangindent=4.1em \hangafter=1
    \textit{* Note 2.} Models \#17 and \#18 were produced by merging models \#10–11 and \#15–16, respectively. Their results are compared with their corresponding source models.\par
}
\end{minipage}
}

\end{table}

\renewcommand{\arraystretch}{1}


\renewcommand{\arraystretch}{1.4}
\begin{table*}[htbp!]

\scriptsize
\centering

\captionsetup{skip=0.5em}  

\caption{Performance comparison of LoRA-SFT models trained under multitask task-type-order (TT) and single-task (ST) settings against the default multitask cascade-order (MT) baseline, using otherwise identical hyperparameters and early stopping. Each metric group shows the raw value for MT and the differences of the alternative methods from MT ($\Delta$ TT = TT$-$MT; $\Delta$ ST = ST$-$MT). Metrics include: 1) mean macro-$F_1$ (``mean task score'') across five ABSA tasks (OE, AOPE, AOC, ASTE, ASQE); 2) total training time (seconds); and 3) number of training epochs used. For ST, metrics 2 and 3 represent means across the five task-specific models.}
\label{table_mtst_ttcc_traintime}

\begin{tabular}{l | rrr | rrr | rrr}
\toprule
 & \multicolumn{3}{c|}{\textbf{Mean Task Score}} & \multicolumn{3}{c|}{\textbf{Train Time (sec.)}} & \multicolumn{3}{c}{\textbf{Training Epoch}} \\

Model & MT & $\Delta$TT & $\Delta$ST & MT & $\Delta$TT & $\Delta$ST & MT & $\Delta$TT & $\Delta$ST \\

\midrule

LoRA\_Phi4\_200\_R4\_0S & 0.51 & 0.01 & -0.01 & 1,362 & 334 & 688 & 2.00 & 0.50 & 1.00 \\
LoRA\_Phi4\_200\_R8\_0S & 0.54 & -0.02 & -0.04 & 1,730 & -69 & 344 & 2.50 & 0.00 & 0.50 \\
LoRA\_Phi4\_500\_R4\_0S & 0.58 & -0.06 & -0.03 & 1,999 & -22 & 1,405 & 1.20 & 0.00 & 0.80 \\
LoRA\_Phi4\_1000\_R16\_4S & 0.62 & -0.04 & 0.02 & 3,004 & -461 & 2,629 & 0.70 & -0.10 & 0.94 \\
LoRA\_Phi4\_2000\_R64\_4S & 0.64 & -0.05 & 0.03 & 3,889 & 127 & 5,344 & 0.45 & 0.00 & 0.93 \\
LoRA\_Qwen2.5\_200\_R4\_0S & 0.43 & -0.14 & 0.06 & 1,204 & -75 & 205 & 2.50 & 0.00 & 0.50 \\
LoRA\_Qwen2.5\_200\_R8\_0S & 0.51 & -0.20 & -0.01 & 1,183 & -48 & 198 & 2.50 & 0.00 & 0.50 \\
LoRA\_Qwen2.5\_500\_R32\_4S & 0.55 & -0.13 & 0.03 & 2,287 & -347 & 1,490 & 1.60 & -0.20 & 1.00 \\
LoRA\_Qwen2.5\_1000\_R8\_4S & 0.58 & -0.04 & 0.03 & 2,577 & -639 & 1,018 & 0.90 & -0.20 & 0.86 \\
LoRA\_Qwen2.5\_2000\_R64\_4S & 0.55 & 0.05 & 0.05 & 2,818 & 2 & 2,023 & 0.50 & 0.00 & 0.66 \\ 

\bottomrule
\end{tabular}

\end{table*}
\renewcommand{\arraystretch}{1}


\subsubsection{Component-level Performance}\label{subsec_result_additional}

Beyond the RQs, we also leverage the component-specific metrics from FTS-OBP to evaluate component-level performance and provide detailed results in Appendix \ref{appendix_E_detailed_results}. 

As illustrated in Figure~\ref{fig_component_metrics} and Tables~\ref{table_matching_result_asp_opn} and~\ref{table_matching_result_cat_sent}, all models perform best on the sentiment classification component. This aligns with previous findings that pre-trained GLMs already outperform traditional ML approaches \citep{ref_2025_gpt_on_sa_1, ref_2025_glms_on_sa_2} on sentiment analysis and ASC \citep{ding_llm_2024} tasks, and is further supported by the results from the two benchmark datasets (Figures~\ref{fig_acos_component_metrics} and \ref{fig_asqp_component_metrics}). Among the two text-extraction components, all models perform better at extracting opinions than aspects. This trend is largely consistent with the results for ASQP Rest16 (Figure~\ref{fig_asqp_component_metrics}), which also extracts implicit opinion expressions, and may be attributed to the EduRABSA dataset’s coreference resolution rules (e.g., prioritising the teacher’s name from an adjacent sentence over a same-sentence pronoun as the aspect) \citep{hua2025edurabsa}, or to a potentially large number of implicit aspects. 

Figure~\ref{fig_asp_opn_heatmap} further reveals the error patterns across models for each component. For aspects, incorrect implicit aspect extraction accounted for 7.15\% to 28.13\% of all rejected pairs, with LoRA-Phi4 on the smallest training set (200 examples/task) performing worse than the pre-trained base model. Moreover, most LoRA-SFT SLMs underperformed the pre-trained GLMs in implicit aspect extraction, potentially due to insufficient examples and/or low adaptor ranks. The pre-trained Qwen2.5-1.5B model had the highest hallucination rate among all models for both aspects (11.77\%, with the second highest being 6.25\%) and opinions (30.70\%, versus the second highest of 10.88\%). Overall, opinion extraction showed greater boundary variability across models than aspect extraction, in terms of the total percentage of under- and over-extraction in both accepted and rejected pairs.  With the two benchmark datasets (Figures~\ref{fig_ACOS_asp_opn_heatmap} and~\ref{fig_ASQP_asp_opn_heatmap}), fine-tuning also seems to negatively affect implicit aspect extraction, particularly with larger training set sizes. In addition, the ``NULL'' notation of implicit opinions in the ACOS Laptop dataset appears to have posed a challenge to all models, accounting for up to 67.69\% of all rejected opinion pairs.

As for aspect categories, Figure~\ref{fig_cat_boxplot_heatmap} shows the greatest variability across models in categories that are less clearly distinguished from others (e.g. ``Staff - Helpfulness'' and ``Staff - Personal traits''; ``Course - Content'' vs. ``Course - Course materials''), as well as in minority labels under the \textit{University} main category.


\section{Discussion and Conclusion}\label{sec6_discussion}

In this study, we proposed \textbf{FTS-OBP}, a new evaluation method tailored for ABSA tasks involving aspect and opinion extraction components. We further used education reviews as an example of a low-resource domain to investigate the effectiveness of three resource-efficient multitask approaches for ABSA. 

Comparison of FTS-OBP with exact-match methods revealed a strong positive correlation between the two evaluation approaches. FTS-OBP overcomes the rigidity of traditional exact-match evaluation by introducing a controllable acceptance range for aspect and opinion boundary variations, while maintaining strictness in classification component matching. The largest metric differences appeared with outputs from pre-trained GLMs, which the traditional method penalised for minor boundary deviations despite otherwise correct components. This suggests that FTS-OBP is particularly relevant to advancing ABSA research as multi-component tasks and generative approaches gain traction in the field \citep{hua2024absa}, while its fine-grained, component-level scoring provides additional metrics valuable for diagnostic analysis.

Regarding resource-efficient ABSA methods, we explored three approaches: ICL with pre-trained GLMs and SLMs, LoRA-SFT on SLMs using only 200--2000 instances per dataset, and weight merging of LoRA-SFT SLMs. Overall, our findings reveal the promising potential of all three approaches: 1) Pre-trained GLMs (GPT-4o, Llama3-70B) achieved strong baseline (0-shot) ICL performance, while 4-shot prompts substantially improved small GLM and SLM performance to match or surpass larger ones, demonstrating the effectiveness of ICL without additional training. 2) Multitask LoRA-SFT proved highly efficient, using only 42\%–86\% of total training time and 1/5 of the memory required for single-task setups to achieve the same level of performance, and outperforming larger GLMs even with as few as 200 examples per task. 3) Weight merging further improved LoRA-SFT performance with minimal additional cost, likely by leveraging complementary error patterns across source models, highlighting its promise for efficient model reuse in low-resource ABSA.

Finally, component-specific analyses revealed consistent cross-task performance patterns. On the EduRABSA dataset, which is more complex and annotated with aspect coreference resolution rules and under-explored implicit opinion extraction components \citep{hua2025edurabsa}, all models performed well on the sentiment classification component but struggled with aspect extraction and categorisation. Low-data fine-tuning also appears to hurt performance on implicit aspect extraction, which warrants further investigation. These findings highlight potential directions for future research, both in improving component-specific model capabilities and in dataset improvement.


\subsection{Limitations and Future Work}\label{subsec6_2_limitations}

Our work has several limitations. First, as there is no prior benchmark for the EduRABSA dataset, we compared model performance only against a limited set of GLMs. Second, although we included two benchmark datasets and compared our results with previously reported benchmarks, we did not evaluate our approaches on datasets from other non-commercial domains. Lastly, as our experiments aimed to identify the resource lower bound, we limited LoRA-SFT training to 2000 examples per task, and therefore cannot draw conclusions about the potential upper performance bound of fine-tuned SLMs.

Future research could further explore and validate FTS-OBP using more diverse datasets and controlled experiments to enable finer-grained analysis and comparison. For low-resource domain ABSA, additional work on discovering and improving resource-efficient approaches remains important, particularly in developing efficient evaluation methods for domains lacking public annotated dataset, advancing data-light or unsupervised learning techniques, and enhancing GLM and SLM performance on domain-specific ABSA tasks, particularly aspect/opinion extraction and aspect categorisation, through strategies such as contrastive learning, data augmentation, and multi-GLM frameworks.

\clearpage
\newpage
\onecolumn

\appendix

\setcounter{figure}{0}
\setcounter{table}{0}

\renewcommand\thefigure{\thesection.\arabic{figure}}
\renewcommand\thetable{\thesection.\arabic{table}}
\renewcommand\theHtable{\thesection.\thetable}
\renewcommand{\thefigure}{\thesection.\arabic{figure}} 
\renewcommand{\theHfigure}{\thesection.\arabic{figure}} 

\begin{center}
{\large\bfseries APPENDICES}
\end{center}
\vspace{1.5ex}


\section{Related Work} \label{Appendix_0_related_work}

\subsection{Existing ABSA approaches} \label{subsec_existing_approaches}

Given ABSA’s fine granularity and dependence on context and domain, past research has focused on incorporating contextual relationships and domain knowledge into solution systems \citep{hua2024absa}, with common approaches involving \citep{hua2024absa, batch2_survey_absa, batch2_survey_absadl, ref_2025_01}: 
1) heuristics based on linguistic rules and syntactic features (e.g. POS tags, dependencies);  
2) embedding layers from learned or pre-trained static models (e.g. Word2Vec \citep{word2vec}, GloVe \citep{glove}) or context-aware encoder-only LLMs (e.g. BERT \citep{bert}, RoBERTa \citep{roberta}) within neural architectures (often RNNs with classifier heads);  
3) linguistic and domain resources such as sentiment lexicons and ontologies; and  
4) context selection mechanisms via model design (e.g. attention, graph-based structures) or training strategies (e.g. auxiliary tasks, multi-task learning).  In terms of architecture, traditional approaches adopt a pipeline framework by chaining single-element task outputs sequentially (e.g. ASC on the output of AE), while an increasing number of recent studies explored the end-to-end unified approach \citep{batch2_survey_absa, jointabsa} that uses a single model to solve one composite task (e.g. ASTE, ASQE) or multiple subtasks simultaneously. 

The aforementioned common ABSA approaches \citep{hua2024absa} have several limitations. First, the pipeline framework's modularised architecture with disjoint subtasks is prone to error propagation, context isolation, and representational bottlenecks between interrelated components \citep{sk2_jointabsa, jointabsa, batch2_survey_absa, implicitOE_7}, which is particularly problematic for ABSA where components are tightly intertwined. Second, manual feature engineering and rule-based methods have limited generalisability \citep{Zhang2024}, and often involve preprocessing (e.g. stop-word removal, stemming) that reduces input richness and struggles to capture linguistic variability and contextual subtleties. Learned or static representations were similarly limited to the dataset vocabulary, lacking contextual understanding, general language knowledge, and generalisability \citep{batch2_survey_absa, ding_llm_2024}. Whilst encoder-only LLMs improve ABSA performance through contextualised features and pre-trained language knowledge \citep{batch2_survey_absa, batch2_survey_absadl}, the prevalent combination of encoder-only LLM embeddings with deep neural modules and classifiers \citep{batch2_survey_absa, hua2024absa} still suffers from pipeline limitations and poor adaptability to new labels, formats, or domains without retraining. Third, traditional ML/DL approaches typically formulate aspect and opinion extraction as sequence labelling tasks, which face challenges with implicit aspects and many-to-one aspect-opinion relationships \citep{joint_aste, jointabsa}.

\subsubsection{The Potential of Generative LLMs (GLMs)}

Pre-trained generative LLMs with decoder-only architecture (``generative LLMs'' or ``GLMs'') offer multiple characteristics that address these limitations. These benefits are particularly pronounced in foundation models \citep{foundationmodels, surveyincontextlearning} with billions to trillions of parameters pre-trained on diverse data (e.g. GPT \citep{gpt4}, Claude \citep{claude3}, Llama \citep{llama3} families). For ABSA tasks, the primary strengths of pre-trained GLMs include:
    \begin{enumerate}
    
        \item The in-context learning (ICL) \citep{refGPT3} capability, which enables models to learn new tasks through prompts alone (zero-shot) or with a few examples (few-shot) without fine-tuning \citep{refGPT3, surveyincontextlearning}, allowing easy adaptation to new output formats and sentiment or category labels.

        \item Rich representations from larger training data (e.g. 15 trillion tokens for Llama3 \citep{llama3} vs. 3300 million for BERT \citep{bert}) and parameter size than earlier LLMs, which can capture broader linguistic patterns and general knowledge beneficial for ABSA.
    
        \item Strong performance across diverse task types, from reasoning to general- and domain-specific knowledge \citep{mmlu, llama3, gpt4, phi4} that are important for ABSA, with early foundation GLMs already outperforming traditional ML on sentiment analysis \citep{ref_2025_gpt_on_sa_1, ref_2025_glms_on_sa_2}.
    
        \item The single-model architecture with unified attention mechanisms avoids representational bottlenecks and can perform multiple ABSA subtasks without any architectural changes. Their generative capability is particularly valuable for handling implicit aspects and complex aspect–opinion relationships that pose challenges for traditional approaches.

        \item The recent arrival of small GLMs (SLMs) offers a promising solution for ABSA in low-resource, high-data-restriction domains such as healthcare and education review. Model families as Phi-3 \citep{phi3}, Phi-4 \citep{phi4mini}, and Qwen-2.5 \citep{qwen25} offer 0.5-7~B parameter variants that fit single GPUs while providing 32k-128k context length and performance comparable to models twice their size \citep{phi4mini, qwen25}. 

    \end{enumerate}

Despite these benefits, GLMs remain underexplored in ABSA \citep{hua2024absa, bai_llm_2024, zhou_llm_2024}. Even recent ABSA studies with generative approaches still mainly used smaller encoder-decoder models like T5 \citep{t5} (e.g. \citep{ref_2025_t5absa, ref_2025_decoupleabsa}). A few studies\footnote{We excluded studies focusing on cross-lingual transfer.} tested GLMs in ABSA, but were all on product reviews (\citep{ding_llm_2024, lee_llm_2024, bai_llm_2024, zhou_llm_2024, llm_czechabsa_2025}). Their results suggested that GLMs (mostly with 7~B or more parameters) after Parameter-efficient fine-tuning (PEFT) consistently outperformed fully-trained or fine-tuned non-GLM baselines across ABSA tasks \citep{lee_llm_2024, ding_llm_2024, zhou_llm_2024, llm_czechabsa_2025}, but both 0-shot and few-shot ICL lagged behind the non-GLM baselines with varied performance gaps \citep{ding_llm_2024, bai_llm_2024, zhou_llm_2024, llm_czechabsa_2025}. Given the rigidity of exact-match evaluation metrics mentioned in Section \ref{subsection_motivation}, we suspect that the prompt-based GLM performance gap partly reflects boundary variations in aspect/opinion extraction, which fine-tuning helps reduce. 

To what extent these GLM ABSA findings on product reviews can be generalised to other domains remains an open question. Three of the five ``GLM in ABSA'' studies mentioned above exclusively used the SemEval benchmark datasets \citep{semeval2014, semeval2015, semeval2016}, which have been criticised for being overly simplified and under-representing realistic review text complexity \citep{datasetsurvey2023, fei2026robustness, mams2019}. This limitation is especially evident when compared with political reviews with greater sentiment subtleties \citep{liu2012sentimentanalysisbook}, and education reviews, which are typically longer and have more multi-aspect/ multi-sentiment sentences and/or implicit aspects and opinions \citep{hua2025edurabsa}. Furthermore, none of these studies involved SLMs nor explored the impact of the training dataset size.

On the other hand, GLMs are particularly absent from education review ABSA studies due to a few key barriers: a lack of relevant annotated datasets for domain evaluation, the fact that student feedback data are often under strict access restriction and server/tool choice, and the high cost of hosting local GLMs due to their hardware requirement \citep{hua2025edurabsa}. These factors create a self-reinforcing cycle in which limited access to data hinders model development, and the absence of suitable models discourages dataset creation and sharing. To break this cycle, it is thus necessary to understand the performance of GLMs and particularly SLMs in low-resource domains under resource-efficient settings.

\newpage

\section{Method Details}\label{appendix_B_method_details}

\subsection{In-context Learning (ICL) Input Examples} \label{appendix_b1_icl_input_example}

Below are the 4-shot ICL input prompts used for ASQE task.  The lines above ``\#\#\# Examples:'' are identical for the 0-shot version.  The input for the other tasks (OE, AOPE, AOC, ASTE) can be obtained from this version by removing the lines on the components not included in that task. 

\begin{prompt}

    \#\#\# Task type: \\
    aspect-sentiment quadruplet extraction (ASQE) \par
    
    \#\#\# Instruction:  \par
    
    Given the input text, extract ALL pairs of opinion expressions and their corresponding aspect terms about the course, staff, or university. Then classify the category and sentiment for each aspect-opinion pair.  \par
    Opinion expressions are words/phrases expressing evaluation, feeling, or judgment (including both explicit and implicit opinions, not objective facts).  \par
    Aspect terms are opinion targets. Only use a pronoun if you cannot find a direct aspect term in the same sentence or adjacent context.   \par
    Each aspect-opinion-category-sentiment combination is a quadruplet.   \par
    
    **Rules:**  \par
    - Extract EVERY opinion in the text, including both explicit and implicit opinion expressions. \par
    - Extract all opinion and aspect terms VERBATIM and as CONSECUTIVE tokens.  \par
    - Use `null' for implicit aspects. Opinions cannot be null. \par
    - If an aspect is mapped to multiple opinion expressions, or vice versa, extract each 1:1 pair separately.  \par
    - Categorise each aspect-opinion pair first into one main category (the keys) in the category\_mapping below, and then into one of its appropriate subcategories (values for the key). The category label follows "Main category - subcategory" format. \par
    category\_mapping = \{ \par
      "Course": ["Content", "Learning activity", "Assessment", "Workload", "Difficulty", "Course materials", "Technology \& tools", "Overall"], \par
      "Staff": ["Teaching", "Knowledge \& skills", "Helpfulness", "Attitude", "Personal traits", "Overall"], \par
      "University": ["Cost", "Opportunities", "Programme", "Campus \& facilities", "Culture \& diversity", "Information \& Services", "Social engagement \& activities", "Overall"] \par
    \} \par
    
    - Classify the sentiment into one of `positive', `neutral', `negative'. \par
    
    - Use these specific tags for each component within each quadruplet: <asp>aspect terms</asp>, <opn>opinion expressions</opn>, <cat>category</cat>, <sen>sentiment</sen>  \par
    
    **Critical formatting requirements:**  \par
    - Output MUST be a valid Python list  \\
    - Quadruplets MUST be separated by commas \par
    
    **Output format:**  \par
    [<asp>...</asp><opn>...</opn><cat>...</cat><sen>...</sen>, <asp>...</asp><opn>...</opn><cat>...</cat><sen>...</sen>, ..., <asp>...</asp><opn>...</opn><cat>...</cat><sen>...</sen>] \par
    
    \#\#\# Examples: \par
    
    Input: "The professor was knowledgeable but the assignments were too hard." \par
    Output: [<asp>professor</asp><opn>knowledgeable</opn><cat>Staff - Knowledge \& skills</cat><sen>positive</sen>, <asp>assignments</asp><opn>too hard</opn><cat>Course - Assessment</cat><sen>negative</sen>] \par
    
    Input: "It was disappointing overall." \par
    Output: [<asp>null</asp><opn>disappointing</opn><cat>Course - Overall</cat><sen>negative</sen>] \par
    
    Input: "She never reply to emails or answer questions" \par
    Output: [<asp>She</asp><opn>never reply to emails or answer questions</opn><cat>Staff - Helpfulness</cat><sen>negative</sen>] \par
    
    Input: "There were 10 assignments, 5 quizzes, 1 final exam." \par
    Output: [<asp></asp><opn></opn><cat></cat><sen></sen>] \par
        
    \#\#\# Input:  \par
    ```<review text entry>'''

\end{prompt}

\newpage

\section{Experimental Setup Details}\label{appendix_C_experiment_details}

\subsection{Dataset} \label{appendix_exp_dataset}

We used three datasets in total. The \textbf{EduRABSA} dataset \citep{hua2025edurabsa} contains courses, teaching staff, and university reviews; and two benchmark datasets: \textbf{ASQP Rest16} \citep{zhang2021_asqp} on restaurant reviews, and \textbf{ACOS Laptop} \citep{cai2021_acos} on laptop reviews. Both the EduRABSA and ASQP Rest16 datasets extract implicit opinions (i.e., opinionated expressions without explicit sentiment-bearing words \citep{hua2025edurabsa}), whereas ACOS Laptop represents implicit opinions as ``NULL'', similar to how all three datasets handle implicit aspects. The ASQP Rest16 training set is limited to 1000 instances due to the original dataset's size constraint. 

Table~\ref{table_dataset_stats} shows the key statistics of these three datasets.


\renewcommand{\arraystretch}{1.25}

\begin{table*}[!h]
\centering
\footnotesize

\caption{Summary statistics of the three annotated ASQE datasets used in this study. For each dataset, the top row reports 1) the number of review entries by dataset split (columns 2--4), and 2) the total number of unique components per quadruplet (columns 5--8). The second row presents quadruplet counts by dataset split and sentiment polarity, where (+), (o), and (–) denote quadruplets with positive, neutral, and negative sentiment labels, respectively. Detailed statistics for the ASQP Rest16 and ACOS Laptop datasets are taken from \citep{compound_task_survey_2025}.}

\label{table_dataset_stats}

\begin{tabular}{l | ccc | cccc}

\toprule
\textbf{Dataset} & \textbf{Train} & \textbf{Val} & \textbf{Test} & \textbf{Aspect} & \textbf{Opinion} & \textbf{Category} & \textbf{Sentiment} \\
 & (+, o, -) & (+, o, -) & (+, o, -) &  &  &  &  \\ 
\midrule

\textbf{ASQP Rest16} & 1264 & 316 & 544 & 2853 & 3040 & 2754 & 2279 \\
 & (1369, 62, 558) & (341, 23, 143) & (584, 40, 177) &  &  &  &  \\
 \lightrule
\textbf{ACOS Laptop} & 2934 & 326 & 816 & 4958 & 5378 & 4992 & 4958 \\
 & (2583, 227, 1364) & (279, 24, 137) & (716, 65, 380) &  &  &  &  \\
 \lightrule
\textbf{EduRABSA} & 4000 & / & 2500 & 16884 \textsuperscript{\textcolor{myblue}{*}} & 26533 & 18148 & 10510 \\
 & (9581, 1713, 5206) & / & (5994, 1049, 3494) &  &  &  &  \\ 

\bottomrule

\end{tabular}

\caption*{
\begin{minipage}{\textwidth}
\footnotesize
\setlength{\baselineskip}{1.2\baselineskip}
    \hangindent=3em \hangafter=1
    \quad \textbf{*} EduRABSA: Quadruplets with implicit aspect = 2,456.
\end{minipage}
}

\end{table*}

\subsubsection{Dataset Processing Details} 
\label{appendix_c2_dataset_processing_details}

We used four mutually exclusive subsets of the EduRABSA dataset \citep{hua2025edurabsa} for all model training and evaluation. The \textbf{``train''} and \textbf{``validation''} sets were split from the default training set, with the latter used for in-training evaluation (for early stopping). For post-training evaluation, we split the default test set into a \textbf{``development''} set and a \textbf{``test''} set, with the former used to inform hyper-parameter tuning, and the latter providing the final experimental results reported in this section. All subsampling maintained equal ratios of course reviews and teaching-staff reviews while maximising the number of available university review entries.

For the ASQP Rest16 and ACOS Laptop datasets, we detokenised the pre-tokenised review text and annotations to restore natural language format, preventing interference with the GLMs' and SLMs' native tokenisers. We then subsampled from the original train, dev (used as our validation set), and test splits, with the ASQP Rest16 training set capped at 1000 instances due to the limited size of the original dataset.

We prepared the datasets for model input as follows: 

\begin{enumerate}
    \item Task conversion: We first converted each review entry's ASQE annotation into five task-specific versions by removing extra components from its four copies. 

    \item Annotation formatting: We then reformatted each input entry's annotation into an array of strings, where each string represents a unit (e.g. triplets, quadruplets) with component boundaries marked by XML-style tags:
        
        \small \centering
                $a_j \rightarrow$  \texttt{<asp>}$a_j$\texttt{</asp>}, 
                $o_j \rightarrow$  \texttt{<opn>}$o_j$\texttt{</opn>}, 
                $c_j \rightarrow$  \texttt{<cat>}$c_j$\texttt{</cat>},
                $s_j \rightarrow$  \texttt{<sen>}$s_j$\texttt{</sen>}.

    \item Model input construction: We combined each review text with one of two instruction prompts (a 4-shot version is demonstrated in Appendix \ref{appendix_b1_icl_input_example}), creating inputs for all five task-specific versions. For the training and validation sets, we appended the formatted annotations and applied each SLM's chat template. 

    \item Task ordering: Finally, we selected and arranged entries according to the MT vs. ST and CT vs. TT configurations described in Section \ref{appendix_exp_dataset}. We chose not to convert the XML-style component tags into custom tokens to facilitate output parsing.  

\end{enumerate}

\subsection{LoRA supervised fine-tuning (SFT)} \label{appendix_exp_sft}

For each training dataset size $T\in$ \{200, 500, 1000, 2000\}, we applied LoRA SFT with different LoRA ranks ($R$) and adjusted the hyperparameters accordingly. The LoRA rank values include $R \in$ \{4, 8, 16, 24, 32, 48\} for all $T$, and additionally $R \in$ \{64, 80\} for $T \geq 500$, $R = 96$ for $T \geq 1000$, and $R=128$ for $T = 2000$. Table~\ref{table_training_param} shows the key hyperparameters of the final selected LoRA-SFT models. Note that to compare the effect of model size, the LoRA\_Qwen2.5 models directly adopted the LoRA rank values of the final selected LoRA\_Phi4 models. 

We report the final selected LoRA-SFT models based on the development-set performance and the \texttt{T200\_R4} model for its minimum training set size $T$ and LoRA rank $R$. We refer to these LoRA-SFT models with the pattern: 

    \texttt{\small LoRA\_<base model>\_<per-task train size>\_<LoRA rank>\_<isMultitask>\_<prompt type>},  
    
    
where  \texttt{isMultitask} $\in$ \{multitask (MT), single-task (ST)\}, \quad \texttt{prompt type} $\in$ \{0-shot (0S), 4-shot (4S)\}.

\renewcommand{\arraystretch}{1.2}

\begin{table*}[!htbp]
\centering
\scriptsize

\captionsetup{skip=0.5em}  

\caption{Key LoRA Supervised-fine-tuning (SFT) hyper-parameters for the final selected models}

\label{table_training_param}

\begin{tabularx}{\textwidth}{
>{\raggedright\arraybackslash}p{0.22\linewidth} |
>{\raggedright\arraybackslash}p{0.05\linewidth} 
>{\raggedright\arraybackslash}p{0.05\linewidth}
>{\raggedright\arraybackslash}p{0.05\linewidth}
>{\raggedright\arraybackslash}p{0.05\linewidth}
>{\raggedright\arraybackslash}p{0.05\linewidth} |
>{\raggedright\arraybackslash}p{0.05\linewidth}
>{\raggedright\arraybackslash}p{0.05\linewidth}
>{\raggedright\arraybackslash}p{0.05\linewidth}
>{\raggedright\arraybackslash}p{0.05\linewidth}
>{\raggedright\arraybackslash}p{0.05\linewidth}
}

\toprule
\multicolumn{1}{r|}{\textbf{Base model}} & \multicolumn{5}{c|}{\textbf{Phi-4-mini-instruct}} & \multicolumn{5}{c}{\textbf{Qwen2.5-1.5B-instruct}} \\
\multicolumn{1}{r|}{\textbf{Version}} & \textbf{T200 R4} & \textbf{T200 R8} & \textbf{T500 R4} & \textbf{T1k R16} & \textbf{T2k R64} & \textbf{T200 R4} & \textbf{T200 R8} & \textbf{T500 R32} & \textbf{T1k\quad R8} & \textbf{T2k R64} \\
\midrule

\textbf{Training dataset size (T) per task} & 200 & 200 & 500 & 1000 & 2000 & 200 & 200 & 500 & 1000 & 2000 \\
\textbf{Validation dataset size per task} & 200 & 200 & 200 & 200 & 200 & 200 & 200 & 200 & 200 & 200 \\
\textbf{Learning rate} & 5.0e-04 & 2.4e-04 & 6.0e-04 & 3.0e-04 & 9.0e-05 & 6.0e-04 & 2.9e-04 & 1.5e-04 & 4.8e-04 & 1.1e-04 \\
\textbf{Weight decay} & 0.0008 & 0.0015 & 0.0015 & 0.003 & 0.0015 & 0.0007 & 0.0012 & 0.0009 & 0.0028 & 0.0015 \\
\textbf{Label smoothing} & 0.15 & 0.15 & 0.12 & 0.1 & 0.05 & 0.15 & 0.15 & 0.06 & 0.1 & 0.06 \\
\textbf{Warmup ratio} & 0.1 & 0.1 & 0.1 & 0.1 & 0.1 & 0.1 & 0.1 & 0.1 & 0.1 & 0.1 \\
\textbf{Epoch} & 3 & 3 & 3 & 3 & 3 & 3 & 3 & 3 & 3 & 3 \\
\textbf{Gradient accumulation steps} & 10 & 10 & 10 & 10 & 10 & 10 & 10 & 10 & 10 & 10 \\
\textbf{Eval steps \textsuperscript{\textcolor{myblue}{1}}} & 50 & 50 & 50 & 50 & 50 & 50 & 50 & 50 & 50 & 50 \\
\textbf{Eval gradient accumulation steps} & 10 & 10 & 10 & 10 & 10 & 10 & 10 & 10 & 10 & 10 \\
\textbf{Use RSLoRA} & TRUE & TRUE & TRUE & TRUE & TRUE & TRUE & TRUE & TRUE & TRUE & TRUE \\
\textbf{LoRA rank (R)} & 4 & 8 & 4 & 16 & 64 & 4 & 8 & 32 & 8 & 64 \\
\textbf{LoRA alpha} & 8 & 16 & 8 & 32 & 128 & 8 & 16 & 64 & 16 & 128 \\
\textbf{LoRA dropout} & 0.15 & 0.12 & 0.12 & 0.06 & 0.015 & 0.12 & 0.1 & 0.04 & 0.08 & 0.025 \\
\textbf{LoRA target\_modules} & AL\textsuperscript{\textcolor{myblue}{2}} & AL & AL & AL & AL & AL & AL & AL & AL & AL \\
\textbf{Balance categories} & TRUE & TRUE & TRUE & TRUE & TRUE & TRUE & TRUE & TRUE & TRUE & TRUE \\
\textbf{Early Stopping} & TRUE & TRUE & TRUE & TRUE & TRUE & TRUE & TRUE & TRUE & TRUE & TRUE \\
\textbf{Early stopping patience} & 3 & 3 & 4 & 4 & 6 & 4 & 4 & 6 & 5 & 7 \\
\textbf{Early stopping threshold} & 0.001 & 0.001 & 0.001 & 0.001 & 0.0002 & 0.0012 & 0.0012 & 0.0005 & 0.0012 & 0.0005 \\
\textbf{Metric for best model} & RL $F_1$\textsuperscript{\textcolor{myblue}{3}} & RL $F_1$ & RL $F_1$ & RL $F_1$ & RL $F_1$ & RL $F_1$ & RL $F_1$ & RL $F_1$ & RL $F_1$ & RL $F_1$ \\

\bottomrule
\end{tabularx}

\begin{minipage}{\textwidth}
\footnotesize
{
    \setlength{\baselineskip}{1.15\baselineskip}
    \setlength{\parskip}{0.25em}
    
    \hangindent=1em \hangafter=1
    \textsuperscript{1} Eval steps: For single-task training, we used 10 for training set sizes 200 and 500, and 20 for training set sizes 1000 and 2000. \par
    
    \hangindent=1em \hangafter=1
    \textsuperscript{2} AL: ``all-linear'', i.e. all linear layers in the model, including attention projections and MLP layers. \par
    
    \hangindent=1em \hangafter=1
    \textsuperscript{3} RL $F_1$: Average RougeL F-measure calculated using the \texttt{\href{https://pypi.org/project/rouge-score/}{rouge\_score}} library.

}
\end{minipage}

\end{table*}
\renewcommand{\arraystretch}{1}

\subsection{Model merging} \label{appendix_exp_merging}

For the SLERP weight-merging, we tested uniform settings with $t=0.5$ and $t=0.7$, and a layer-wise scheme defined by five anchor values, with per-layer $t$ obtained by linear interpolation between successive anchors across the depth of the module. With the scheme setting, we tested both orderings of the two source models and the following configurations: For the self-attention layers, the anchors were $[0,\,0.5,\,0.3,\,0.7,\,1]$, for the MLP layers the inverse ordering $[1,\,0.5,\,0.7,\,0.3,\,0]$, and for remaining components uniform $t=0.5$ was used.

\newpage
\section{FTS-OBP Validation}\label{Appendix_ftsobp}

\subsection{FTS Matching Validation} \label{appendix_fts_validation}

A comprehensive comparison of the FTS-OBP method and the traditional approach, based on experimental results, is presented in Section \ref{subsec_exp_metric_comparison}. In this section, we focus on validating the core FTS Matching, specifically by examining the non-identical text-extraction pairs that are accepted as matches by the FTS scoring and the selected threshold values. We conducted an experiment to identify the accepted variation range as follows: We used ``$a_1\ a_2\ ...\ a_{50}$'' as the input text and systematically generated 10 gold cases with lengths of 1--10 tokens (from ``$a_1$'' to ``$a_1\ a_2\ ...\ a_{10}$''). For each gold case, we designed three scenarios of variations: 1) ``gold in pred'' -  pred extends beyond gold by up to 20 tokens (``over by $n$''), 2) ``pred in gold'' - pred truncates gold by removing the last $n \leq |g|-1$ tokens (``under by $n$''), and 3) ``partial overlap'' - pred shifts the gold boundary by up to 10 tokens whilst maintaining equal length (``shifted by $n$''). We did not include cases where pred is not in the input text, as FTS automatically rejects such invalid extractions. This generated 300 test pairs (21--39 pairs per gold length).

Table~\ref{table_eval_validation} summarises the acceptable variation ranges for a match under FTS with our chosen threshold values. The results show that FTS and our threshold values together allow more over-extraction (i.e. pred contains gold) than under-extraction, and harshly penalise boundary shifts (i.e. partial pred-gold overlap). This behaviour aligns with realistic use cases based on our observations. To illustrate this more concretely, Table~\ref{table_fts_example} shows 20 randomly chosen non-full-match gold-pred unit pairs and the FTS scores from 0-shot GPT-4o ICL (reported in Section \ref{sec5_experiments}). 

As a result of the more relaxed matching criteria, we expect FTS-OBP to produce higher scores more frequently for tasks that involve aspect and/or opinion components than the traditional approach.

\renewcommand{\arraystretch}{1}
\begin{table*}[hbp!]

\scriptsize
\centering

\captionsetup{skip=0.5em}  

\caption{Accepted boundary variation range and number of accepted cases for each gold sequence length and scenario combination under Flexible Text Similarity (FTS) scoring with dynamic threshold $T$. Total cases = 300. Exact-match cases (one per combination) are not shown in this table.}

\label{table_eval_validation}

\begin{tabular}{cc | cccc | cccc | cccc}

\toprule
& &  
\multicolumn{4}{c|}{\textbf{Scenario 1: Over by $n$}} 
& \multicolumn{4}{c|}{\textbf{Scenario 2: Under by $n$}} 
& \multicolumn{4}{c}{\textbf{Scenario 3: Shifted by $n$}} \\
\textbf{Gold len} & \textbf{Threshold} 
& \textbf{$\boldsymbol{n}$-range} & \textbf{N}\textsuperscript{\textcolor{myblue}{*}} & \textbf{A}\textsuperscript{\textcolor{myblue}{*}} & \textbf{A/N}\textsuperscript{\textcolor{myblue}{*}} 
& \textbf{$\boldsymbol{n}$-range} & \textbf{N} & \textbf{A} & \textbf{A/N} 
& \textbf{$\boldsymbol{n}$-range} & \textbf{N} & \textbf{A} & \textbf{A/N} 
\\ 

\midrule
1 & 0.5 & 1-2 & 20 & 2 & 0.10 & - & 0 & 0 & 0.00 & - & 0 & 0 & 0.00 \\
2 & 0.5 & 1-4 & 20 & 4 & 0.20 & 1 & 1 & 1 & 1.00 & 1 & 1 & 1 & 1.00 \\
3 & 0.6 & 1-4 & 20 & 4 & 0.20 & 1 & 2 & 1 & 0.50 & 1 & 2 & 1 & 0.50 \\
4 & 0.6 & 1-5 & 20 & 5 & 0.20 & 1-2 & 3 & 2 & 0.70 & 1 & 3 & 1 & 0.30 \\
5 & 0.7 & 1-4 & 20 & 4 & 0.20 & 1-2 & 4 & 2 & 0.50 & 1 & 4 & 1 & 0.20 \\
6 & 0.7 & 1-5 & 20 & 5 & 0.20 & 1-2 & 5 & 2 & 0.40 & 1 & 5 & 1 & 0.20 \\
7 & 0.7 & 1-6 & 20 & 6 & 0.30 & 1-3 & 6 & 3 & 0.50 & 1-2 & 6 & 2 & 0.30 \\
8 & 0.7 & 1-6 & 20 & 6 & 0.30 & 1-3 & 7 & 3 & 0.40 & 1-2 & 7 & 2 & 0.30 \\
9 & 0.7 & 1-7 & 20 & 7 & 0.40 & 1-4 & 8 & 4 & 0.50 & 1-2 & 8 & 2 & 0.20 \\
10 & 0.7 & 1-8 & 20 & 8 & 0.40 & 1-4 & 9 & 4 & 0.40 & 1-3 & 9 & 3 & 0.30 \\
\lightrule
TOTAL & - & - & 200 & 51 & 0.26 & - & 45 & 22 & 0.49 & - & 45 & 14 & 0.31 \\ 

\bottomrule
\end{tabular}

\smallskip
\begin{minipage}{\textwidth}
\footnotesize
\setlength{\baselineskip}{1.05\baselineskip}
    \hangindent=3em \hangafter=1
    \textbf{*  }: \textbf{N}: total possible cases for each gold length and scenario. \quad \textbf{A}: number of accepted cases. \quad \textbf{A/N}: acceptance ratio. \par
    \hangindent=3em \hangafter=1
    \textit{Note 1}:  The threshold $T$ varies with gold sequence length $|g|$ as shown, with $T = 0.7$ for all $|g| \geq 5$. \par
    \hangindent=3em \hangafter=1
    \textit{Note 2}:  Results are derived from 300 systematically generated test cases exploring three variation scenarios: 1) extending beyond gold by $n \leq 20$ tokens (``over by $n$''), 2) truncating the last $n$ tokens from gold (``under by $n$''), and 3) shifting the token window by $n$ positions whilst maintaining gold length (``shifted by $n$''). 
\end{minipage}

\end{table*}
\renewcommand{\arraystretch}{1}

\renewcommand{\arraystretch}{1.2}

\begin{table*}[!htbp]
\centering
\scriptsize
\caption{FTS Matching results for 20 randomly sampled non-identical gold-pred pairs of aspect and opinion components from 0-shot GPT-4o ICL on the EduRABSA dataset. Columns show FTS scores (FTS), gold extraction length after tokenisation and stopword removal ($g$), threshold values (T), and match outcomes (Match).}
\label{table_fts_example}


\begin{tabularx}{\textwidth}{
>{\raggedright\arraybackslash}p{0.06\linewidth} 
>{\raggedright\arraybackslash}p{0.275\linewidth} 
>{\raggedright\arraybackslash}p{0.275\linewidth} |
>{\raggedleft\arraybackslash}p{0.04\linewidth}  
>{\raggedleft\arraybackslash}p{0.04\linewidth} 
>{\raggedleft\arraybackslash}p{0.05\linewidth}  
>{\raggedleft\arraybackslash}p{0.05\linewidth}  
}

\toprule
\textbf{Type} & \textbf{Gold Extraction (G)} & \textbf{Pred Extraction} & \textbf{FTS} & \textbf{$g$} & \textbf{T} & \textbf{Match} \\ \midrule
Aspect & Beginner Italian I and II & my interest in Italian & 0.29 & 4 & 0.6 & False \\
Aspect & booking Accommodation with Exeter & Accommodation & 0.40 & 4 & 0.6 & False \\
Aspect & desmos & tools like desmos & 0.50 & 1 & 0.5 & True \\
Aspect & BioE 102 class & the class & 0.50 & 3 & 0.6 & False \\
Aspect & opportunities to get involved & opportunities to get involved in societies and meet likeminded people & 0.60 & 3 & 0.6 & True \\
Aspect & his lectures & lectures & 0.67 & 2 & 0.5 & True \\
Aspect & midterm and final & the midterm & 0.67 & 2 & 0.5 & True \\
Aspect & standard of teaching & teaching & 0.67 & 2 & 0.5 & True \\
Aspect & amount of careers help & careers help & 0.80 & 3 & 0.6 & True \\
Aspect & tutorials and in-residence sessions & in-residence sessions & 0.86 & 3 & 0.6 & True \\
\lightrule
Opinion & made the class good / engaging & good & 0.40 & 5 & 0.7 & False \\
Opinion & great & Has a great sense of humor & 0.40 & 1 & 0.5 & False \\
Opinion & amazing place to study & amazing & 0.50 & 3 & 0.6 & False \\
Opinion & hard & hard at the end & 0.50 & 1 & 0.5 & True \\
Opinion & engaging and intuitive & engaging & 0.67 & 2 & 0.5 & True \\
Opinion & Not easy , but not that bad either & not that bad either & 0.67 & 7 & 0.7 & False \\
Opinion & Fairly easy & fairly easy to do well in & 0.67 & 2 & 0.5 & True \\
Opinion & gets easier & only gets easier & 0.80 & 2 & 0.5 & True \\
Opinion & give a really thorough and satisfying answer & really thorough and satisfying answer & 0.86 & 4 & 0.6 & True \\
Opinion & Mostly common sense stuff & common sense stuff & 0.86 & 4 & 0.6 & True \\ \bottomrule
\end{tabularx}
\end{table*}
\renewcommand{\arraystretch}{1}


\subsection{FTS-OBP vs. Exact-match-based Method}\label{subsec_exp_metric_comparison}

To assess the agreement between FTS-OBP and the traditional exact-match-based evaluation method\footnote{For the Exact-match implementation, we used the \href{https://github.com/IsakZhang/ABSA-QUAD/blob/master/eval_utils.py}{ \texttt{compute\_f1\_scores} function} from \citep{zhang2021_asqp} for unit-level matching.} (hereafter ``Exact-match''), we analysed the outputs of 34 model-prompt pairs on the EduRABSA test set across five tasks (OE, AOPE, AOC, ASTE, ASQE). This subsection focuses on the metric-level differences; model performance results are reported in Section~\ref{subsec_results}. The complete data are provided in Table~\ref{table_classical_vs_ftsobp_f1}.

\renewcommand{\arraystretch}{1.2}
\begin{table*}[htbp!]
\centering
\footnotesize

\caption{Correlation and difference statistics of macro-$F_1$ scores computed using FTS-OBP and the exact-match-based method (``Exact-match'') on outputs from 34 model-prompt pairs across five ABSA tasks using the EduRABSA test dataset. All Pearson’s $r$ and Spearman’s $\rho$ values are significant at $p$ \textless 0.001. $\Delta$ = FTS-OBP $-$ Exact-match.}
\label{table_metric_task_correlation}

\begin{tabular}{c | cccc}

\toprule
\textbf{Task} & \textbf{Spearman's $\boldsymbol{\rho}$} & \textbf{Pearson's $\boldsymbol{r}$} & \textbf{Mean $\Delta$} & \textbf{Std. $\Delta$} \\ 

\midrule
OE & 0.744 & 0.840 & 0.242 & 0.072 \\
AOPE & 0.936 & 0.966 & 0.167 & 0.034 \\
AOC & 0.959 & 0.977 & 0.111 & 0.039 \\
ASTE & 0.909 & 0.962 & 0.157 & 0.039 \\
ASQE & 0.947 & 0.982 & 0.102 & 0.039 \\
Task mean & 0.932 & 0.961 & 0.156 & 0.038 \\ 

\bottomrule

\end{tabular}
\end{table*}
\renewcommand{\arraystretch}{1}

\begin{figure*}[ht!]
    \centering
    \footnotesize
    \begin{subfigure}[t]{0.48\textwidth} 
        \includegraphics[width=\textwidth]{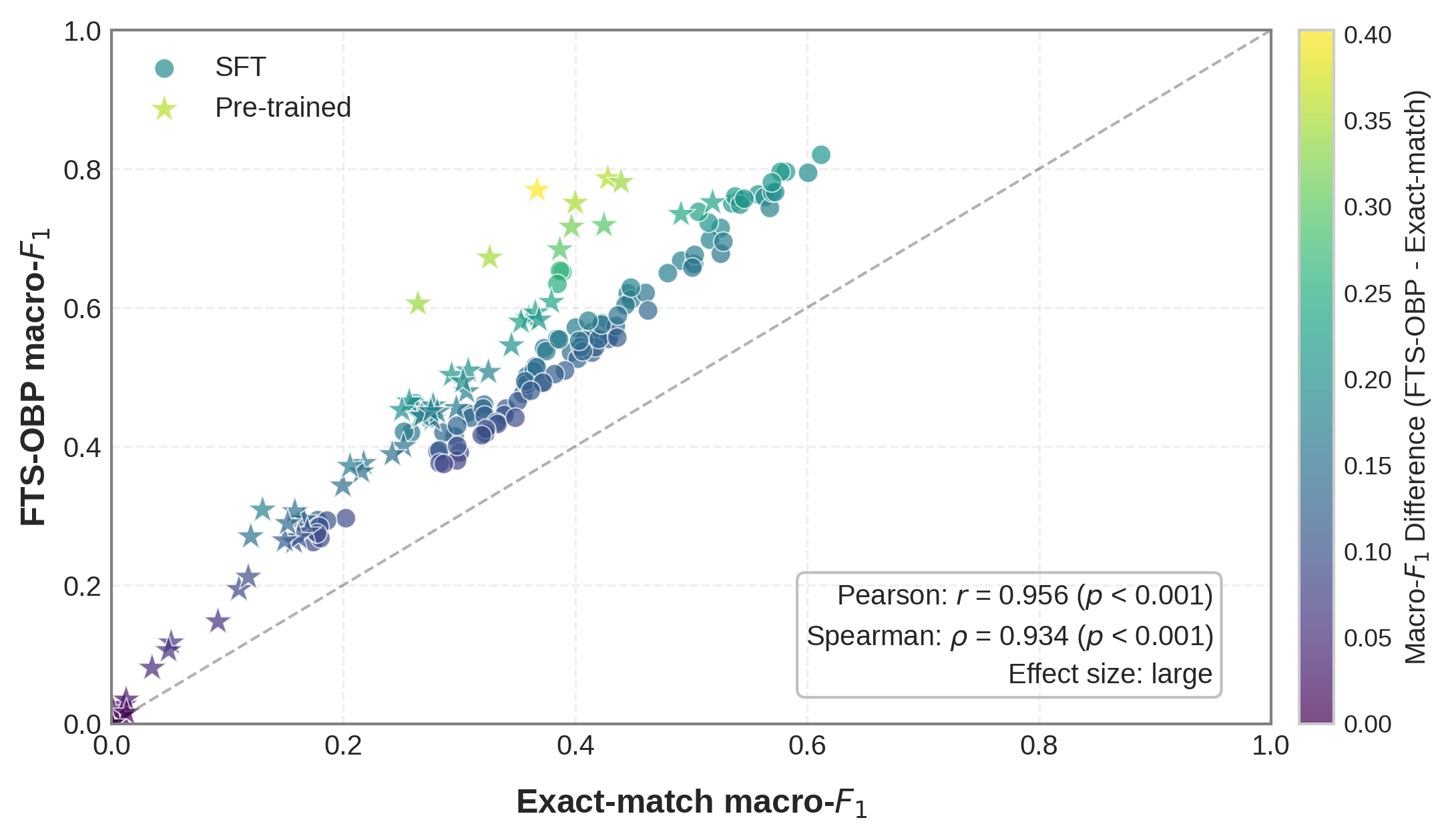}
        \caption{Scatter plot of macro-$F_1$ scores computed using FTS-OBP and Exact-match across 170 data points (34 model-prompt pairs $\times$ 5 tasks), coloured by the magnitude of their difference (FTS-OBP $-$ Exact-match). A strong correlation is observed (Pearson’s $r$ = 0.956; Spearman’s $\rho$ = 0.934; both $p$ \textless 0.001), with larger differences (yellow) primarily occurring at lower to medium performance levels and with pre-trained models.}
        \label{fig_metric_correlation_scatter}
    \end{subfigure}
    \hfill
    \begin{subfigure}[t]{0.5\textwidth}  
        \includegraphics[width=\textwidth]{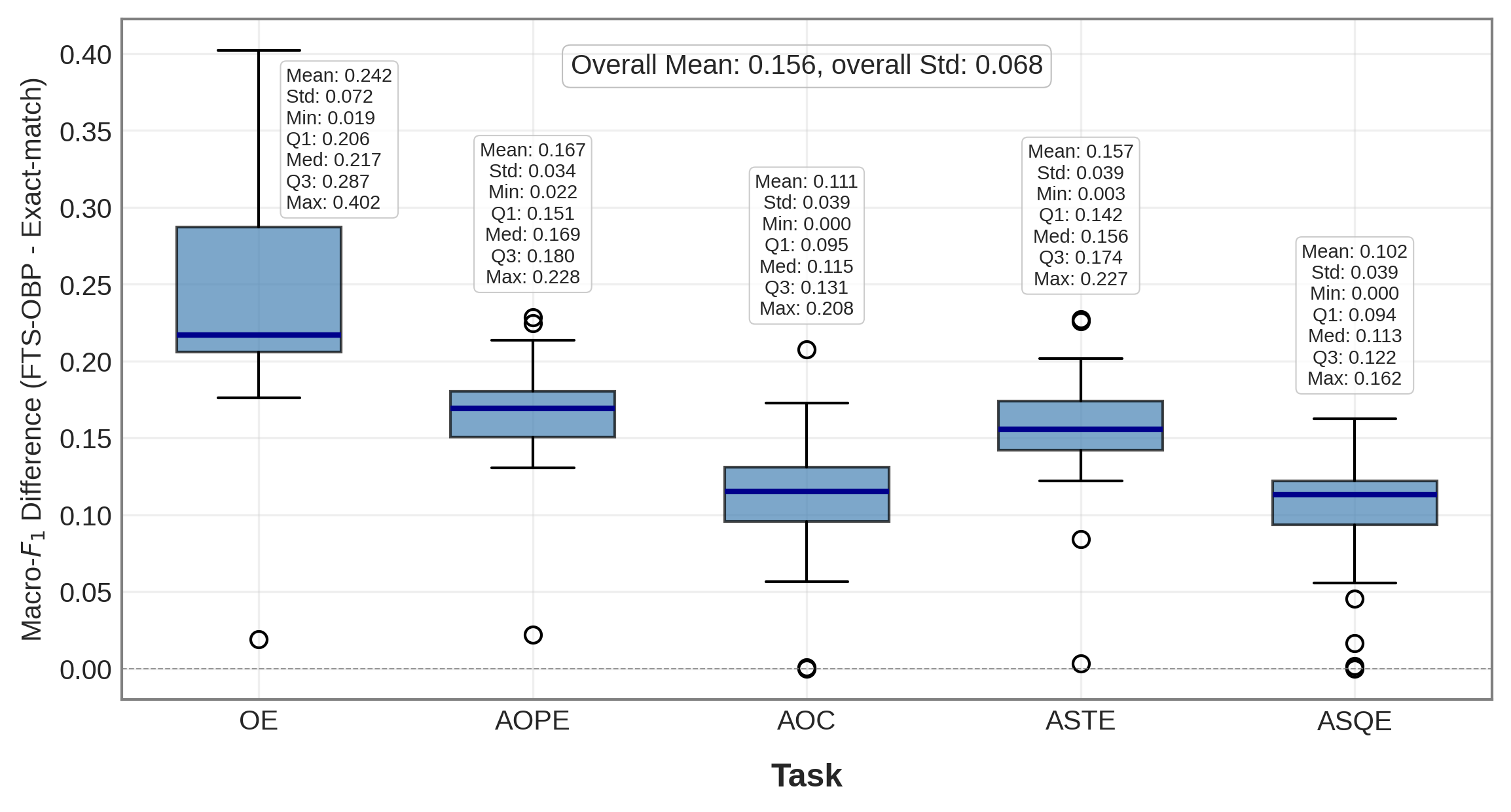}
        \captionsetup{margin=11pt}
        \caption{Distribution of macro-$F_1$ score differences (FTS-OBP $-$ Exact-match) for 34 model-prompt pairs across five ABSA tasks. FTS-OBP consistently yields higher scores, with an overall mean difference of 0.156 (std. = 0.068). The extraction-focused OE task shows the largest difference, whereas the more complex tasks exhibit smaller gaps and reduced variability.}
        \label{fig_metric_correlation_boxplot}
    \end{subfigure}
    
    \caption{Correlation and differences between macro-$F_1$ scores from FTS-OBP and exact-match-based evaluation (``Exact-match'') on outputs from 34 model-prompt pairs across 5 ABSA tasks (OE, AOPE, AOC, ASTE, ASQE) using the EduRABSA test dataset.}
    \label{fig_metric_correlation}
\end{figure*}

Overall, as shown in Table~\ref{table_metric_task_correlation} and Figure~\ref{fig_metric_correlation_scatter}, the two evaluation metrics exhibit a \textbf{strong positive correlation} both within and across tasks (overall Spearman's $\rho$ = 0.934, Pearson's $r$ = 0.956, both $p$ \textless 0.001, $n$ = 170). 

\begin{figure*}[htbp!]
  \centering
  \includegraphics[width=\linewidth]{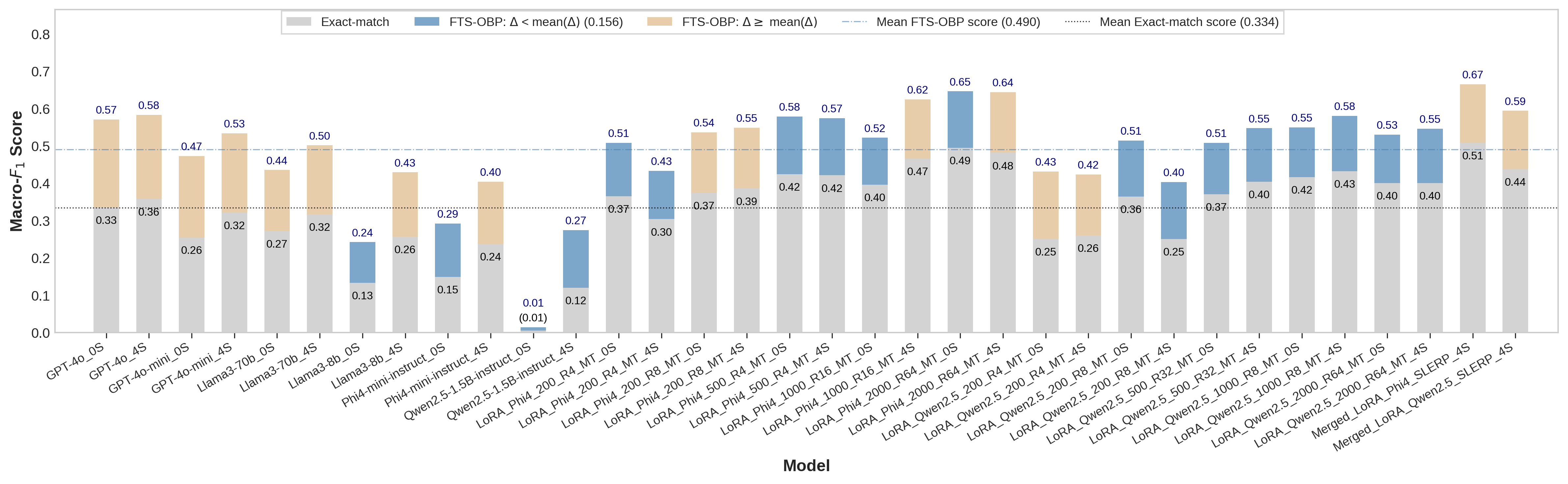}
  \caption{Macro-$F_1$ scores computed with FTS-OBP and exact-match-based method (``Exact-match'') from 34 model-prompt pairs on 5 ABSA subtasks (OE, AOPE, AOC, ASTE, ASQE) using the EduRABSA test dataset. The models include pre-trained, LoRA SFT (``LoRA\_''), and LoRA weight-merged (``Merged\_'')  GLMs and SLMs, with 0-shot (0S) and 4-shot (4S) prompt inputs. The upper section of each bar represents the score difference ($\Delta$ = FTS-OBP $-$ Exact-match, mean($\Delta$) = 0.156).}

  \label{fig4_metric_diff_bargraph}
\end{figure*}

\begin{figure*}[htbp!]
  \centering
  \includegraphics[width=\linewidth]{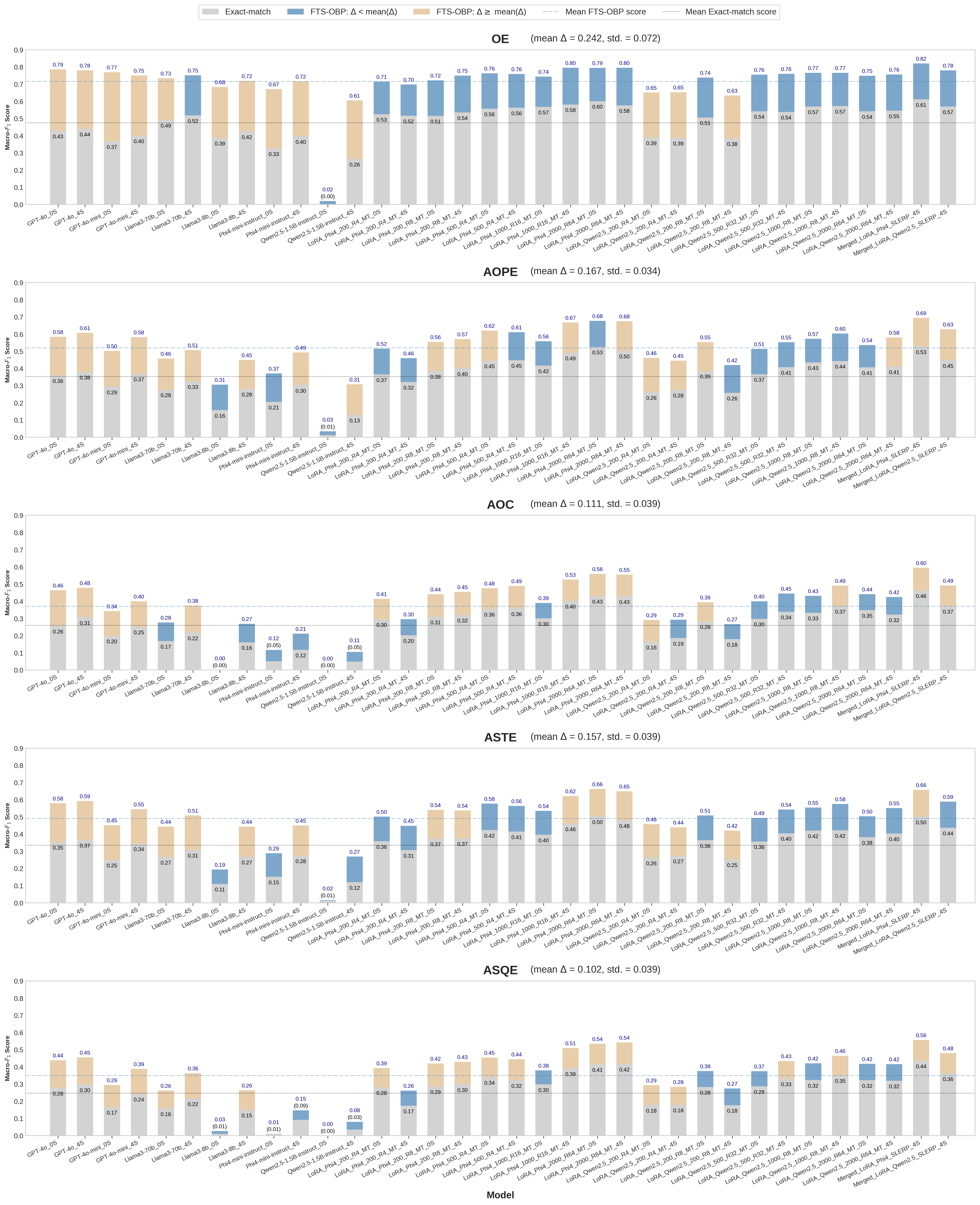}
  \caption{Macro-$F_1$ scores computed with FTS-OBP and exact-match-based method (``Exact-match'') from 34 model-prompt pairs across 5 ABSA subtasks (OE, AOPE, AOC, ASTE, ASQE) using the EduRABSA test dataset. The models include pre-trained GLMs and SLMs, and LoRA SFT (``LoRA\_'') and LoRA weight-merged (``Merged\_'') SLMs, with 0-shot (0S) and 4-shot (4S) prompts. The top section of each bar shows the score difference ($\Delta$ = FTS-OBP $-$ Exact-match).}
  \label{fig5_metric_diff_bargraph_subplots}
\end{figure*}

In addition, FTS-OBP macro-$F_1$ scores are consistently higher than those from Exact-match, with a mean difference of 0.156 (std. = 0.068, 95\% CI = [0.145, 0.166], Cohen's $d$ = 2.282). A paired $t$-test confirmed the significance of this difference ($t$(169) = 29.667, $p$ \textless 0.001). We further examined the score difference $\Delta$ (FTS-OBP $-$ Exact-match) and observed the following patterns:

\begin{enumerate}
    \item \textbf{Metric difference $\Delta$ decreases with task complexity.} Table~\ref{table_metric_task_correlation}, Figure~\ref{fig_metric_correlation_boxplot}, and Figure~\ref{fig5_metric_diff_bargraph_subplots} on per-task metric differences indicate that $\Delta$ decreases in both magnitude and dispersion as task complexity increases. The OE task shows the largest difference due to its single-component, text-extraction nature, whereas AOC and ASQE show the smallest differences, owing to their relatively low performance in both metric systems (Figure~\ref{fig5_metric_diff_bargraph_subplots}).

    \item \textbf{FTS-OBP leniency is more pronounced in extraction tasks with pre-trained models.} As shown in Figures~\ref{fig_metric_correlation_scatter}~and~\ref{fig4_metric_diff_bargraph}, above-average $\Delta$ values occur primarily in pre-trained models (66.67\%, $N$ = 8/12) and low-rank LoRA models trained on small datasets (18.18\%, 4/22). Figure~\ref{fig5_metric_diff_bargraph_subplots} further indicates that this pattern is most evident in text extraction tasks (OE, AOPE). This may be attributed to boundary decision variability implicitly learned from their pre-training data, particularly when evaluated on challenging, novel, and/or highly domain-specific datasets such as EduRABSA \citep{hua2025edurabsa}.
\end{enumerate}

These patterns and relationships are consistent with analyses conducted on the two benchmark datasets, as shown in Appendix Figures~\ref{fig_semeval_metric_comparison_bar_graph}--\ref{fig_semeval_metric_diff_boxplot}. In particular, Figure~\ref{fig_semeval_metric_diff_boxplot} supports point 1 above: the metric differences diminish for more challenging tasks, such as ASQE on the ACOS Laptop dataset, where scores on both metrics are lower than on ASQP Rest16. Nevertheless, FTS-OBP leniency continued to benefit pre-trained models more strongly as shown in Figures~\ref{fig_semeval_metric_comparison_bar_graph}--\ref{fig_semeval_scatterplot}.

These results reflect the design intent of FTS-OBP: to tolerate minor boundary deviations in aspect and opinion extraction that would otherwise be rejected by Exact-match, while maintaining strict exact-matching criteria for the classification components within each evaluation unit. Detailed result analyses in Figure~\ref{fig_asp_opn_heatmap} further support this, showing that: 1) across tasks and models outputs, 87.33\%--96.53\% aspect pairs and 81.41\%--88.96\% opinion pairs accepted by FTS-OBP were exact matches; 2) among the accepted pairs, the boundary variations in the ``over'' and ``under'' cases only took up at most 2.22\% and 2.28\% for aspects, and 9.62\% and 10.69\% for opinion pairs, respectively; and 3) boundary window shift cases were rarely accepted, accounting for only up to 0.12\% for aspect and 0.23\%  for opinion among the matched pairs. The same analyses of the ASQE task outputs on the benchmark datasets are presented in Appendix~\ref{appendix_F_semeval_results} and are consistent with the results above.

\clearpage

\renewcommand{\arraystretch}{1.25}
\begin{table}[htbp!]

\scriptsize
\centering

\captionsetup{skip=0.5em}  

\caption{Macro-$F_1$ scores obtained using traditional exact-matching evaluation methods and FTS-OBP across five tasks in multitask testing on the EduRABSA dataset (300 entries per task). Tasks: OE, AOPE, AOC, ASTE, and ASQE. Task mean is the average task macro-$F_1$ value.}
\label{table_classical_vs_ftsobp_f1}

\begin{tabularx}{\textwidth}{
>{\raggedright\arraybackslash}p{0.23\linewidth} |
>{\centering\arraybackslash}p{0.03\linewidth}
>{\centering\arraybackslash}p{0.03\linewidth}
>{\centering\arraybackslash}p{0.03\linewidth}
>{\centering\arraybackslash}p{0.03\linewidth}
>{\centering\arraybackslash}p{0.03\linewidth}
>{\centering\arraybackslash}X |
>{\centering\arraybackslash}p{0.05\linewidth}
>{\centering\arraybackslash}p{0.03\linewidth}
>{\centering\arraybackslash}p{0.03\linewidth}
>{\centering\arraybackslash}p{0.03\linewidth}
>{\centering\arraybackslash}p{0.03\linewidth}
>{\centering\arraybackslash}X
}

\toprule
\textbf{} & \multicolumn{6}{c|}{\textbf{Exact Match macro-$F_1$}} & \multicolumn{6}{c}{\textbf{FTS-OBP macro-$F_1$}} \\ [0.6em]
\textbf{Model} & \textbf{OE} & \textbf{AOPE} & \textbf{AOC} & \textbf{ASTE} & \textbf{ASQE} & \textbf{Task mean} & \textbf{OE} & \textbf{AOPE} & \textbf{AOC} & \textbf{ASTE} & \textbf{ASQE} & \textbf{Task mean} \\

 \midrule

GPT-4o\_0S & 0.43 & 0.36 & 0.26 & 0.35 & 0.28 & 0.33 & 0.79 & 0.58 & 0.46 & 0.58 & 0.44 & 0.57 \\
GPT-4o\_4S & 0.44 & 0.38 & 0.31 & 0.37 & 0.30 & 0.36 & 0.78 & 0.61 & 0.48 & 0.59 & 0.45 & 0.58 \\
GPT-4o-mini\_0S & 0.37 & 0.29 & 0.20 & 0.25 & 0.17 & 0.26 & 0.77 & 0.50 & 0.34 & 0.45 & 0.29 & 0.47 \\
GPT-4o-mini\_4S & 0.40 & 0.37 & 0.25 & 0.34 & 0.24 & 0.32 & 0.75 & 0.58 & 0.40 & 0.55 & 0.39 & 0.53 \\
Llama3-70b\_0S & 0.49 & 0.28 & 0.17 & 0.27 & 0.16 & 0.27 & 0.73 & 0.46 & 0.28 & 0.44 & 0.26 & 0.44 \\
Llama3-70b\_4S & 0.52 & 0.33 & 0.22 & 0.31 & 0.22 & 0.32 & 0.75 & 0.51 & 0.38 & 0.51 & 0.36 & 0.50 \\
Llama3-8b\_0S & 0.39 & 0.16 & 0.00 & 0.11 & 0.01 & 0.13 & 0.68 & 0.31 & 0.00 & 0.19 & 0.03 & 0.24 \\
Llama3-8b\_4S & 0.42 & 0.28 & 0.16 & 0.27 & 0.15 & 0.26 & 0.72 & 0.45 & 0.27 & 0.44 & 0.26 & 0.43 \\
Phi4-mini-instruct\_0S & 0.33 & 0.21 & 0.05 & 0.15 & 0.01 & 0.15 & 0.67 & 0.37 & 0.12 & 0.29 & 0.01 & 0.29 \\
Phi4-mini-instruct\_4S & 0.40 & 0.30 & 0.12 & 0.28 & 0.09 & 0.24 & 0.72 & 0.49 & 0.21 & 0.45 & 0.15 & 0.40 \\
Qwen2.5-1.5B-instruct\_0S & 0.00 & 0.01 & 0.00 & 0.01 & 0.00 & 0.01 & 0.02 & 0.03 & 0.00 & 0.02 & 0.00 & 0.01 \\
Qwen2.5-1.5B-instruct\_4S & 0.26 & 0.13 & 0.05 & 0.12 & 0.03 & 0.12 & 0.61 & 0.31 & 0.11 & 0.27 & 0.08 & 0.27 \\
\lightrule
LoRA\_Phi4\_200\_R4\_MT\_0S & 0.53 & 0.37 & 0.30 & 0.36 & 0.28 & 0.37 & 0.71 & 0.52 & 0.41 & 0.50 & 0.39 & 0.51 \\
LoRA\_Phi4\_200\_R4\_MT\_4S & 0.52 & 0.32 & 0.20 & 0.31 & 0.17 & 0.30 & 0.70 & 0.46 & 0.30 & 0.45 & 0.26 & 0.43 \\
LoRA\_Phi4\_200\_R8\_MT\_0S & 0.51 & 0.38 & 0.31 & 0.37 & 0.29 & 0.37 & 0.72 & 0.56 & 0.44 & 0.54 & 0.42 & 0.54 \\
LoRA\_Phi4\_200\_R8\_MT\_4S & 0.54 & 0.40 & 0.32 & 0.37 & 0.30 & 0.39 & 0.75 & 0.57 & 0.45 & 0.54 & 0.43 & 0.55 \\
LoRA\_Phi4\_500\_R4\_MT\_0S & 0.56 & 0.45 & 0.36 & 0.42 & 0.34 & 0.42 & 0.76 & 0.62 & 0.48 & 0.58 & 0.45 & 0.58 \\
LoRA\_Phi4\_500\_R4\_MT\_4S & 0.56 & 0.45 & 0.36 & 0.41 & 0.32 & 0.42 & 0.76 & 0.61 & 0.49 & 0.56 & 0.44 & 0.57 \\
LoRA\_Phi4\_1000\_R16\_MT\_0S & 0.57 & 0.42 & 0.30 & 0.40 & 0.30 & 0.40 & 0.74 & 0.56 & 0.39 & 0.54 & 0.38 & 0.52 \\
LoRA\_Phi4\_1000\_R16\_MT\_4S & 0.58 & 0.49 & 0.40 & 0.46 & 0.39 & 0.47 & 0.80 & 0.67 & 0.53 & 0.62 & 0.51 & 0.62 \\
LoRA\_Phi4\_2000\_R64\_MT\_0S & 0.60 & 0.53 & 0.43 & 0.50 & 0.41 & 0.49 & 0.79 & 0.68 & 0.56 & 0.66 & 0.54 & 0.65 \\
LoRA\_Phi4\_2000\_R64\_MT\_4S & 0.58 & 0.50 & 0.43 & 0.48 & 0.42 & 0.48 & 0.80 & 0.68 & 0.55 & 0.65 & 0.54 & 0.64 \\
\lightrule
LoRA\_Qwen2.5\_200\_R4\_MT\_0S & 0.39 & 0.26 & 0.16 & 0.26 & 0.18 & 0.25 & 0.65 & 0.46 & 0.29 & 0.46 & 0.29 & 0.43 \\
LoRA\_Qwen2.5\_200\_R4\_MT\_4S & 0.39 & 0.28 & 0.19 & 0.27 & 0.18 & 0.26 & 0.65 & 0.45 & 0.29 & 0.44 & 0.28 & 0.42 \\
LoRA\_Qwen2.5\_200\_R8\_MT\_0S & 0.51 & 0.39 & 0.28 & 0.36 & 0.28 & 0.36 & 0.74 & 0.55 & 0.39 & 0.51 & 0.38 & 0.51 \\
LoRA\_Qwen2.5\_200\_R8\_MT\_4S & 0.38 & 0.26 & 0.18 & 0.25 & 0.18 & 0.25 & 0.63 & 0.42 & 0.27 & 0.42 & 0.27 & 0.40 \\
LoRA\_Qwen2.5\_500\_R32\_MT\_0S & 0.54 & 0.37 & 0.30 & 0.36 & 0.29 & 0.37 & 0.76 & 0.51 & 0.40 & 0.49 & 0.37 & 0.51 \\
LoRA\_Qwen2.5\_500\_R32\_MT\_4S & 0.54 & 0.41 & 0.34 & 0.40 & 0.33 & 0.40 & 0.76 & 0.55 & 0.45 & 0.54 & 0.43 & 0.55 \\
LoRA\_Qwen2.5\_1000\_R8\_MT\_0S & 0.57 & 0.43 & 0.33 & 0.42 & 0.32 & 0.42 & 0.77 & 0.57 & 0.43 & 0.55 & 0.42 & 0.55 \\
LoRA\_Qwen2.5\_1000\_R8\_MT\_4S & 0.57 & 0.44 & 0.37 & 0.42 & 0.35 & 0.43 & 0.77 & 0.60 & 0.49 & 0.58 & 0.46 & 0.58 \\
LoRA\_Qwen2.5\_2000\_R64\_MT\_0S & 0.54 & 0.41 & 0.35 & 0.38 & 0.32 & 0.40 & 0.75 & 0.54 & 0.44 & 0.50 & 0.42 & 0.53 \\
LoRA\_Qwen2.5\_2000\_R64\_MT\_4S & 0.55 & 0.41 & 0.32 & 0.40 & 0.32 & 0.40 & 0.76 & 0.58 & 0.42 & 0.55 & 0.42 & 0.55 \\
\lightrule
Merged\_LoRA\_Phi4\_SLERP\_4S & 0.61 & 0.53 & 0.46 & 0.50 & 0.44 & 0.51 & 0.82 & 0.69 & 0.60 & 0.66 & 0.56 & 0.67 \\
Merged\_LoRA\_Qwen2.5\_SLERP\_4S & 0.57 & 0.45 & 0.37 & 0.44 & 0.36 & 0.44 & 0.78 & 0.63 & 0.49 & 0.59 & 0.48 & 0.59 \\

\bottomrule
\end{tabularx}
\end{table}


\newpage
\clearpage \newpage
\section{Detailed Experiment Results for the EduRABSA Dataset}\label{appendix_E_detailed_results}

All results reported in this appendix section were obtained using the EduRABSA test dataset, with 300 examples per task. Unless otherwise specified, all models were evaluated and/or fine-tuned in a multi-task setting with five tasks: OE, AOPE, AOC, ASTE, and ASQE. \\

\begin{figure*}[hb!]
  \centering
  \includegraphics[width=0.85\textwidth]{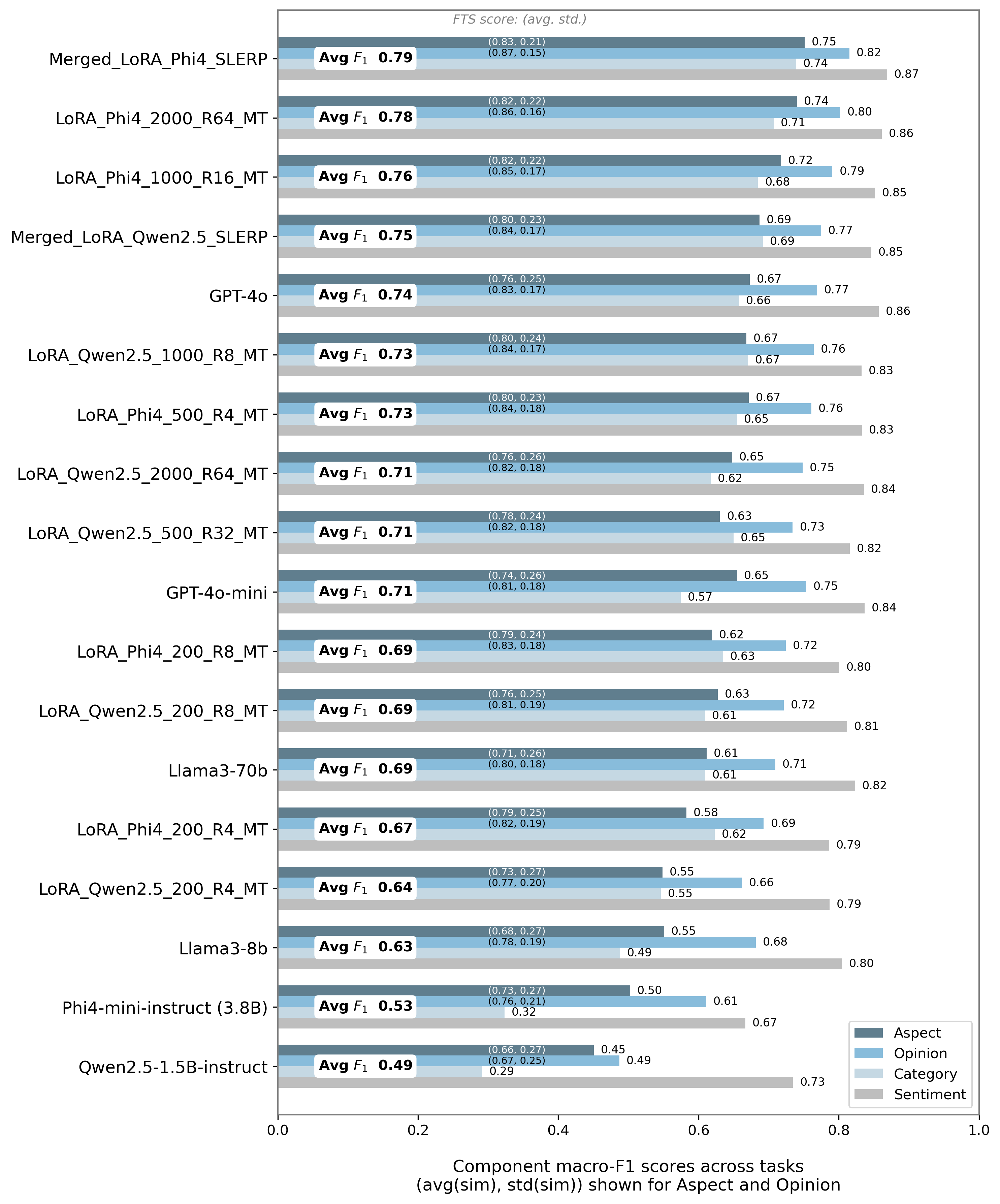}
  
  \caption{Macro-$F_1$ scores per component across entries and tasks (OE, AOPE, AOC, ASTE, ASQE) on the EduRABSA dataset (300 test examples per task). Models include pre-trained GLMs and SLMs, and LoRA-SFT and LoRA weight-merged SLMs, and are ordered by descending cross-component average scores. ``Avg. $F_1$'' represents the mean value of a model's component macro-$F_1$ scores. The pair of numbers shown within each aspect or opinion bar represents the mean and standard deviation of that component’s FTS (similarity) score.}

  \label{fig_component_metrics}
\end{figure*}
\newpage

\renewcommand{\arraystretch}{1.05}
\begin{table}[h!]

\footnotesize
\centering

\captionsetup{skip=0.5em}  

\caption{Total ground-truth (gold) and model output (pred) \textbf{aspect and opinion pairs} across tasks in multitask testing on the EduRABSA dataset (300 entries per task), and the percentage accepted as a match by the FTS-OBP method. The tasks include OE, AOPE, AOC, ASTE, and ASQE. The two merged models are compared with their sources (the 2 best-performing LoRA\_phi4 and LoRA\_Qwen models).}

\label{table_matching_result_asp_opn}

\begin{tabularx}{\textwidth}{
>{\raggedright\arraybackslash}p{0.3\linewidth} |
>{\centering\arraybackslash}p{0.07\linewidth} 
>{\centering\arraybackslash}p{0.08\linewidth} 
>{\centering\arraybackslash}p{0.08\linewidth} 
>{\centering\arraybackslash}p{0.1\linewidth} |
>{\raggedleft\arraybackslash}p{0.07\linewidth} 
>{\centering\arraybackslash}X 
}

\toprule
\textbf{} & \multicolumn{4}{c|}{\textbf{Aspect Pairs}} & \multicolumn{2}{c}{\textbf{Opinion Pairs}} \\ 
\lightrule
\textbf{Model} & \textbf{Total} & \textbf{Match \%} & \textbf{Implicit Total} & \textbf{Implicit Match \%} & \textbf{Total} & \textbf{Match \%} \\
\midrule

\textbf{GPT-4o\_4S} & 4776 & \textbf{67.06} & 434 & 54.84 & 5942 & \textbf{72.79} \\
GPT-4o-mini\_4S & 4775 & 63.83 & 439 & 65.60 & 5932 & 69.07 \\
\textbf{Llama3-70b\_4S} & 4925 & 61.36 & 454 & \textbf{70.04} & 6084 & 68.82 \\
Llama3-8b\_4S & 4696 & 55.64 & 422 & 55.21 & 5855 & 64.90 \\
Phi4-mini-instruct (3.8B)\_4S & 3782 & 60.18 & 296 & 0.00 & 4940 & 62.73 \\
Qwen2.5-1.5B-instruct\_4S & 3526 & 47.22 & 297 & 0.00 & 4577 & 45.49 \\
\lightrule
LoRA\_Phi4\_200\_R4\_MT\_0S & 4053 & 65.36 & 396 & 0.25 & 5136 & 72.14 \\
LoRA\_Phi4\_200\_R8\_MT\_0S & 4173 & 68.10 & 382 & 17.80 & 5272 & 73.58 \\
\textbf{LoRA\_Phi4\_500\_R4\_MT\_0S} & 4471 & 70.25 & 398 & \textbf{58.54} & 5622 & 73.78 \\
LoRA\_Phi4\_1000\_R16\_MT\_4S & 4380 & 74.11 & 413 & 55.45 & 5493 & 76.44 \\
\textbf{LoRA\_Phi4\_2000\_R64\_MT\_4S} & 4702 & \textbf{75.14} & 425 & 47.29 & 5865 & \textbf{77.19} \\
\lightrule
LoRA\_Qwen2.5\_200\_R4\_MT\_0S & 4030 & 55.29 & 361 & 36.01 & 4996 & 64.35 \\
LoRA\_Qwen2.5\_200\_R8\_MT\_0S & 4352 & 63.99 & 393 & 47.84 & 5486 & 69.65 \\
LoRA\_Qwen2.5\_500\_R32\_MT\_4S & 4450 & 65.06 & 398 & 32.91 & 5566 & 71.60 \\
\textbf{LoRA\_Qwen2.5\_1000\_R8\_MT\_4S} & 4524 & \textbf{69.19} & 405 & 53.33 & 5677 & \textbf{74.95} \\
\textbf{LoRA\_Qwen2.5\_2000\_R64\_MT\_4S} & 4417 & 63.66 & 389 & \textbf{68.12} & 5562 & 71.75 \\
\lightrule
\textbf{Merged\_Phi4\_SLERP\_4S} & 4659 & \textbf{76.52} & 423 & \textbf{57.68} & 5813 & \textbf{78.38} \\
\textbf{Merged\_Qwen2.5\_SLERP\_4S} & 4523 & \textbf{70.88} & 401 & 62.59 & 5687 & 74.64 \\ 

\bottomrule
\end{tabularx}
\end{table}



\renewcommand{\arraystretch}{1.05}
\begin{table}[h!]

\footnotesize
\centering

\captionsetup{skip=0.5em}  

\caption{Total counts and percentage of matched ground-truth (gold) and model output (pred) \textbf{category and sentiment pairs} across tasks in multitask testing on the EduRABSA dataset (300 entries per task). The tasks include AOC, ASTE, and ASQE. The two merged models are compared with their sources (the 2 best-performing LoRA\_phi4 and LoRA\_Qwen models, respectively).}

\label{table_matching_result_cat_sent}

\begin{tabularx}{\textwidth}{
>{\raggedright\arraybackslash}p{0.3\linewidth} |
>{\centering\arraybackslash}p{0.04\linewidth} 
>{\centering\arraybackslash}p{0.08\linewidth} |
>{\centering\arraybackslash}p{0.04\linewidth} 
>{\centering\arraybackslash}p{0.08\linewidth} |
>{\centering\arraybackslash}p{0.04\linewidth} 
>{\centering\arraybackslash}p{0.08\linewidth} |
>{\centering\arraybackslash}p{0.04\linewidth} 
>{\centering\arraybackslash}p{0.08\linewidth} 
}

\toprule
\textbf{} & \multicolumn{4}{c|}{\textbf{Category}} & \multicolumn{4}{c}{\textbf{Sentiment}} \\ 
\lightrule
\textbf{} & \multicolumn{2}{c|}{\textbf{AOC}} & \multicolumn{2}{c|}{\textbf{ASQE}} & \multicolumn{2}{c|}{\textbf{ASTE}} & \multicolumn{2}{c}{\textbf{ASQE}} \\ 

\textbf{Model} & \textbf{Total} & \textbf{Match \%} & \textbf{Total} & \textbf{Match \%} & \textbf{Total} & \textbf{Match \%} & \textbf{Total} & \textbf{Match \%} \\

\midrule
\textbf{GPT-4o\_4S} & 1198 & \textbf{71.54} & 1196 & \textbf{69.65} & 1186 & \textbf{93.93} & 1196 & \textbf{93.31} \\
GPT-4o-mini\_4S & 1190 & 61.43 & 1189 & 61.31 & 1207 & 91.96 & 1189 & 91.76 \\
Llama3-70b\_4S & 1230 & 66.34 & 1228 & 65.96 & 1225 & 93.22 & 1228 & 92.67 \\
Llama3-8b\_4S & 1189 & 54.58 & 1157 & 56.01 & 1168 & 90.84 & 1157 & 91.01 \\
Phi4-mini-instruct (3.8B)\_4S & 909 & 47.30 & 760 & 45.26 & 1023 & 90.62 & 760 & 89.47 \\
Qwen2.5-1.5B-instruct\_4S & 854 & 37.94 & 890 & 34.61 & 926 & 88.98 & 890 & 89.21 \\
\lightrule
LoRA\_Phi4\_200\_R4\_MT\_0S & 992 & 72.28 & 1015 & 72.22 & 1043 & 92.04 & 1015 & 91.33 \\
LoRA\_Phi4\_200\_R8\_MT\_0S & 1020 & 73.04 & 1021 & 70.13 & 1078 & 92.30 & 1021 & 91.58 \\
LoRA\_Phi4\_500\_R4\_MT\_0S & 1109 & 71.42 & 1121 & 71.19 & 1127 & 92.64 & 1121 & 91.61 \\
LoRA\_Phi4\_1000\_R16\_MT\_4S & 1089 & 74.01 & 1085 & 73.64 & 1088 & \textbf{93.20} & 1085 & 93.00 \\
\textbf{LoRA\_Phi4\_2000\_R64\_MT\_4S} & 1186 & \textbf{74.37} & 1181 & \textbf{75.11} & 1160 & 93.02 & 1181 & \textbf{93.23} \\
\lightrule
LoRA\_Qwen2.5\_200\_R4\_MT\_0S & 975 & 61.54 & 1021 & 60.43 & 1042 & 90.21 & 1021 & 90.11 \\
LoRA\_Qwen2.5\_200\_R8\_MT\_0S & 1083 & 67.59 & 1095 & 67.49 & 1085 & 91.34 & 1095 & 91.14 \\
LoRA\_Qwen2.5\_500\_R32\_MT\_4S & 1089 & 70.62 & 1100 & 71.55 & 1117 & 91.94 & 1100 & 90.55 \\
\textbf{LoRA\_Qwen2.5\_1000\_R8\_MT\_4S} & 1129 & \textbf{75.02} & 1109 & \textbf{74.66} & 1131 & \textbf{93.19} & 1109 & \textbf{92.52} \\
LoRA\_Qwen2.5\_2000\_R64\_MT\_4S & 1077 & 68.15 & 1109 & 66.73 & 1118 & 92.58 & 1109 & 91.61 \\
\lightrule
\textbf{Merged\_Phi4\_SLERP\_4S} & 1166 & \textbf{78.99} & 1157 & \textbf{78.57} & 1167 & 93.06 & 1157 & \textbf{93.43} \\
\textbf{Merged\_Qwen2.5\_SLERP\_4S} & 1113 & 74.75 & 1133 & \textbf{74.76} & 1142 & 92.99 & 1133 & \textbf{92.67} \\ 

\bottomrule

\end{tabularx}
\end{table}


\begin{figure}[htbp!]
    \centering   
    
    \begin{subfigure}[c]{0.8\textwidth}
        \centering
        \includegraphics[height=0.41\textheight]{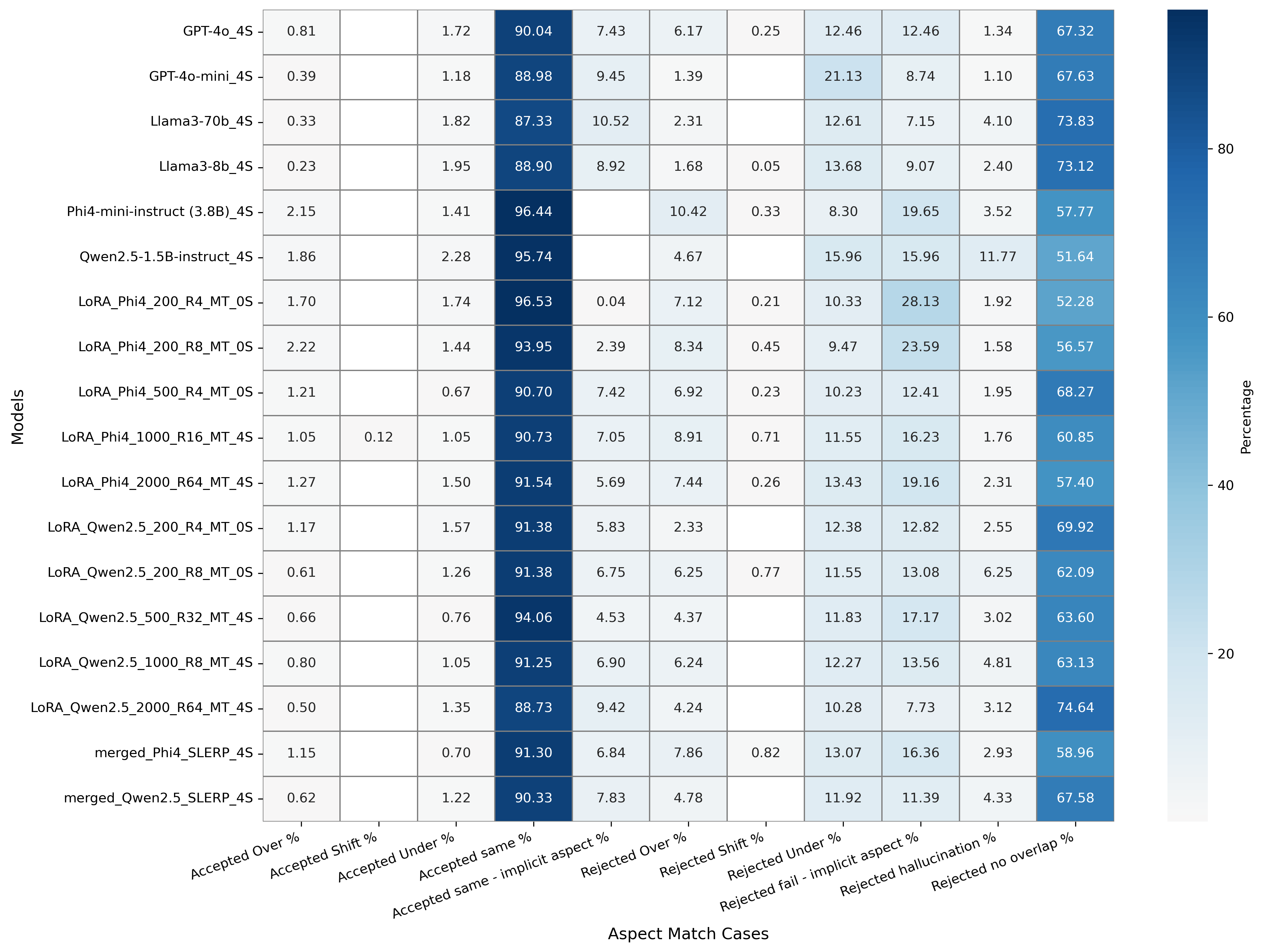}
        \caption{Percentage of matched aspect pairs among all accepted and rejected}
        \label{fig_asp_heatmap}
    \end{subfigure}

    \begin{subfigure}[c]{0.8\textwidth}
        \centering
        \includegraphics[height=0.41\textheight]{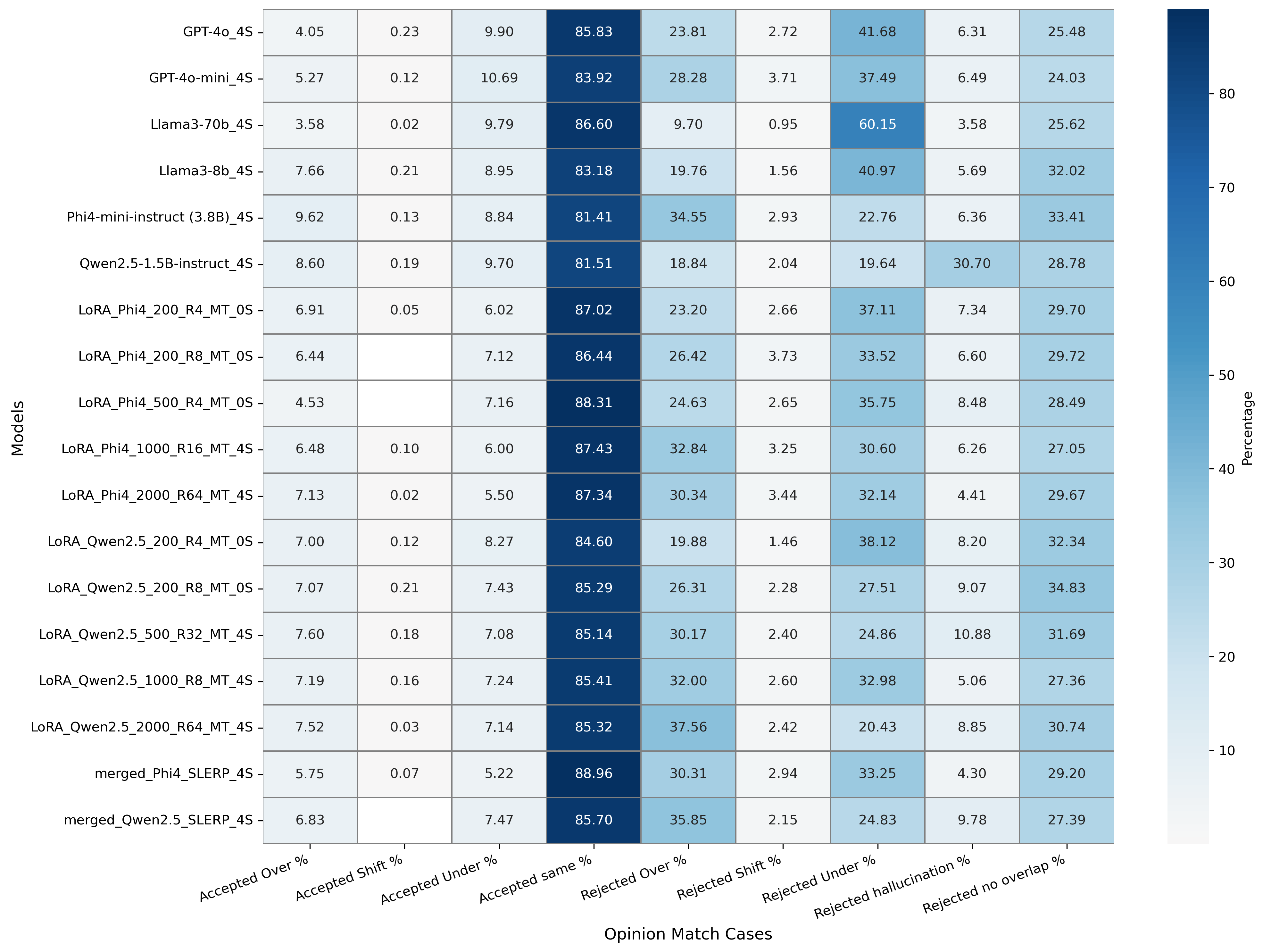}
        \caption{Percentage of matched opinion pairs among all accepted and rejected}
        \label{fig_opn_heatmap}
    \end{subfigure}

    \caption{Percentage of ground-truth (gold) and model output (pred) aspect (a) and opinion (b) pairs in each match case, calculated over all pairs either accepted as a match or rejected by the FTS-OBP evaluation method, across related tasks using the EduRAbsa test dataset ($N$ = 300/task). The match cases describe how pred differs from gold: 1) outside the original input text (``\textbf{hallucination}''), 2) extending beyond gold (``\textbf{over}''), 3) is a substring of gold (``\textbf{under}''), 4) partially overlapping with the gold and is not a substring of it (``\textbf{shift}''), and 5) \textbf{no overlap}. Cases 2)--4) were determined after removing the stopwords specified in Sec.~\ref{subsection_exp_metric}.}
    \label{fig_asp_opn_heatmap}
\end{figure}

\begin{figure}[htbp!]
    \centering   
    \footnotesize
    
    \begin{subfigure}[c]{\textwidth}
        \centering
        \includegraphics[width=\textwidth]{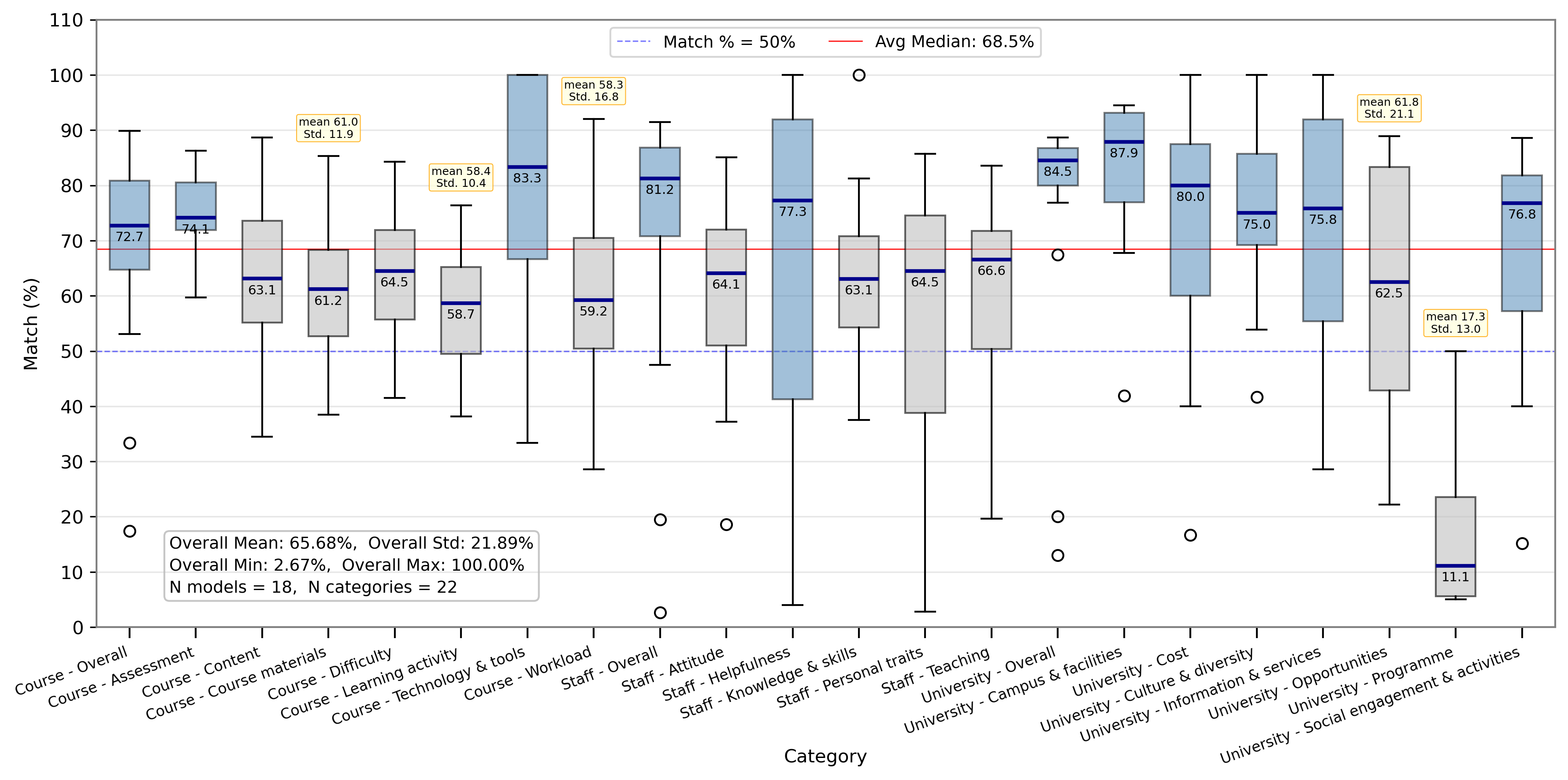}
        \caption{Distribution of match percentages per category label across models on the EduRABSA dataset for the ASQE task. The models involved are identical to those in sub-figure (b). Grey boxes indicate categories with below-average median values. Mean and standard deviation (std.) values are additionally provided for the five categories with the lowest median performance. }
        \label{fig_cat_boxplot}
    \end{subfigure}

    \bigskip

    \begin{subfigure}[c]{\textwidth}
        \centering
        \includegraphics[width=\textwidth]{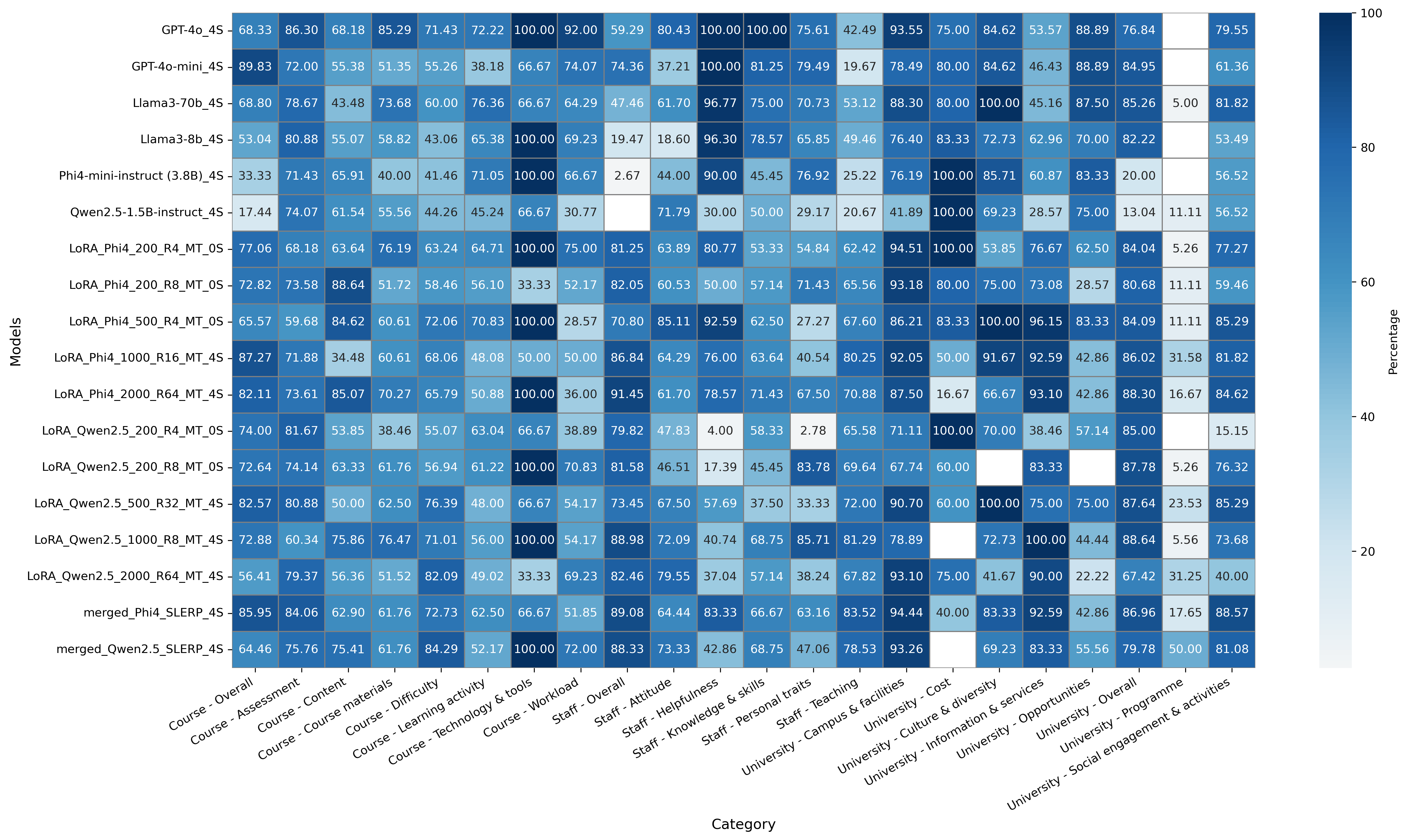}
        \caption{Percentage of matched pairs per category label and model on the EduRABSA dataset for the ASQE task.}
        \label{fig_cat_heatmap}
    \end{subfigure}

    \caption{Overview of category label matching between ground-truth (gold) and model output (pred) pairs for the EduRABSA dataset (N = 300) on the ASQE task.  Subfigure (a) shows the distribution of match percentages per category across models, while subfigure (b) presents a heatmap of matches by category and model.}
        
    \label{fig_cat_boxplot_heatmap}
\end{figure}

\newpage
\clearpage \newpage
\section{Results on the ASQE Task Using the ACOS Laptop and ASQP Rest16 Datasets}\label{appendix_F_semeval_results}

The figures in this section present the experiment results on the ASQE task using the ASQP Rest16 and ACOS Laptop benchmark datasets (300 entries per test set). 

The dataset split and the settings for fine-tuning, weight merging, and inference were identical to those described in Section~\ref{subsec5_1_experiment_setup} for the EduRABSA dataset, with three exceptions: 1) all models were trained and evaluated in a single-task setting; 2) the ASQP Rest16 training set was capped at 1000 entries due to the source dataset size limit; and 3) The Llama3-70B and Llama3-8B models were excluded due to API endpoint deprecation, and no replacements were introduced to maintain consistency across datasets.

As with EduRABSA, the models include pre-trained GLMs and SLMs, as well as LoRA SFT (``LoRA\_'') and LoRA weight-merged (``Merged\_'') SLMs, with 0-shot (0S) and 4-shot (4S) prompt inputs. The LoRA-SFT models were fine-tuned and evaluated using identical prompts, and the two weight-merged LoRA models were derived from the best two (per each base model) 4-shot LoRA checkpoints.

\begin{figure*}[htbp!]
  \centering
  \includegraphics[width=\linewidth]{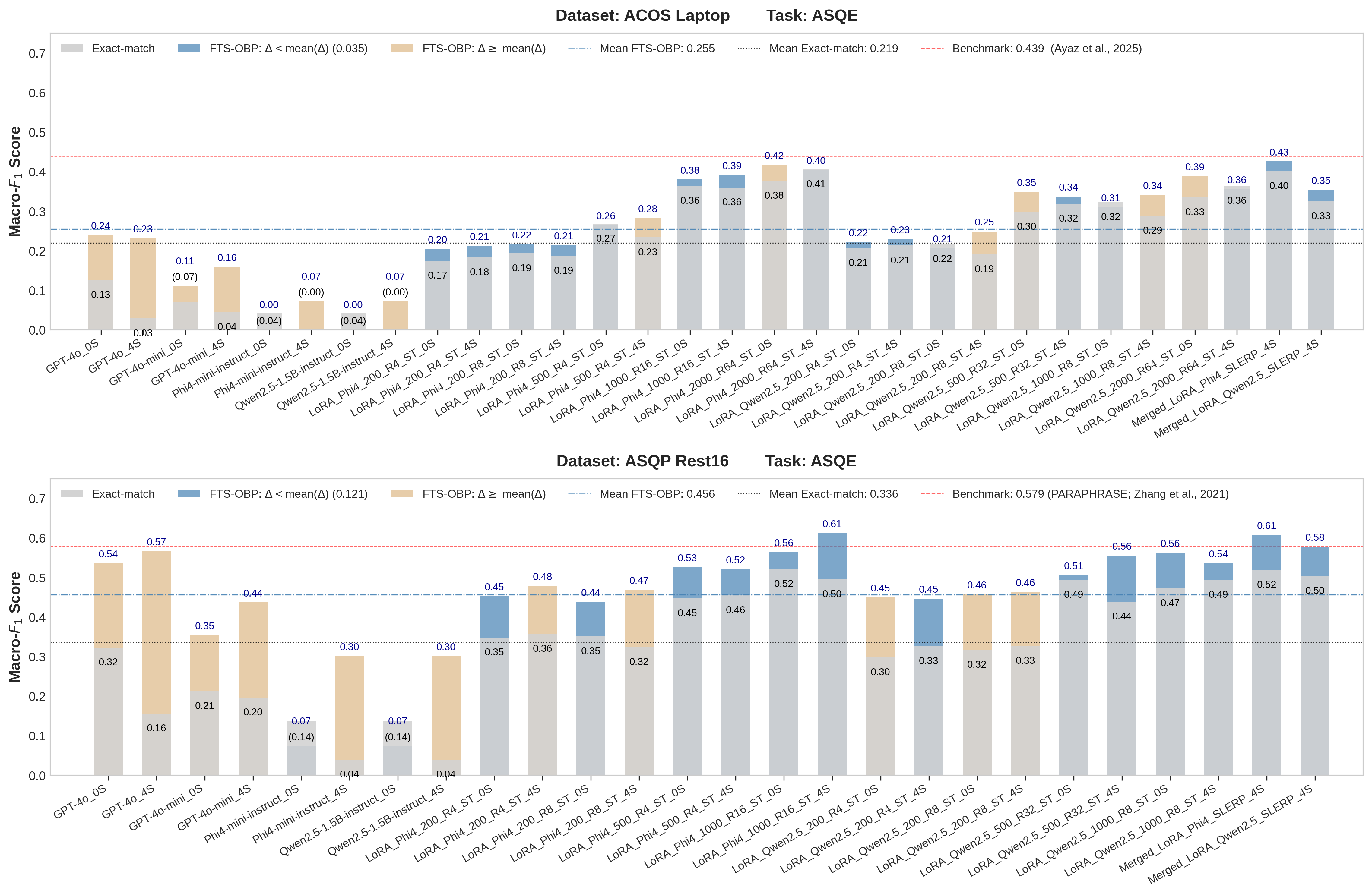}
  
  \caption{Macro-$F_1$ scores computed with FTS-OBP and exact-match-based method (``Exact-match'') on the ASQE task with two benchmark datasets: ACOS Laptop ($N$ = 30 model-prompt pairs) and ASQP Rest16 ($N$ = 26 model-prompt pairs). The upper section of each bar represents the score difference ($\Delta$ = FTS-OBP $-$ Exact-match). The benchmark scores are the highest ASQE results (under the exact-match criterion) reported among existing benchmark models by Ayaz et al. (2025)~\citep{ref_2025_t5absa} for ACOS Laptop and Zhang et al. (2021)~\citep{zhang2021_asqp} for ASQP Rest16, respectively. Above-average $\Delta$ distributions across both datasets mirror the pattern in Figure~\ref{fig4_metric_diff_bargraph} for EduRABSA, where FTS-OBP tolerance to small extraction-boundary differences mainly benefited pre-trained and LoRA-SFT models on small training sets.}
  \label{fig_semeval_metric_comparison_bar_graph}
\end{figure*}

\begin{figure}[htbp!]
    \centering   
    
    \begin{subfigure}[t]{0.48\textwidth}
        \centering
        \includegraphics[width=\textwidth]{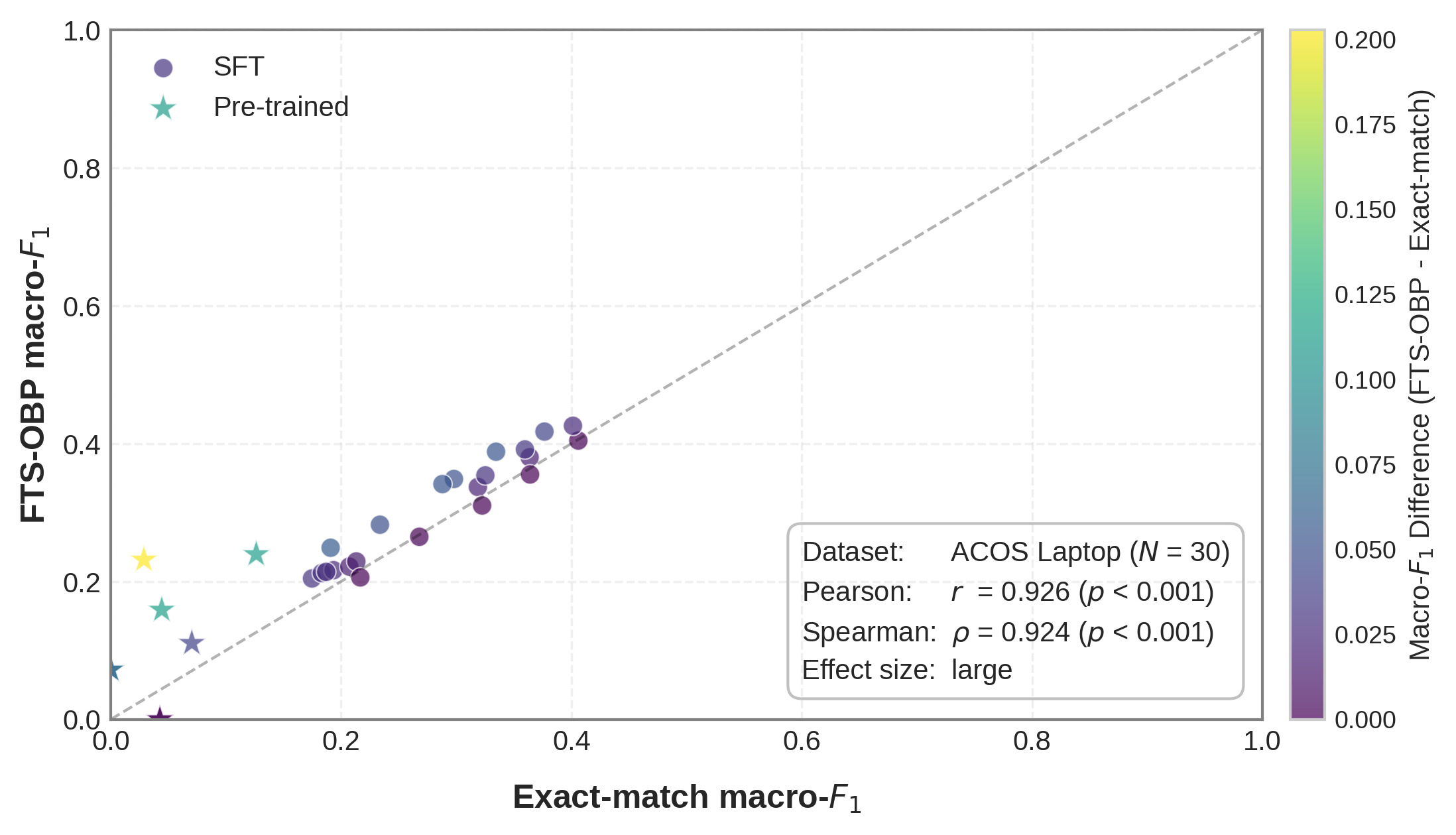}
        \caption{Dataset: ACOS Laptop}
        \label{fig_semeval_ACOS_Laptop_correlation_scatter}
    \end{subfigure}
    \hfill
    \begin{subfigure}[t]{0.48\textwidth}
        \centering
        \includegraphics[width=\textwidth]{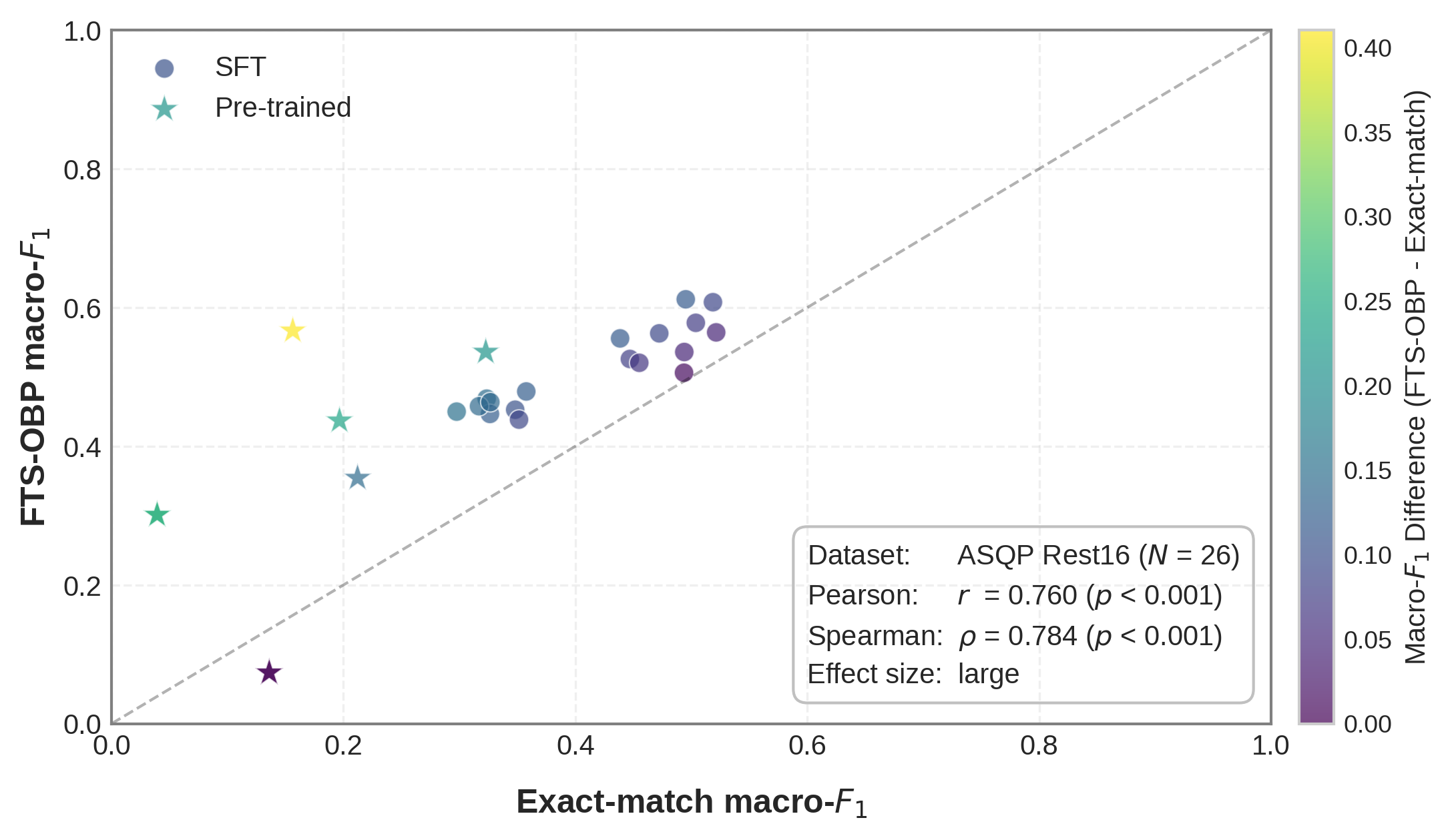}
        \caption{Dataset: ASQP Rest16}
        \label{fig_semeval_ASQP_Rest16_correlation_scatter}
    \end{subfigure}

    \caption{Scatter plot of macro-$F_1$ scores computed using FTS-OBP and Exact-match on the ASQE task with (a) the ACOS Laptop dataset ($N$ = 30 model-prompt pairs), and (b) the ASQP Rest16 dataset ($N$ = 26 model-prompt pairs).  The data points are coloured by the magnitude of their difference (FTS-OBP $-$ Exact-match). The patterns mirror those observed with the EduRABSA dataset in Figure \ref{fig_metric_correlation_scatter}: Strong inter-metric correlations are seen in both datasets (Pearson’s $r$ and Spearman’s $\rho$ are 0.926 and 0.924 for ACOS Laptop; and 0.760 and 0.784 for ASQP Rest16, all $p$ \textless 0.001), with larger differences (green to yellow) primarily occurring at relatively lower to medium performance levels and with pre-trained models. }
    \label{fig_semeval_scatterplot}
    
\end{figure}

\begin{figure*}[htbp!]
  \centering
  \includegraphics[width=0.55\linewidth]{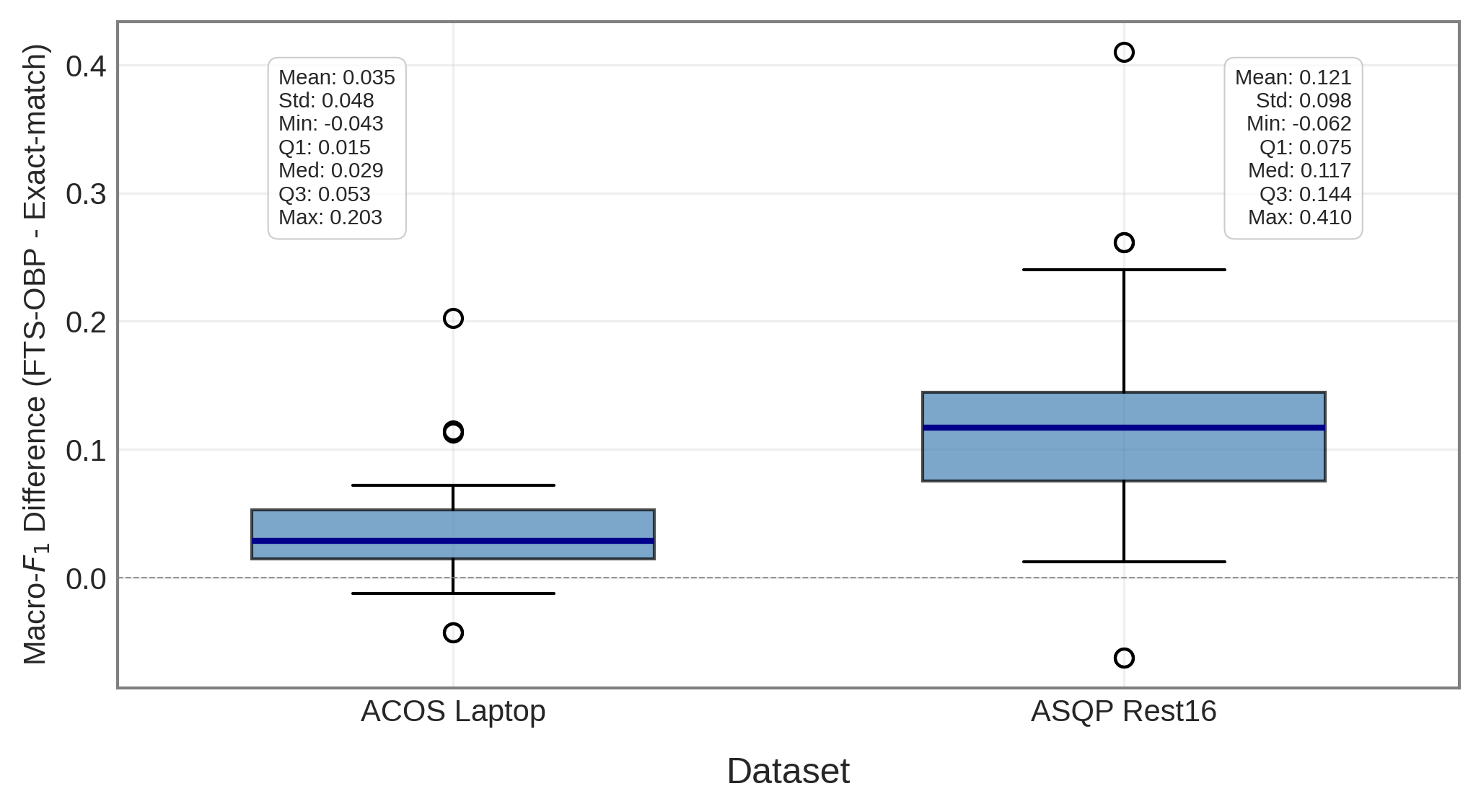}
  \caption{Distribution of macro-$F_1$ score differences (FTS-OBP $-$ Exact-match) on the ASQE task with the ACOS Laptop dataset ($N$ = 30 model-prompt pairs) and the ASQP Rest16 dataset ($N$ = 26 model-prompt pairs). FTS-OBP consistently yields higher scores. Compared with both the EduRABSA and ASQP Rest16 datasets, the ACOS Laptop dataset exhibits smaller metric differences in both magnitude and dispersion, as both metrics achieved lower scores than in the other two datasets.}

  \label{fig_semeval_metric_diff_boxplot}
\end{figure*}

\begin{figure*}[htbp!]
  \centering
  \includegraphics[width=\linewidth]{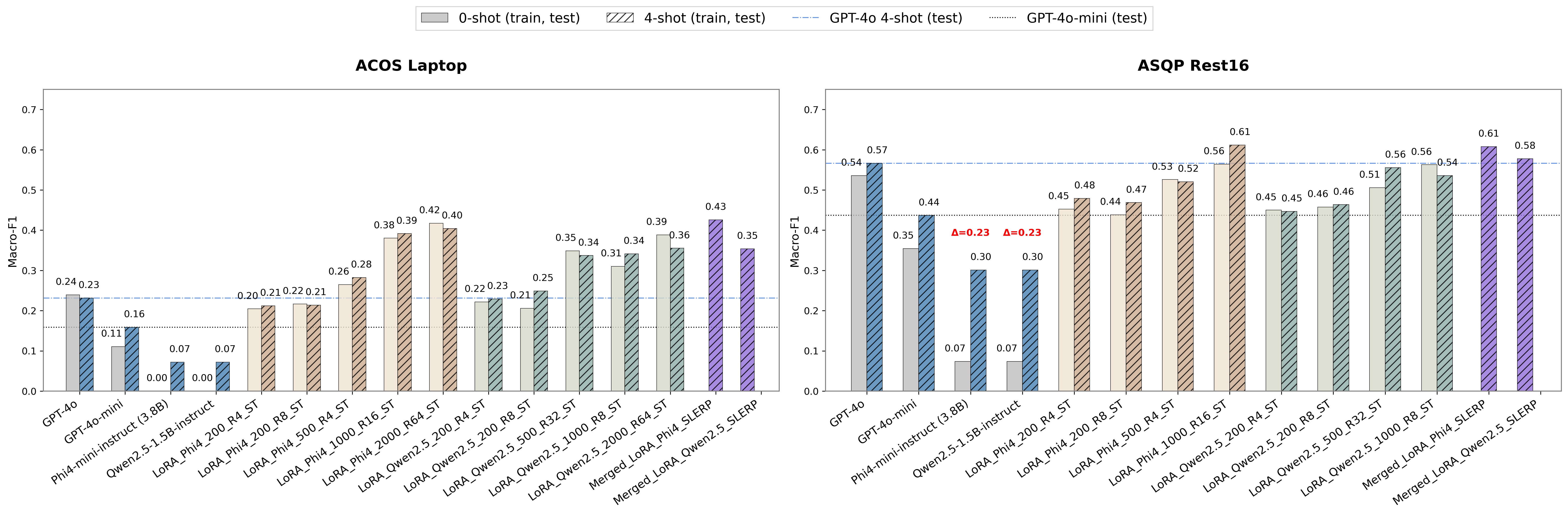}
  \caption{FTS-OBP Macro-$F_1$ scores on the ASQE task for pre-trained GLMs and SLMs, and LoRA-SFT and LoRA weight-merged SLMs using two benchmark test datasets (300 entries each), with 0-shot (0S) and 4-shot (4S) prompt input. $\Delta$ = 4S - 0S score (\textgreater 0.15). The figure shows a similar pattern to that observed in Figure~\ref{fig_overall_narrow_2bar_allmodels} and detailed in Section \ref{subsec_result_RQ2} for the EduRABSA dataset.}

  \label{fig_semeval_2bar_allmodels}
\end{figure*}

\newpage \clearpage
\subsection{Detailed Results for the ACOS Laptop Dataset}

\begin{figure*}[hb!]
  \centering
  \includegraphics[width=0.85\textwidth]{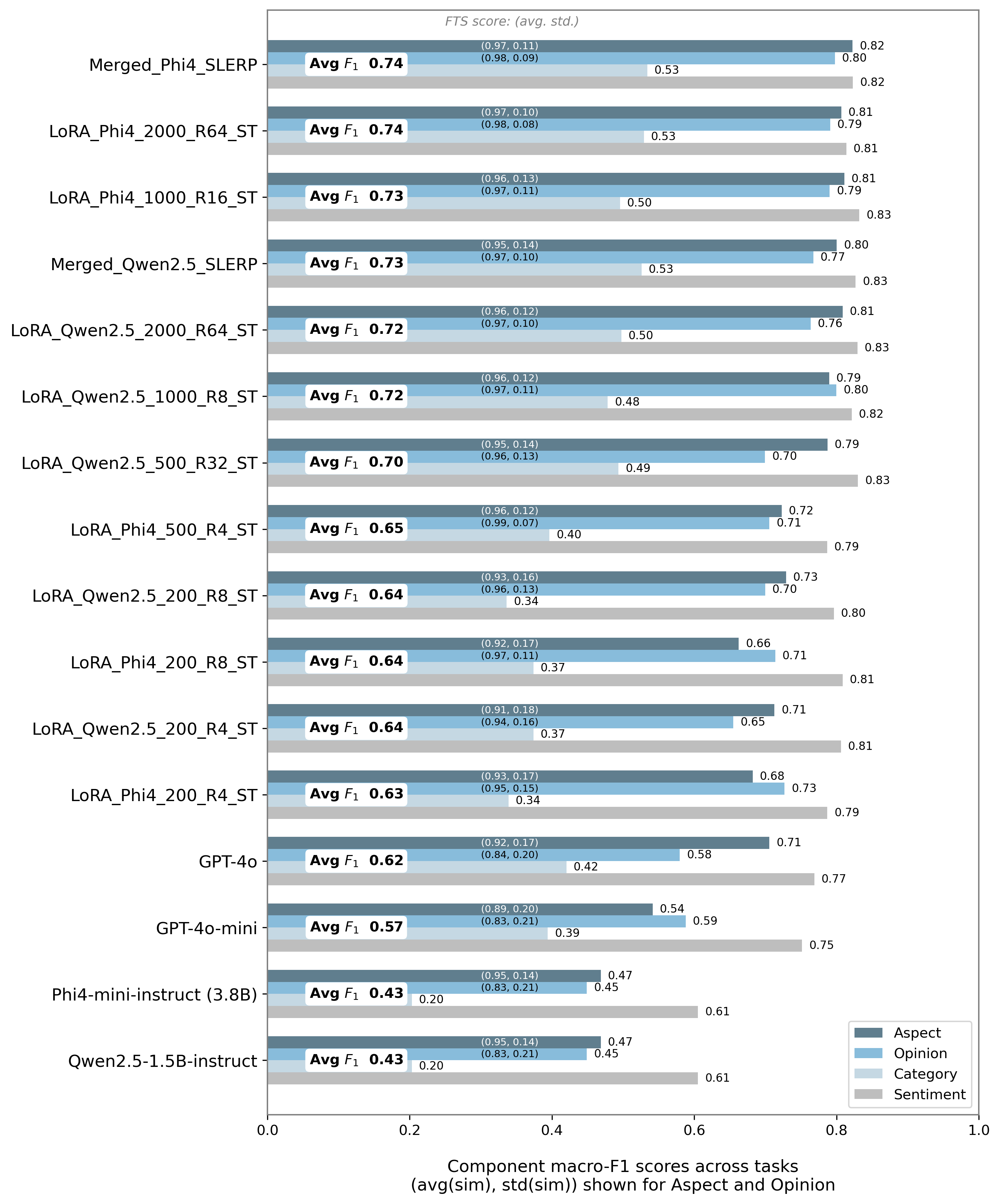}
  
  \caption{Macro-$F_1$ scores per component across entries on the ASQE task using the \textbf{ACOS Laptop} test dataset ($N$ = 300). Models include pre-trained GLMs and SLMs, and LoRA-SFT and LoRA weight-merged SLMs, and are ordered by descending cross-component average scores. ``Avg. $F_1$'' represents the mean value of a model's component macro-$F_1$ scores. The pair of numbers shown within each aspect or opinion bar represents the mean and standard deviation of that component’s FTS (similarity) score.}

  \label{fig_acos_component_metrics}
\end{figure*}

\begin{figure}[htbp!]
    \centering   
    
    \begin{subfigure}[c]{0.8\textwidth}
        \centering
        \includegraphics[height=0.41\textheight]{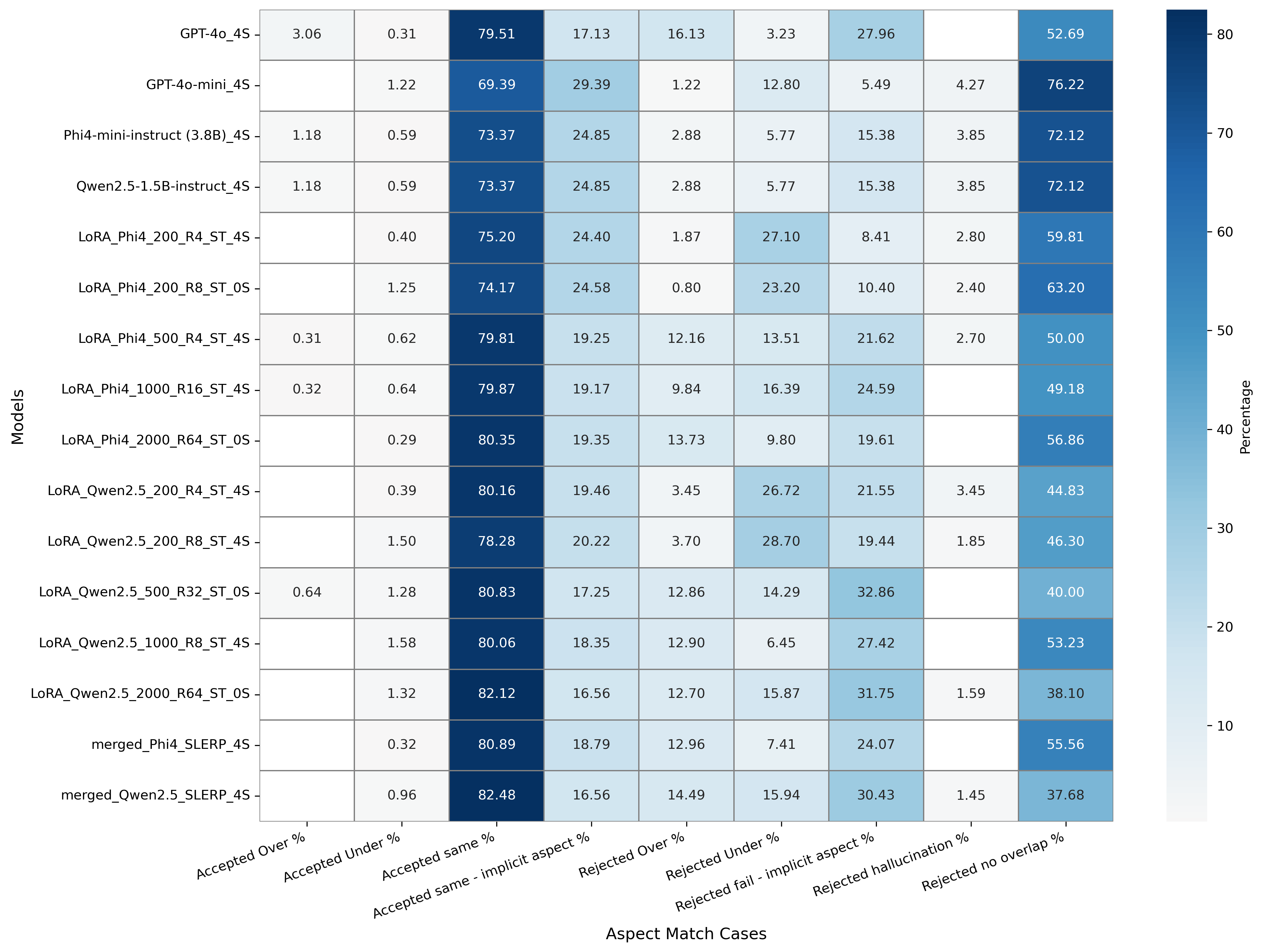}
        \caption{Percentage of matched aspect pairs among all accepted and rejected}
        \label{fig_ACOS_asp_heatmap}
    \end{subfigure}

    \begin{subfigure}[c]{0.8\textwidth}
        \centering
        \includegraphics[height=0.41\textheight]{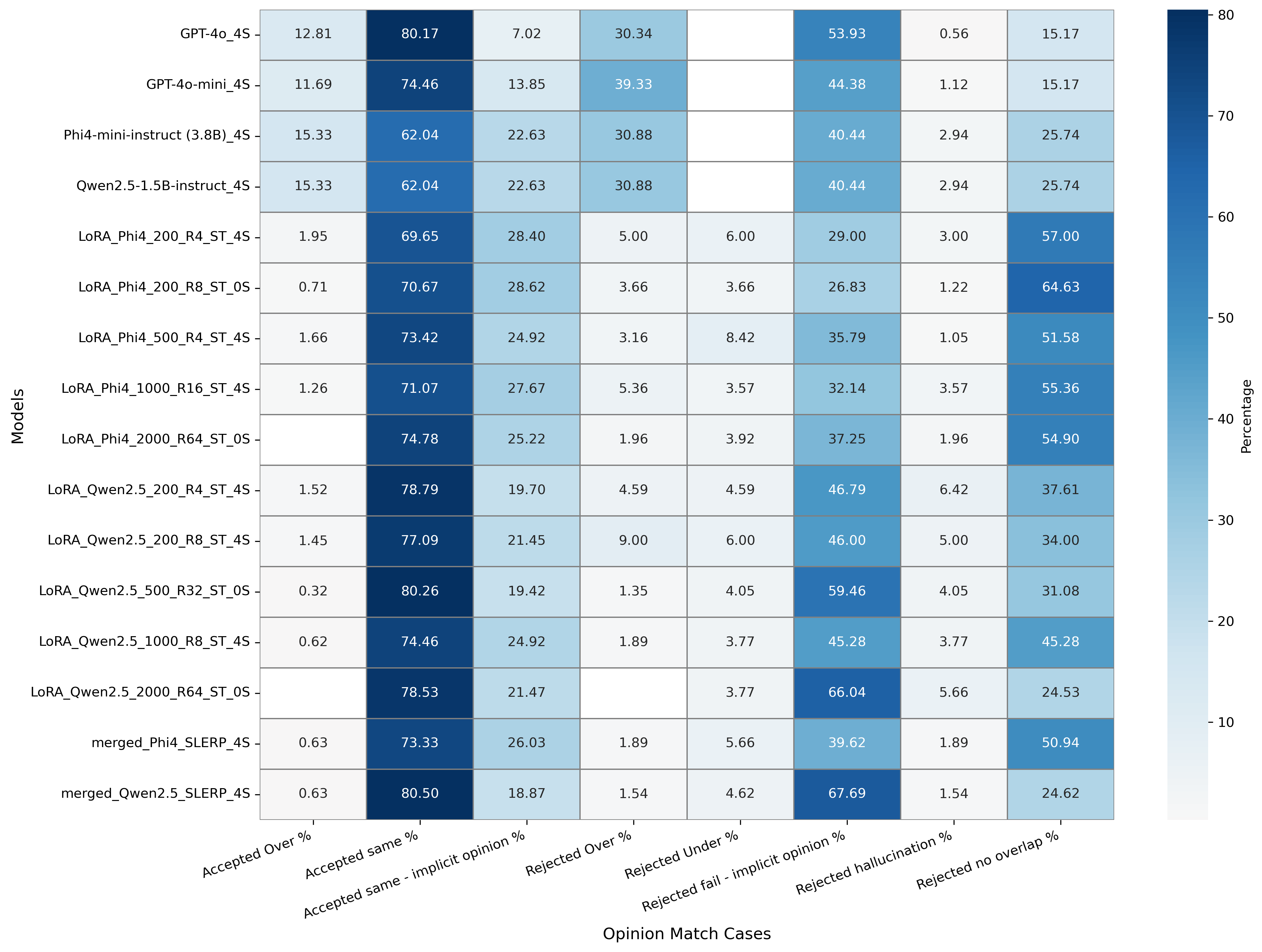}
        \caption{Percentage of matched opinion pairs among all accepted and rejected}
        \label{fig_ACOS_opn_heatmap}
    \end{subfigure}

    \caption{Percentage of ground-truth (gold) and model output (pred) aspect (a) and opinion (b) pairs in each match case, calculated over all pairs either accepted as a match or rejected by the FTS-OBP evaluation method, across related tasks using the \textbf{ACOS Laptop} test dataset ($N$ = 300). The match cases describe how pred differs from gold: 1) outside the original input text (``\textbf{hallucination}''), 2) extending beyond gold (``\textbf{over}''), 3) is a substring of gold (``\textbf{under}''), 4) partially overlapping with the gold and is not a substring of it (``\textbf{shift}''), and 5) \textbf{no overlap}. Cases 2)--4) were determined after removing the stopwords specified in Sec.~\ref{subsection_exp_metric}.}
    \label{fig_ACOS_asp_opn_heatmap}
\end{figure}

\newpage
\subsection{Detailed Results for the ASQP Rest16 Dataset}

\begin{figure*}[hb!]
  \centering
  \includegraphics[width=0.85\textwidth]{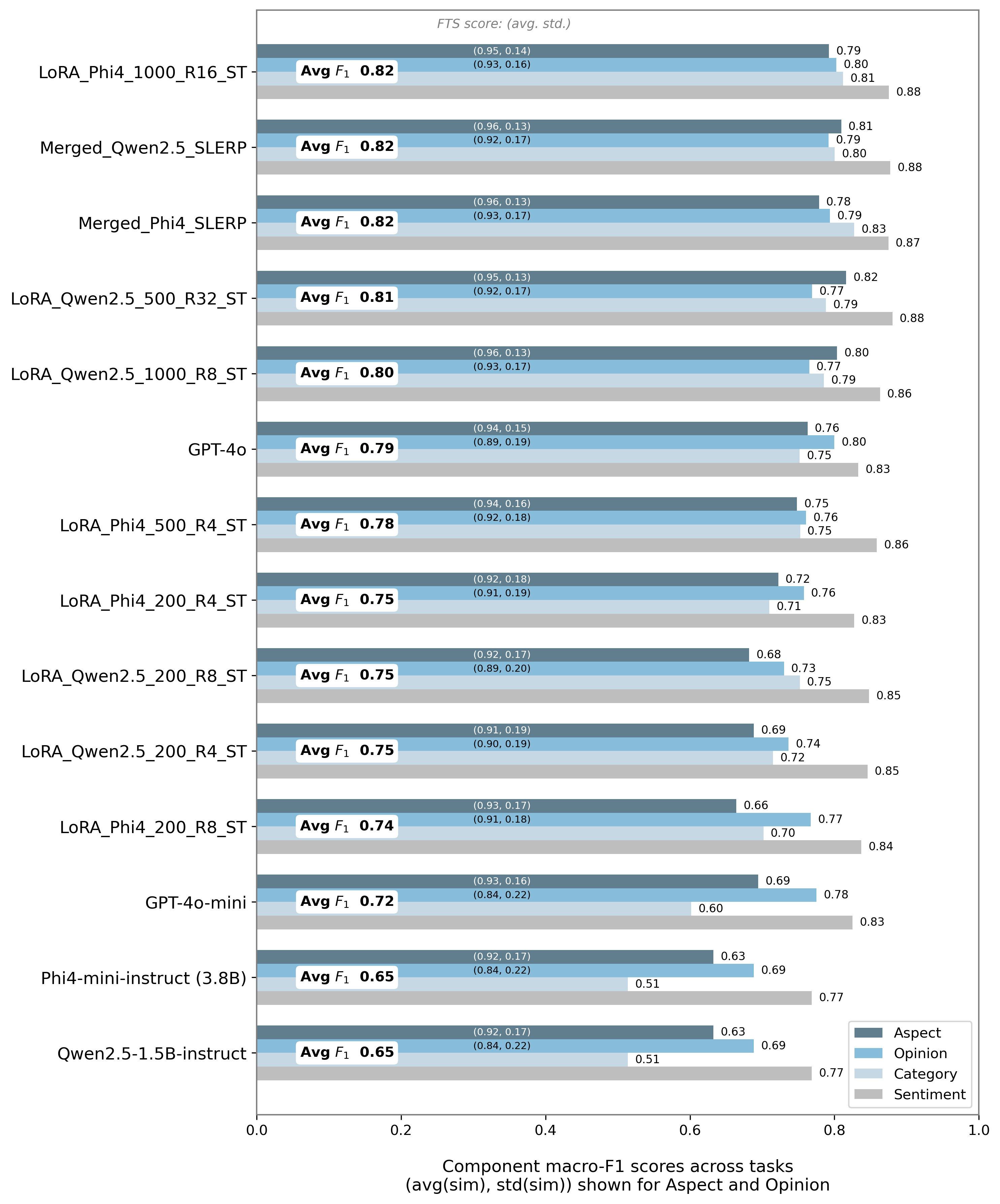}
  
  \caption{Macro-$F_1$ scores per component across entries on the ASQE task using the \textbf{ASQP Rest16} test dataset ($N$ = 300). Models include pre-trained GLMs and SLMs, and LoRA-SFT and LoRA weight-merged SLMs, and are ordered by descending cross-component average scores. ``Avg. $F_1$'' represents the mean value of a model's component macro-$F_1$ scores. The pair of numbers shown within each aspect or opinion bar represents the mean and standard deviation of that component’s FTS (similarity) score.}

  \label{fig_asqp_component_metrics}
\end{figure*}

\begin{figure}[htbp!]
    \centering   
    
    \begin{subfigure}[c]{0.8\textwidth}
        \centering
        \includegraphics[height=0.41\textheight]{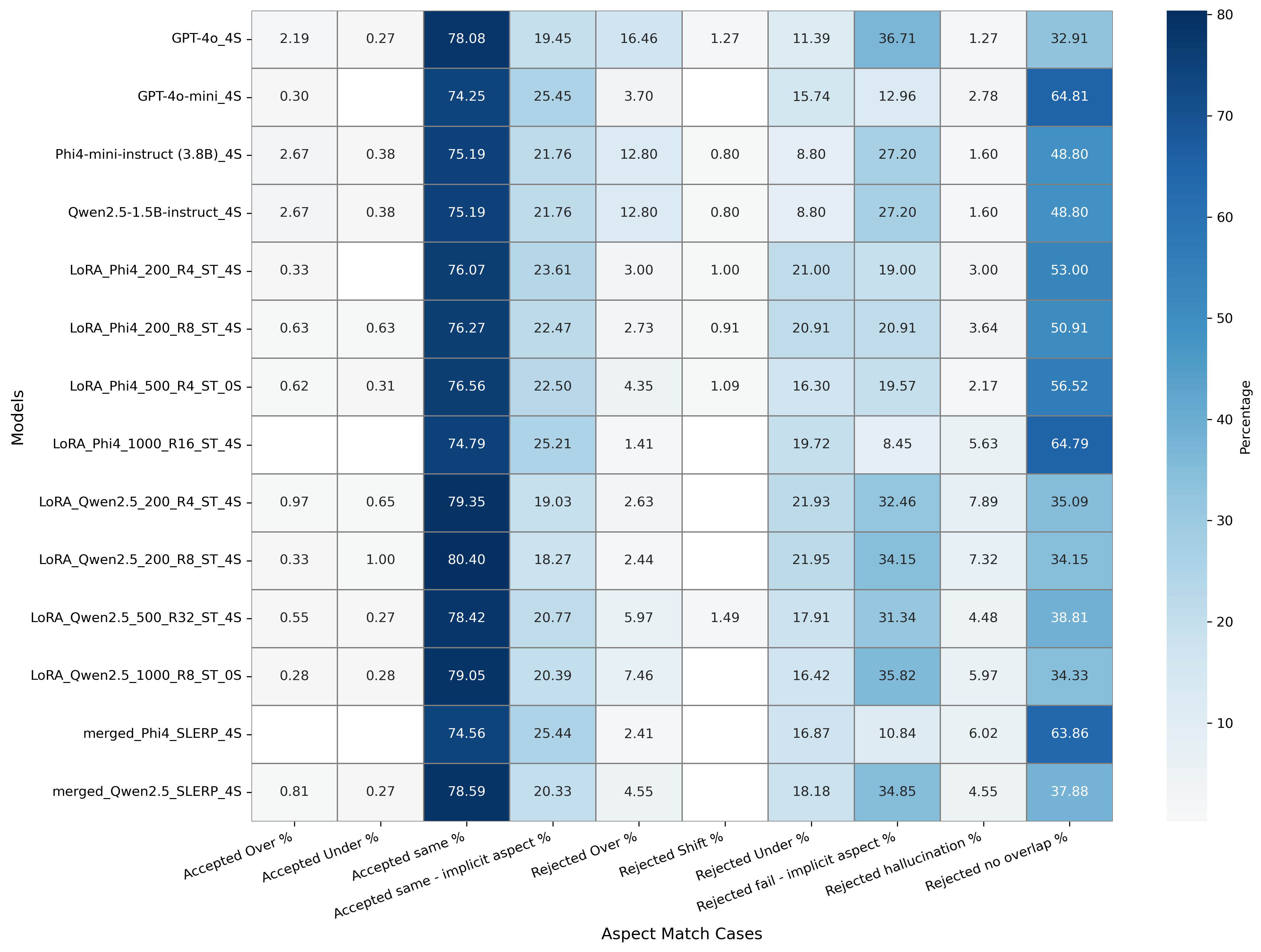}
        \caption{Percentage of matched aspect pairs among all accepted and rejected}
        \label{fig_ASQP_asp_heatmap}
    \end{subfigure}

    \begin{subfigure}[c]{0.8\textwidth}
        \centering
        \includegraphics[height=0.41\textheight]{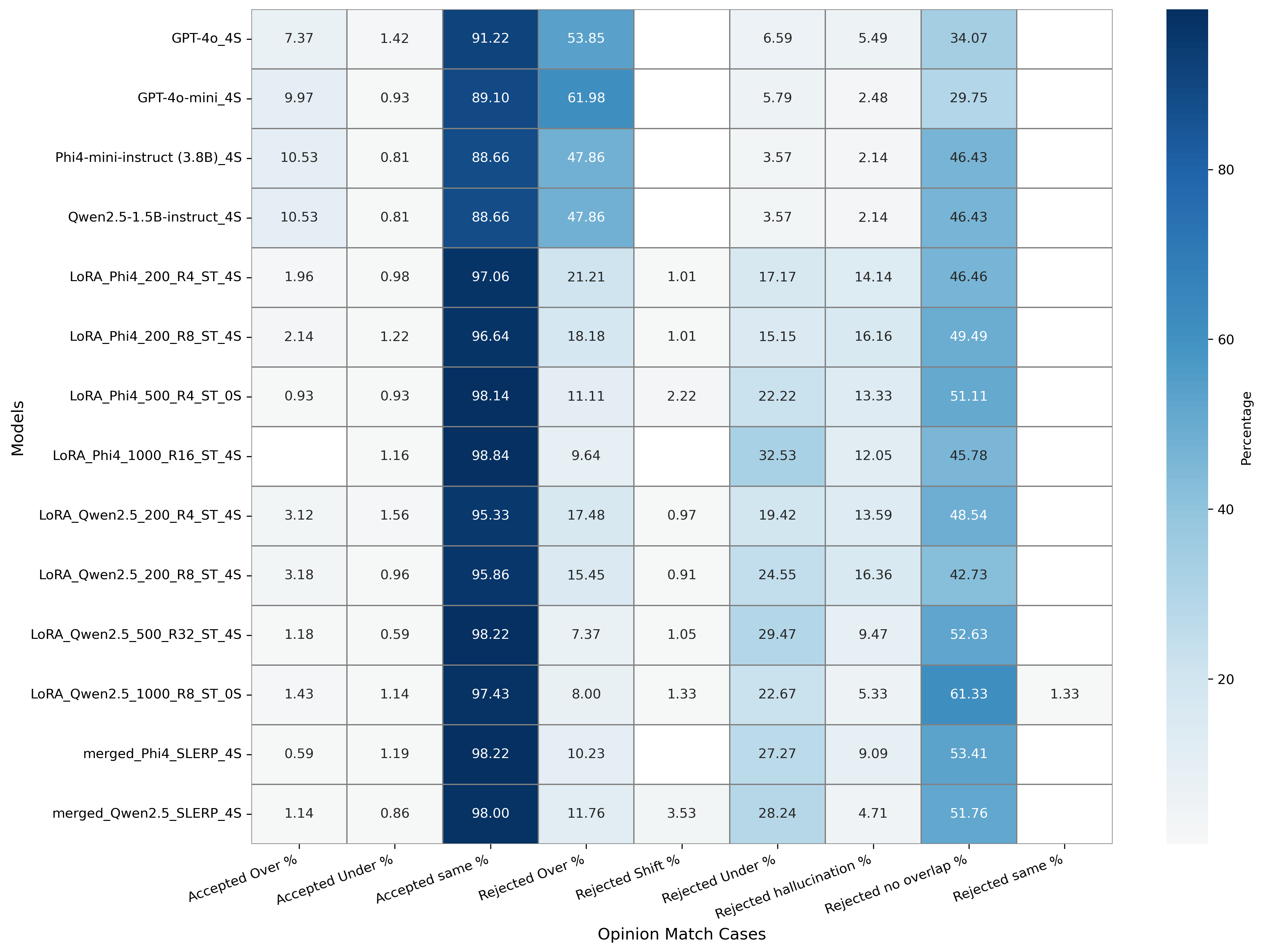}
        \caption{Percentage of matched opinion pairs among all accepted and rejected}
        \label{fig_ASQP_opn_heatmap}
    \end{subfigure}

    \caption{Percentage of ground-truth (gold) and model output (pred) aspect (a) and opinion (b) pairs in each match case, calculated over all pairs either accepted as a match or rejected by the FTS-OBP evaluation method, across related tasks using the \textbf{ASQP Rest16} test dataset ($N$ = 300). The match cases describe how pred differs from gold: 1) outside the original input text (``\textbf{hallucination}''), 2) extending beyond gold (``\textbf{over}''), 3) is a substring of gold (``\textbf{under}''), 4) partially overlapping with the gold and is not a substring of it (``\textbf{shift}''), and 5) \textbf{no overlap}. Cases 2)--4) were determined after removing the stopwords specified in Sec.~\ref{subsection_exp_metric}.}
    \label{fig_ASQP_asp_opn_heatmap}
\end{figure}


\clearpage
\bibliography{references}

@inproceedings{semeval2014,
    title = "{S}em{E}val-2014 Task 4: Aspect Based Sentiment Analysis",
    author = "Pontiki, Maria  and
      Galanis, Dimitris  and
      Pavlopoulos, John  and
      Papageorgiou, Harris  and
      Androutsopoulos, Ion  and
      Manandhar, Suresh",
    editor = "Nakov, Preslav  and
      Zesch, Torsten",
    booktitle = "Proceedings of the 8th International Workshop on Semantic Evaluation ({S}em{E}val 2014)",
    month = aug,
    year = "2014",
    address = "Dublin, Ireland",
    publisher = "Association for Computational Linguistics",
    url = "https://aclanthology.org/S14-2004",
    doi = "10.3115/v1/S14-2004",
    pages = "27--35",
}

@inproceedings{semeval2015,
    title = "{S}em{E}val-2015 Task 12: Aspect Based Sentiment Analysis",
    author = "Pontiki, Maria  and
      Galanis, Dimitris  and
      Papageorgiou, Haris  and
      Manandhar, Suresh  and
      Androutsopoulos, Ion",
    editor = "Nakov, Preslav  and
      Zesch, Torsten  and
      Cer, Daniel  and
      Jurgens, David",
    booktitle = "Proceedings of the 9th International Workshop on Semantic Evaluation ({S}em{E}val 2015)",
    month = jun,
    year = "2015",
    address = "Denver, Colorado",
    publisher = "Association for Computational Linguistics",
    url = "https://aclanthology.org/S15-2082",
    doi = "10.18653/v1/S15-2082",
    pages = "486--495",
}

@inproceedings{semeval2016,
    title = "{S}em{E}val-2016 Task 5: Aspect Based Sentiment Analysis",
    author = {Pontiki, Maria  and
      Galanis, Dimitris  and
      Papageorgiou, Haris  and
      Androutsopoulos, Ion  and
      Manandhar, Suresh  and
      AL-Smadi, Mohammad  and
      Al-Ayyoub, Mahmoud  and
      Zhao, Yanyan  and
      Qin, Bing  and
      De Clercq, Orph{\'e}e  and
      Hoste, V{\'e}ronique  and
      Apidianaki, Marianna  and
      Tannier, Xavier  and
      Loukachevitch, Natalia  and
      Kotelnikov, Evgeniy  and
      Bel, Nuria  and
      Jim{\'e}nez-Zafra, Salud Mar{\'\i}a  and
      Eryi{\u{g}}it, G{\"u}l{\c{s}}en},
    editor = "Bethard, Steven  and
      Carpuat, Marine  and
      Cer, Daniel  and
      Jurgens, David  and
      Nakov, Preslav  and
      Zesch, Torsten",
    booktitle = "Proceedings of the 10th International Workshop on Semantic Evaluation ({S}em{E}val-2016)",
    month = jun,
    year = "2016",
    address = "San Diego, California",
    publisher = "Association for Computational Linguistics",
    url = "https://aclanthology.org/S16-1002",
    doi = "10.18653/v1/S16-1002",
    pages = "19--30",
}

@inproceedings{mams2019,
 address = {Hong Kong, China},
 author = {Jiang, Qingnan and Chen, Lei and Xu, Ruifeng and Ao, Xiang and Yang, Min},
 booktitle = {Proceedings of the 2019 Conference on Empirical Methods in Natural Language Processing and the 9th International Joint Conference on Natural Language Processing ({EMNLP}-{IJCNLP})},
 doi = {10.18653/v1/D19-1654},
 eventtitle = {Proceedings of the 2019 Conference on Empirical Methods in Natural Language Processing and the 9th International Joint Conference on Natural Language Processing ({EMNLP}-{IJCNLP})},
 langid = {english},
 pages = {6279--6284},
 publisher = {Association for Computational Linguistics},
 title = {A Challenge Dataset and Effective Models for Aspect-Based Sentiment Analysis},
 url = {https://www.aclweb.org/anthology/D19-1654},
 urldate = {2023-09-16},
 year = {2019}
}

@article{hua2024absa,
  author    = {Hua, Yan Cathy and Denny, Paul and Wicker, J{\"o}rg and Ta\v{s}kova, Katerina},
  title     = {A systematic review of aspect-based sentiment analysis: domains, methods, and trends},
  journal   = {Artificial Intelligence Review},
  year      = {2024},
  volume    = {57},
  number    = {11},
  pages     = {296},
  month     = sep,
  doi       = {10.1007/s10462-024-10906-z},
  url       = {https://doi.org/10.1007/s10462-024-10906-z},
  issn      = {1573-7462}
}

@misc{hua2025edurabsa,
      title={EduRABSA: An Education Review Dataset for Aspect-based Sentiment Analysis Tasks}, 
      author={Hua, Yan Cathy and Denny, Paul and Wicker, J{\"o}rg and Ta\v{s}kova, Katerina},
      year={2025},
      eprint={2508.17008},
      archivePrefix={arXiv},
      primaryClass={cs.CL},
      url={https://arxiv.org/abs/2508.17008}, 
}

@article{fei2026robustness,
 author = {Fei, Hao and Chua, Tat-Seng and Li, Chenliang and Ji, Donghong and Zhang, Meishan and Ren, Yafeng},
 doi = {10.1145/3564281},
 issn = {1046-8188, 1558-2868},
 journal = {{ACM} Transactions on Information Systems},
 langid = {english},
 number = {2},
 pages = {1--32},
 shortjournal = {{ACM} Trans. Inf. Syst.},
 shorttitle = {On the Robustness of Aspect-based Sentiment Analysis},
 title = {On the Robustness of Aspect-based Sentiment Analysis: Rethinking Model, Data, and Training},
 url = {https://dl.acm.org/doi/10.1145/3564281},
 urldate = {2023-09-16},
 volume = {41},
 year = {2023}
}

@misc{datasetsurvey2023,
 archivePrefix = {arXiv},
 author = {Siva Uday Sampreeth Chebolu and Franck Dernoncourt and Nedim Lipka and Thamar Solorio},
 eprint = {2204.05232},
 primaryclass = {cs.CL},
 title = {Survey of Aspect-based Sentiment Analysis Datasets},
 year = {2023},
 doi = {10.48550/arXiv.2204.05232},
 url={https://arxiv.org/abs/2204.05232}
}

@inproceedings{sk2_jointabsa,
author = {Li, Jia and Zhao, Yuyuan and Jin, Zhi and Li, Ge and Shen, Tao and Tao, Zhengwei and Tao, Chongyang},
title = {SK2: Integrating Implicit Sentiment Knowledge and Explicit Syntax Knowledge for Aspect-Based Sentiment Analysis},
year = {2022},
isbn = {9781450392365},
publisher = {Association for Computing Machinery},
address = {New York, NY, USA},
url = {https://doi.org/10.1145/3511808.3557452},
doi = {10.1145/3511808.3557452},
booktitle = {Proceedings of the 31st ACM International Conference on Information \& Knowledge Management},
pages = {1114–1123},
numpages = {10},
location = {Atlanta, GA, USA},
series = {CIKM '22}
}

@article{jointabsa,
 author = {Li, You and Lin, Yongdong and Lin, Yuming and Chang, Liang and Zhang, Huibing},
 doi = {10.1016/j.knosys.2022.108366},
 issn = {09507051},
 journal = {Knowledge-Based Systems},
 langid = {english},
 pages = {108366},
 shortjournal = {Knowledge-Based Systems},
 title = {A span-sharing joint extraction framework for harvesting aspect sentiment triplets},
 url = {https://linkinghub.elsevier.com/retrieve/pii/S0950705122001381},
 urldate = {2023-09-16},
 volume = {242},
 year = {2022}
}

@article{joint_aste,
author = {Zhang, Zhihao and Zuo, Yuan and Wu, Junjie},
title = {Aspect Sentiment Triplet Extraction: A Seq2Seq Approach With Span Copy Enhanced Dual Decoder},
year = {2022},
issue_date = {2022},
publisher = {IEEE Press},
volume = {30},
issn = {2329-9290},
url = {https://doi.org/10.1109/TASLP.2022.3198802},
doi = {10.1109/TASLP.2022.3198802},
journal = {IEEE/ACM Trans. Audio, Speech and Lang. Proc.},
month = aug,
pages = {2729–2742},
numpages = {14}
}

@INPROCEEDINGS{batch2_survey_absadl,
  author={Satyarthi, Shailendra and Sharma, Sanjiv},
  booktitle={2023 IEEE International Conference on Paradigm Shift in Information Technologies with Innovative Applications in Global Scenario (ICPSITIAGS)}, 
  title={Identification of Effective Deep Learning Approaches for Classifying Sentiments at Aspect Level in Different Domain}, 
  year={2023},
  volume={},
  number={},
  pages={496-508},
  doi={10.1109/ICPSITIAGS59213.2023.10527695},
url={https://ieeexplore-ieee-org.ezproxy.auckland.ac.nz/document/10527695}
}

@article{batch2_survey_absa,
author = {Zhang, Wenxuan and Li, Xin and Deng, Yang and Bing, Lidong and Lam, Wai},
title = {A Survey on Aspect-Based Sentiment Analysis: Tasks, Methods, and Challenges},
year = {2022},
issue_date = {Nov. 2023},
publisher = {IEEE Educational Activities Department},
address = {USA},
volume = {35},
number = {11},
issn = {1041-4347},
url = {https://doi.org/10.1109/TKDE.2022.3230975},
doi = {10.1109/TKDE.2022.3230975},
journal = {IEEE Trans. on Knowl. and Data Eng.},
month = {dec},
pages = {11019–11038},
numpages = {20}
}

@misc{surveyincontextlearning,
      title={A Survey on In-context Learning}, 
      author={Qingxiu Dong and Lei Li and Damai Dai and Ce Zheng and Jingyuan Ma and Rui Li and Heming Xia and Jingjing Xu and Zhiyong Wu and Baobao Chang and Xu Sun and Lei Li and Zhifang Sui},
      year={2024},
      eprint={2301.00234},
      archivePrefix={arXiv},
      primaryClass={cs.CL},
      url={https://arxiv.org/abs/2301.00234}, 
}

@article{Zhang2024,
    author    = {Hao Zhang and Yu-N Cheah and Osamah Mohammed Alyasiri and Jieyu An},
    title     = {Exploring aspect-based sentiment quadruple extraction with implicit aspects, opinions, and ChatGPT: a comprehensive survey},
    journal   = {Artificial Intelligence Review},
    volume    = {57},
    number    = {2},
    pages     = {17},
    year      = {2024},
    doi       = {10.1007/s10462-023-10633-x},
    url       = {https://doi.org/10.1007/s10462-023-10633-x},
    issn      = {1573-7462}
}

@misc{foundationmodels,
      title={On the Opportunities and Risks of Foundation Models}, 
      author={Rishi Bommasani and Drew A. Hudson and Ehsan Adeli and Russ Altman and Simran Arora and Sydney von Arx and Michael S. Bernstein and Jeannette Bohg and Antoine Bosselut and Emma Brunskill and Erik Brynjolfsson and Shyamal Buch and Dallas Card and Rodrigo Castellon and Niladri Chatterji and Annie Chen and Kathleen Creel and Jared Quincy Davis and Dora Demszky and Chris Donahue and Moussa Doumbouya and Esin Durmus and Stefano Ermon and John Etchemendy and Kawin Ethayarajh and Li Fei-Fei and Chelsea Finn and Trevor Gale and Lauren Gillespie and Karan Goel and Noah Goodman and Shelby Grossman and Neel Guha and Tatsunori Hashimoto and Peter Henderson and John Hewitt and Daniel E. Ho and Jenny Hong and Kyle Hsu and Jing Huang and Thomas Icard and Saahil Jain and Dan Jurafsky and Pratyusha Kalluri and Siddharth Karamcheti and Geoff Keeling and Fereshte Khani and Omar Khattab and Pang Wei Koh and Mark Krass and Ranjay Krishna and Rohith Kuditipudi and Ananya Kumar and Faisal Ladhak and Mina Lee and Tony Lee and Jure Leskovec and Isabelle Levent and Xiang Lisa Li and Xuechen Li and Tengyu Ma and Ali Malik and Christopher D. Manning and Suvir Mirchandani and Eric Mitchell and Zanele Munyikwa and Suraj Nair and Avanika Narayan and Deepak Narayanan and Ben Newman and Allen Nie and Juan Carlos Niebles and Hamed Nilforoshan and Julian Nyarko and Giray Ogut and Laurel Orr and Isabel Papadimitriou and Joon Sung Park and Chris Piech and Eva Portelance and Christopher Potts and Aditi Raghunathan and Rob Reich and Hongyu Ren and Frieda Rong and Yusuf Roohani and Camilo Ruiz and Jack Ryan and Christopher Ré and Dorsa Sadigh and Shiori Sagawa and Keshav Santhanam and Andy Shih and Krishnan Srinivasan and Alex Tamkin and Rohan Taori and Armin W. Thomas and Florian Tramèr and Rose E. Wang and William Wang and Bohan Wu and Jiajun Wu and Yuhuai Wu and Sang Michael Xie and Michihiro Yasunaga and Jiaxuan You and Matei Zaharia and Michael Zhang and Tianyi Zhang and Xikun Zhang and Yuhui Zhang and Lucia Zheng and Kaitlyn Zhou and Percy Liang},
      year={2022},
      eprint={2108.07258},
      archivePrefix={arXiv},
      primaryClass={cs.LG},
      url={https://arxiv.org/abs/2108.07258}, 
}

@inproceedings{word2vec,
 author = {Mikolov, Tomas and Sutskever, Ilya and Chen, Kai and Corrado, Greg S and Dean, Jeff},
 booktitle = {Advances in Neural Information Processing Systems},
 editor = {C.J. Burges and L. Bottou and M. Welling and Z. Ghahramani and K.Q. Weinberger},
 pages = {},
 publisher = {Curran Associates, Inc.},
 title = {Distributed Representations of Words and Phrases and their Compositionality},
 url = {https://proceedings.neurips.cc/paper_files/paper/2013/file/9aa42b31882ec039965f3c4923ce901b-Paper.pdf},
 volume = {26},
 year = {2013}
}

@inproceedings{glove,
    title = "{G}lo{V}e: Global Vectors for Word Representation",
    author = "Pennington, Jeffrey  and
      Socher, Richard  and
      Manning, Christopher",
    editor = "Moschitti, Alessandro  and
      Pang, Bo  and
      Daelemans, Walter",
    booktitle = "Proceedings of the 2014 Conference on Empirical Methods in Natural Language Processing ({EMNLP})",
    month = oct,
    year = "2014",
    address = "Doha, Qatar",
    publisher = "Association for Computational Linguistics",
    url = "https://aclanthology.org/D14-1162/",
    doi = "10.3115/v1/D14-1162",
    pages = "1532--1543"
}

@inproceedings{bert,
  author       = {Jacob Devlin and
                  Ming{-}Wei Chang and
                  Kenton Lee and
                  Kristina Toutanova},
  editor       = {Jill Burstein and
                  Christy Doran and
                  Thamar Solorio},
  title        = {{BERT:} Pre-training of Deep Bidirectional Transformers for Language
                  Understanding},
  booktitle    = {Proceedings of the 2019 Conference of the North American Chapter of
                  the Association for Computational Linguistics: Human Language Technologies,
                  {NAACL-HLT} 2019, Minneapolis, MN, USA, June 2-7, 2019, Volume 1 (Long
                  and Short Papers)},
  pages        = {4171--4186},
  publisher    = {Association for Computational Linguistics},
  year         = {2019},
  url          = {https://doi.org/10.18653/v1/n19-1423},
  doi          = {10.18653/V1/N19-1423},
  bibsource    = {dblp computer science bibliography, https://dblp.org}
}

@misc{roberta,
      title={RoBERTa: A Robustly Optimized BERT Pretraining Approach}, 
      author={Yinhan Liu and Myle Ott and Naman Goyal and Jingfei Du and Mandar Joshi and Danqi Chen and Omer Levy and Mike Lewis and Luke Zettlemoyer and Veselin Stoyanov},
      year={2019},
      eprint={1907.11692},
      archivePrefix={arXiv},
      primaryClass={cs.CL},
      url={https://arxiv.org/abs/1907.11692}, 
}

@article{T5,
    author = {Raffel, Colin and Shazeer, Noam and Roberts, Adam and Lee, Katherine and Narang, Sharan and Matena, Michael and Zhou, Yanqi and Li, Wei and Liu, Peter J.},
    title = {Exploring the limits of transfer learning with a unified text-to-text transformer},
    year = {2020},
    issue_date = {January 2020},
    publisher = {JMLR.org},
    volume = {21},
    number = {1},
    issn = {1532-4435},
    journal = {J. Mach. Learn. Res.},
    month = {jan},
    articleno = {140},
    numpages = {67},
    url={https://dl.acm.org/doi/abs/10.5555/3455716.3455856}
}

@misc{refGPT3,
    archivePrefix = {arXiv},
    author = {Tom B. Brown and Benjamin Mann and Nick Ryder and Melanie Subbiah and Jared Kaplan and Prafulla Dhariwal and Arvind Neelakantan and Pranav Shyam and Girish Sastry and Amanda Askell and Sandhini Agarwal and Ariel Herbert-Voss and Gretchen Krueger and Tom Henighan and Rewon Child and Aditya Ramesh and Daniel M. Ziegler and Jeffrey Wu and Clemens Winter and Christopher Hesse and Mark Chen and Eric Sigler and Mateusz Litwin and Scott Gray and Benjamin Chess and Jack Clark and Christopher Berner and Sam McCandlish and Alec Radford and Ilya Sutskever and Dario Amodei},
    eprint = {2005.14165},
    primaryclass = {cs.CL},
    title = {Language Models are Few-Shot Learners},
    year = {2020},
    doi = {10.48550/arXiv.2005.14165}
}

@techreport{claude3,
  author       = {Anthropic},
  title        = {The Claude 3 Model Family: Opus, Sonnet, Haiku},
  institution  = {Anthropic},
  type         = {Model Card},
  year         = {2024},
  url         = {https://www-cdn.anthropic.com/de8ba9b01c9ab7cbabf5c33b80b7bbc618857627/Model_Card_Claude_3.pdf},
}

@misc{llama3,
      title={The Llama 3 Herd of Models}, 
      author={Aaron Grattafiori and Abhimanyu Dubey and Abhinav Jauhri and Abhinav Pandey and Abhishek Kadian and Ahmad Al-Dahle and Aiesha Letman and Akhil Mathur and Alan Schelten and Alex Vaughan and Amy Yang and Angela Fan and Anirudh Goyal and Anthony Hartshorn and Aobo Yang and Archi Mitra and Archie Sravankumar and Artem Korenev and Arthur Hinsvark and Arun Rao and Aston Zhang and Aurelien Rodriguez and Austen Gregerson and Ava Spataru and Baptiste Roziere and Bethany Biron and Binh Tang and Bobbie Chern and Charlotte Caucheteux and Chaya Nayak and Chloe Bi and Chris Marra and Chris McConnell and Christian Keller and Christophe Touret and Chunyang Wu and Corinne Wong and Cristian Canton Ferrer and Cyrus Nikolaidis and Damien Allonsius and Daniel Song and Danielle Pintz and Danny Livshits and Danny Wyatt and David Esiobu and Dhruv Choudhary and Dhruv Mahajan and Diego Garcia-Olano and Diego Perino and Dieuwke Hupkes and Egor Lakomkin and Ehab AlBadawy and Elina Lobanova and Emily Dinan and Eric Michael Smith and Filip Radenovic and Francisco Guzmán and Frank Zhang and Gabriel Synnaeve and Gabrielle Lee and Georgia Lewis Anderson and Govind Thattai and Graeme Nail and Gregoire Mialon and Guan Pang and Guillem Cucurell and Hailey Nguyen and Hannah Korevaar and Hu Xu and Hugo Touvron and Iliyan Zarov and Imanol Arrieta Ibarra and Isabel Kloumann and Ishan Misra and Ivan Evtimov and Jack Zhang and Jade Copet and Jaewon Lee and Jan Geffert and Jana Vranes and Jason Park and Jay Mahadeokar and Jeet Shah and Jelmer van der Linde and Jennifer Billock and Jenny Hong and Jenya Lee and Jeremy Fu and Jianfeng Chi and Jianyu Huang and Jiawen Liu and Jie Wang and Jiecao Yu and Joanna Bitton and Joe Spisak and Jongsoo Park and Joseph Rocca and Joshua Johnstun and Joshua Saxe and Junteng Jia and Kalyan Vasuden Alwala and Karthik Prasad and Kartikeya Upasani and Kate Plawiak and Ke Li and Kenneth Heafield and Kevin Stone and Khalid El-Arini and Krithika Iyer and Kshitiz Malik and Kuenley Chiu and Kunal Bhalla and Kushal Lakhotia and Lauren Rantala-Yeary and Laurens van der Maaten and Lawrence Chen and Liang Tan and Liz Jenkins and Louis Martin and Lovish Madaan and Lubo Malo and Lukas Blecher and Lukas Landzaat and Luke de Oliveira and Madeline Muzzi and Mahesh Pasupuleti and Mannat Singh and Manohar Paluri and Marcin Kardas and Maria Tsimpoukelli and Mathew Oldham and Mathieu Rita and Maya Pavlova and Melanie Kambadur and Mike Lewis and Min Si and Mitesh Kumar Singh and Mona Hassan and Naman Goyal and Narjes Torabi and Nikolay Bashlykov and Nikolay Bogoychev and Niladri Chatterji and Ning Zhang and Olivier Duchenne and Onur Çelebi and Patrick Alrassy and Pengchuan Zhang and Pengwei Li and Petar Vasic and Peter Weng and Prajjwal Bhargava and Pratik Dubal and Praveen Krishnan and Punit Singh Koura and Puxin Xu and Qing He and Qingxiao Dong and Ragavan Srinivasan and Raj Ganapathy and Ramon Calderer and Ricardo Silveira Cabral and Robert Stojnic and Roberta Raileanu and Rohan Maheswari and Rohit Girdhar and Rohit Patel and Romain Sauvestre and Ronnie Polidoro and Roshan Sumbaly and Ross Taylor and Ruan Silva and Rui Hou and Rui Wang and Saghar Hosseini and Sahana Chennabasappa and Sanjay Singh and Sean Bell and Seohyun Sonia Kim and Sergey Edunov and Shaoliang Nie and Sharan Narang and Sharath Raparthy and Sheng Shen and Shengye Wan and Shruti Bhosale and Shun Zhang and Simon Vandenhende and Soumya Batra and Spencer Whitman and Sten Sootla and Stephane Collot and Suchin Gururangan and Sydney Borodinsky and Tamar Herman and Tara Fowler and Tarek Sheasha and Thomas Georgiou and Thomas Scialom and Tobias Speckbacher and Todor Mihaylov and Tong Xiao and Ujjwal Karn and Vedanuj Goswami and Vibhor Gupta and Vignesh Ramanathan and Viktor Kerkez and Vincent Gonguet and Virginie Do and Vish Vogeti and Vítor Albiero and Vladan Petrovic and Weiwei Chu and Wenhan Xiong and Wenyin Fu and Whitney Meers and Xavier Martinet and Xiaodong Wang and Xiaofang Wang and Xiaoqing Ellen Tan and Xide Xia and Xinfeng Xie and Xuchao Jia and Xuewei Wang and Yaelle Goldschlag and Yashesh Gaur and Yasmine Babaei and Yi Wen and Yiwen Song and Yuchen Zhang and Yue Li and Yuning Mao and Zacharie Delpierre Coudert and Zheng Yan and Zhengxing Chen and Zoe Papakipos and Aaditya Singh and Aayushi Srivastava and Abha Jain and Adam Kelsey and Adam Shajnfeld and Adithya Gangidi and Adolfo Victoria and Ahuva Goldstand and Ajay Menon and Ajay Sharma and Alex Boesenberg and Alexei Baevski and Allie Feinstein and Amanda Kallet and Amit Sangani and Amos Teo and Anam Yunus and Andrei Lupu and Andres Alvarado and Andrew Caples and Andrew Gu and Andrew Ho and Andrew Poulton and Andrew Ryan and Ankit Ramchandani and Annie Dong and Annie Franco and Anuj Goyal and Aparajita Saraf and Arkabandhu Chowdhury and Ashley Gabriel and Ashwin Bharambe and Assaf Eisenman and Azadeh Yazdan and Beau James and Ben Maurer and Benjamin Leonhardi and Bernie Huang and Beth Loyd and Beto De Paola and Bhargavi Paranjape and Bing Liu and Bo Wu and Boyu Ni and Braden Hancock and Bram Wasti and Brandon Spence and Brani Stojkovic and Brian Gamido and Britt Montalvo and Carl Parker and Carly Burton and Catalina Mejia and Ce Liu and Changhan Wang and Changkyu Kim and Chao Zhou and Chester Hu and Ching-Hsiang Chu and Chris Cai and Chris Tindal and Christoph Feichtenhofer and Cynthia Gao and Damon Civin and Dana Beaty and Daniel Kreymer and Daniel Li and David Adkins and David Xu and Davide Testuggine and Delia David and Devi Parikh and Diana Liskovich and Didem Foss and Dingkang Wang and Duc Le and Dustin Holland and Edward Dowling and Eissa Jamil and Elaine Montgomery and Eleonora Presani and Emily Hahn and Emily Wood and Eric-Tuan Le and Erik Brinkman and Esteban Arcaute and Evan Dunbar and Evan Smothers and Fei Sun and Felix Kreuk and Feng Tian and Filippos Kokkinos and Firat Ozgenel and Francesco Caggioni and Frank Kanayet and Frank Seide and Gabriela Medina Florez and Gabriella Schwarz and Gada Badeer and Georgia Swee and Gil Halpern and Grant Herman and Grigory Sizov and Guangyi and Zhang and Guna Lakshminarayanan and Hakan Inan and Hamid Shojanazeri and Han Zou and Hannah Wang and Hanwen Zha and Haroun Habeeb and Harrison Rudolph and Helen Suk and Henry Aspegren and Hunter Goldman and Hongyuan Zhan and Ibrahim Damlaj and Igor Molybog and Igor Tufanov and Ilias Leontiadis and Irina-Elena Veliche and Itai Gat and Jake Weissman and James Geboski and James Kohli and Janice Lam and Japhet Asher and Jean-Baptiste Gaya and Jeff Marcus and Jeff Tang and Jennifer Chan and Jenny Zhen and Jeremy Reizenstein and Jeremy Teboul and Jessica Zhong and Jian Jin and Jingyi Yang and Joe Cummings and Jon Carvill and Jon Shepard and Jonathan McPhie and Jonathan Torres and Josh Ginsburg and Junjie Wang and Kai Wu and Kam Hou U and Karan Saxena and Kartikay Khandelwal and Katayoun Zand and Kathy Matosich and Kaushik Veeraraghavan and Kelly Michelena and Keqian Li and Kiran Jagadeesh and Kun Huang and Kunal Chawla and Kyle Huang and Lailin Chen and Lakshya Garg and Lavender A and Leandro Silva and Lee Bell and Lei Zhang and Liangpeng Guo and Licheng Yu and Liron Moshkovich and Luca Wehrstedt and Madian Khabsa and Manav Avalani and Manish Bhatt and Martynas Mankus and Matan Hasson and Matthew Lennie and Matthias Reso and Maxim Groshev and Maxim Naumov and Maya Lathi and Meghan Keneally and Miao Liu and Michael L. Seltzer and Michal Valko and Michelle Restrepo and Mihir Patel and Mik Vyatskov and Mikayel Samvelyan and Mike Clark and Mike Macey and Mike Wang and Miquel Jubert Hermoso and Mo Metanat and Mohammad Rastegari and Munish Bansal and Nandhini Santhanam and Natascha Parks and Natasha White and Navyata Bawa and Nayan Singhal and Nick Egebo and Nicolas Usunier and Nikhil Mehta and Nikolay Pavlovich Laptev and Ning Dong and Norman Cheng and Oleg Chernoguz and Olivia Hart and Omkar Salpekar and Ozlem Kalinli and Parkin Kent and Parth Parekh and Paul Saab and Pavan Balaji and Pedro Rittner and Philip Bontrager and Pierre Roux and Piotr Dollar and Polina Zvyagina and Prashant Ratanchandani and Pritish Yuvraj and Qian Liang and Rachad Alao and Rachel Rodriguez and Rafi Ayub and Raghotham Murthy and Raghu Nayani and Rahul Mitra and Rangaprabhu Parthasarathy and Raymond Li and Rebekkah Hogan and Robin Battey and Rocky Wang and Russ Howes and Ruty Rinott and Sachin Mehta and Sachin Siby and Sai Jayesh Bondu and Samyak Datta and Sara Chugh and Sara Hunt and Sargun Dhillon and Sasha Sidorov and Satadru Pan and Saurabh Mahajan and Saurabh Verma and Seiji Yamamoto and Sharadh Ramaswamy and Shaun Lindsay and Shaun Lindsay and Sheng Feng and Shenghao Lin and Shengxin Cindy Zha and Shishir Patil and Shiva Shankar and Shuqiang Zhang and Shuqiang Zhang and Sinong Wang and Sneha Agarwal and Soji Sajuyigbe and Soumith Chintala and Stephanie Max and Stephen Chen and Steve Kehoe and Steve Satterfield and Sudarshan Govindaprasad and Sumit Gupta and Summer Deng and Sungmin Cho and Sunny Virk and Suraj Subramanian and Sy Choudhury and Sydney Goldman and Tal Remez and Tamar Glaser and Tamara Best and Thilo Koehler and Thomas Robinson and Tianhe Li and Tianjun Zhang and Tim Matthews and Timothy Chou and Tzook Shaked and Varun Vontimitta and Victoria Ajayi and Victoria Montanez and Vijai Mohan and Vinay Satish Kumar and Vishal Mangla and Vlad Ionescu and Vlad Poenaru and Vlad Tiberiu Mihailescu and Vladimir Ivanov and Wei Li and Wenchen Wang and Wenwen Jiang and Wes Bouaziz and Will Constable and Xiaocheng Tang and Xiaojian Wu and Xiaolan Wang and Xilun Wu and Xinbo Gao and Yaniv Kleinman and Yanjun Chen and Ye Hu and Ye Jia and Ye Qi and Yenda Li and Yilin Zhang and Ying Zhang and Yossi Adi and Youngjin Nam and Yu and Wang and Yu Zhao and Yuchen Hao and Yundi Qian and Yunlu Li and Yuzi He and Zach Rait and Zachary DeVito and Zef Rosnbrick and Zhaoduo Wen and Zhenyu Yang and Zhiwei Zhao and Zhiyu Ma},
      year={2024},
      eprint={2407.21783},
      archivePrefix={arXiv},
      primaryClass={cs.AI},
      url={https://arxiv.org/abs/2407.21783}, 
}

@misc{gpt4,
      title={GPT-4 Technical Report}, 
      author={OpenAI and Josh Achiam and Steven Adler and Sandhini Agarwal and Lama Ahmad and Ilge Akkaya and Florencia Leoni Aleman and Diogo Almeida and Janko Altenschmidt and Sam Altman and Shyamal Anadkat and Red Avila and Igor Babuschkin and Suchir Balaji and Valerie Balcom and Paul Baltescu and Haiming Bao and Mohammad Bavarian and Jeff Belgum and Irwan Bello and Jake Berdine and Gabriel Bernadett-Shapiro and Christopher Berner and Lenny Bogdonoff and Oleg Boiko and Madelaine Boyd and Anna-Luisa Brakman and Greg Brockman and Tim Brooks and Miles Brundage and Kevin Button and Trevor Cai and Rosie Campbell and Andrew Cann and Brittany Carey and Chelsea Carlson and Rory Carmichael and Brooke Chan and Che Chang and Fotis Chantzis and Derek Chen and Sully Chen and Ruby Chen and Jason Chen and Mark Chen and Ben Chess and Chester Cho and Casey Chu and Hyung Won Chung and Dave Cummings and Jeremiah Currier and Yunxing Dai and Cory Decareaux and Thomas Degry and Noah Deutsch and Damien Deville and Arka Dhar and David Dohan and Steve Dowling and Sheila Dunning and Adrien Ecoffet and Atty Eleti and Tyna Eloundou and David Farhi and Liam Fedus and Niko Felix and Simón Posada Fishman and Juston Forte and Isabella Fulford and Leo Gao and Elie Georges and Christian Gibson and Vik Goel and Tarun Gogineni and Gabriel Goh and Rapha Gontijo-Lopes and Jonathan Gordon and Morgan Grafstein and Scott Gray and Ryan Greene and Joshua Gross and Shixiang Shane Gu and Yufei Guo and Chris Hallacy and Jesse Han and Jeff Harris and Yuchen He and Mike Heaton and Johannes Heidecke and Chris Hesse and Alan Hickey and Wade Hickey and Peter Hoeschele and Brandon Houghton and Kenny Hsu and Shengli Hu and Xin Hu and Joost Huizinga and Shantanu Jain and Shawn Jain and Joanne Jang and Angela Jiang and Roger Jiang and Haozhun Jin and Denny Jin and Shino Jomoto and Billie Jonn and Heewoo Jun and Tomer Kaftan and Łukasz Kaiser and Ali Kamali and Ingmar Kanitscheider and Nitish Shirish Keskar and Tabarak Khan and Logan Kilpatrick and Jong Wook Kim and Christina Kim and Yongjik Kim and Jan Hendrik Kirchner and Jamie Kiros and Matt Knight and Daniel Kokotajlo and Łukasz Kondraciuk and Andrew Kondrich and Aris Konstantinidis and Kyle Kosic and Gretchen Krueger and Vishal Kuo and Michael Lampe and Ikai Lan and Teddy Lee and Jan Leike and Jade Leung and Daniel Levy and Chak Ming Li and Rachel Lim and Molly Lin and Stephanie Lin and Mateusz Litwin and Theresa Lopez and Ryan Lowe and Patricia Lue and Anna Makanju and Kim Malfacini and Sam Manning and Todor Markov and Yaniv Markovski and Bianca Martin and Katie Mayer and Andrew Mayne and Bob McGrew and Scott Mayer McKinney and Christine McLeavey and Paul McMillan and Jake McNeil and David Medina and Aalok Mehta and Jacob Menick and Luke Metz and Andrey Mishchenko and Pamela Mishkin and Vinnie Monaco and Evan Morikawa and Daniel Mossing and Tong Mu and Mira Murati and Oleg Murk and David Mély and Ashvin Nair and Reiichiro Nakano and Rajeev Nayak and Arvind Neelakantan and Richard Ngo and Hyeonwoo Noh and Long Ouyang and Cullen O'Keefe and Jakub Pachocki and Alex Paino and Joe Palermo and Ashley Pantuliano and Giambattista Parascandolo and Joel Parish and Emy Parparita and Alex Passos and Mikhail Pavlov and Andrew Peng and Adam Perelman and Filipe de Avila Belbute Peres and Michael Petrov and Henrique Ponde de Oliveira Pinto and Michael and Pokorny and Michelle Pokrass and Vitchyr H. Pong and Tolly Powell and Alethea Power and Boris Power and Elizabeth Proehl and Raul Puri and Alec Radford and Jack Rae and Aditya Ramesh and Cameron Raymond and Francis Real and Kendra Rimbach and Carl Ross and Bob Rotsted and Henri Roussez and Nick Ryder and Mario Saltarelli and Ted Sanders and Shibani Santurkar and Girish Sastry and Heather Schmidt and David Schnurr and John Schulman and Daniel Selsam and Kyla Sheppard and Toki Sherbakov and Jessica Shieh and Sarah Shoker and Pranav Shyam and Szymon Sidor and Eric Sigler and Maddie Simens and Jordan Sitkin and Katarina Slama and Ian Sohl and Benjamin Sokolowsky and Yang Song and Natalie Staudacher and Felipe Petroski Such and Natalie Summers and Ilya Sutskever and Jie Tang and Nikolas Tezak and Madeleine B. Thompson and Phil Tillet and Amin Tootoonchian and Elizabeth Tseng and Preston Tuggle and Nick Turley and Jerry Tworek and Juan Felipe Cerón Uribe and Andrea Vallone and Arun Vijayvergiya and Chelsea Voss and Carroll Wainwright and Justin Jay Wang and Alvin Wang and Ben Wang and Jonathan Ward and Jason Wei and CJ Weinmann and Akila Welihinda and Peter Welinder and Jiayi Weng and Lilian Weng and Matt Wiethoff and Dave Willner and Clemens Winter and Samuel Wolrich and Hannah Wong and Lauren Workman and Sherwin Wu and Jeff Wu and Michael Wu and Kai Xiao and Tao Xu and Sarah Yoo and Kevin Yu and Qiming Yuan and Wojciech Zaremba and Rowan Zellers and Chong Zhang and Marvin Zhang and Shengjia Zhao and Tianhao Zheng and Juntang Zhuang and William Zhuk and Barret Zoph},
      year={2024},
      eprint={2303.08774},
      archivePrefix={arXiv},
      primaryClass={cs.CL},
      url={https://arxiv.org/abs/2303.08774}, 
}

@misc{phi3,
      title={Phi-3 Technical Report: A Highly Capable Language Model Locally on Your Phone}, 
      author={Marah Abdin and Jyoti Aneja and Hany Awadalla and Ahmed Awadallah and Ammar Ahmad Awan and Nguyen Bach and Amit Bahree and Arash Bakhtiari and Jianmin Bao and Harkirat Behl and Alon Benhaim and Misha Bilenko and Johan Bjorck and Sébastien Bubeck and Martin Cai and Qin Cai and Vishrav Chaudhary and Dong Chen and Dongdong Chen and Weizhu Chen and Yen-Chun Chen and Yi-Ling Chen and Hao Cheng and Parul Chopra and Xiyang Dai and Matthew Dixon and Ronen Eldan and Victor Fragoso and Jianfeng Gao and Mei Gao and Min Gao and Amit Garg and Allie Del Giorno and Abhishek Goswami and Suriya Gunasekar and Emman Haider and Junheng Hao and Russell J. Hewett and Wenxiang Hu and Jamie Huynh and Dan Iter and Sam Ade Jacobs and Mojan Javaheripi and Xin Jin and Nikos Karampatziakis and Piero Kauffmann and Mahoud Khademi and Dongwoo Kim and Young Jin Kim and Lev Kurilenko and James R. Lee and Yin Tat Lee and Yuanzhi Li and Yunsheng Li and Chen Liang and Lars Liden and Xihui Lin and Zeqi Lin and Ce Liu and Liyuan Liu and Mengchen Liu and Weishung Liu and Xiaodong Liu and Chong Luo and Piyush Madan and Ali Mahmoudzadeh and David Majercak and Matt Mazzola and Caio César Teodoro Mendes and Arindam Mitra and Hardik Modi and Anh Nguyen and Brandon Norick and Barun Patra and Daniel Perez-Becker and Thomas Portet and Reid Pryzant and Heyang Qin and Marko Radmilac and Liliang Ren and Gustavo de Rosa and Corby Rosset and Sambudha Roy and Olatunji Ruwase and Olli Saarikivi and Amin Saied and Adil Salim and Michael Santacroce and Shital Shah and Ning Shang and Hiteshi Sharma and Yelong Shen and Swadheen Shukla and Xia Song and Masahiro Tanaka and Andrea Tupini and Praneetha Vaddamanu and Chunyu Wang and Guanhua Wang and Lijuan Wang and Shuohang Wang and Xin Wang and Yu Wang and Rachel Ward and Wen Wen and Philipp Witte and Haiping Wu and Xiaoxia Wu and Michael Wyatt and Bin Xiao and Can Xu and Jiahang Xu and Weijian Xu and Jilong Xue and Sonali Yadav and Fan Yang and Jianwei Yang and Yifan Yang and Ziyi Yang and Donghan Yu and Lu Yuan and Chenruidong Zhang and Cyril Zhang and Jianwen Zhang and Li Lyna Zhang and Yi Zhang and Yue Zhang and Yunan Zhang and Xiren Zhou},
      year={2024},
      eprint={2404.14219},
      archivePrefix={arXiv},
      primaryClass={cs.CL},
      url={https://arxiv.org/abs/2404.14219}, 
}

@misc{phi4,
      title={Phi-4 Technical Report}, 
      author={Marah Abdin and Jyoti Aneja and Harkirat Behl and Sébastien Bubeck and Ronen Eldan and Suriya Gunasekar and Michael Harrison and Russell J. Hewett and Mojan Javaheripi and Piero Kauffmann and James R. Lee and Yin Tat Lee and Yuanzhi Li and Weishung Liu and Caio C. T. Mendes and Anh Nguyen and Eric Price and Gustavo de Rosa and Olli Saarikivi and Adil Salim and Shital Shah and Xin Wang and Rachel Ward and Yue Wu and Dingli Yu and Cyril Zhang and Yi Zhang},
      year={2024},
      eprint={2412.08905},
      archivePrefix={arXiv},
      primaryClass={cs.CL},
      url={https://arxiv.org/abs/2412.08905}, 
}

@misc{phi4mini,
      title={Phi-4-Mini Technical Report: Compact yet Powerful Multimodal Language Models via Mixture-of-LoRAs}, 
      author={Microsoft and Abdelrahman Abouelenin and Atabak Ashfaq and Adam Atkinson and Hany Awadalla and Nguyen Bach and Jianmin Bao and Alon Benhaim and Martin Cai and Vishrav Chaudhary and Congcong Chen and Dong Chen and Dongdong Chen and Junkun Chen and Weizhu Chen and Yen-Chun Chen and Yi-ling Chen and Qi Dai and Xiyang Dai and Ruchao Fan and Mei Gao and Min Gao and Amit Garg and Abhishek Goswami and Junheng Hao and Amr Hendy and Yuxuan Hu and Xin Jin and Mahmoud Khademi and Dongwoo Kim and Young Jin Kim and Gina Lee and Jinyu Li and Yunsheng Li and Chen Liang and Xihui Lin and Zeqi Lin and Mengchen Liu and Yang Liu and Gilsinia Lopez and Chong Luo and Piyush Madan and Vadim Mazalov and Arindam Mitra and Ali Mousavi and Anh Nguyen and Jing Pan and Daniel Perez-Becker and Jacob Platin and Thomas Portet and Kai Qiu and Bo Ren and Liliang Ren and Sambuddha Roy and Ning Shang and Yelong Shen and Saksham Singhal and Subhojit Som and Xia Song and Tetyana Sych and Praneetha Vaddamanu and Shuohang Wang and Yiming Wang and Zhenghao Wang and Haibin Wu and Haoran Xu and Weijian Xu and Yifan Yang and Ziyi Yang and Donghan Yu and Ishmam Zabir and Jianwen Zhang and Li Lyna Zhang and Yunan Zhang and Xiren Zhou},
      year={2025},
      eprint={2503.01743},
      archivePrefix={arXiv},
      primaryClass={cs.CL},
      url={https://arxiv.org/abs/2503.01743}, 
}

@misc{qwen25,
      title={Qwen2.5 Technical Report}, 
      author={Qwen and An Yang and Baosong Yang and Beichen Zhang and Binyuan Hui and Bo Zheng and Bowen Yu and Chengyuan Li and Dayiheng Liu and Fei Huang and Haoran Wei and Huan Lin and Jian Yang and Jianhong Tu and Jianwei Zhang and Jianxin Yang and Jiaxi Yang and Jingren Zhou and Junyang Lin and Kai Dang and Keming Lu and Keqin Bao and Kexin Yang and Le Yu and Mei Li and Mingfeng Xue and Pei Zhang and Qin Zhu and Rui Men and Runji Lin and Tianhao Li and Tianyi Tang and Tingyu Xia and Xingzhang Ren and Xuancheng Ren and Yang Fan and Yang Su and Yichang Zhang and Yu Wan and Yuqiong Liu and Zeyu Cui and Zhenru Zhang and Zihan Qiu},
      year={2025},
      eprint={2412.15115},
      archivePrefix={arXiv},
      primaryClass={cs.CL},
      url={https://arxiv.org/abs/2412.15115}, 
}

@ARTICLE{edu_1_hmm,
    author={Sindhu, Irum and Muhammad Daudpota, Sher and Badar, Kamal and Bakhtyar, Maheen and Baber, Junaid and Nurunnabi, Mohammad},
    journal={IEEE Access}, 
    title={Aspect-Based Opinion Mining on Student’s Feedback for Faculty Teaching Performance Evaluation}, 
    year={2019},
    volume={7},
    number={},
    pages={108729-108741},
    doi={10.1109/ACCESS.2019.2928872}
}

@article{edu_5_autoscoring,
    author = {Ren, Ping and Yang, Liu and Luo, Fang},
    title = {Automatic scoring of student feedback for teaching evaluation based on aspect-level sentiment analysis},
    year = {2022},
    issue_date = {Jan 2023},
    publisher = {Kluwer Academic Publishers},
    address = {USA},
    volume = {28},
    number = {1},
    issn = {1360-2357},
    url = {https://doi.org/10.1007/s10639-022-11151-z},
    doi = {10.1007/s10639-022-11151-z},
    journal = {Education and Information Technologies},
    month = jul,
    pages = {797–814},
    numpages = {18}
}

@Article{implicitOE_7,
    AUTHOR = {Lei, Zhou and Zhang, Yawei and Chen, Shengbo},
    TITLE = {A Dual-Template Prompted Mutual Learning Generative Model for Implicit Aspect-Based Sentiment Analysis},
    JOURNAL = {Applied Sciences},
    VOLUME = {14},
    YEAR = {2024},
    NUMBER = {19},
    ARTICLE-NUMBER = {8719},
    URL = {https://www.mdpi.com/2076-3417/14/19/8719},
    ISSN = {2076-3417},
    DOI = {10.3390/app14198719}
}

@INPROCEEDINGS{ref_2025_01,
    author={Singhi, Vishal and Chauhan, Charulata and Soni, Piyush Kumar},
    booktitle={2024 IEEE 9th International Conference for Convergence in Technology (I2CT)}, 
    title={Exploring Progress in Aspect-based Sentiment Analysis: An In-depth Survey}, 
    year={2024},
    volume={},
    number={},
    pages={1-10},
    doi={10.1109/I2CT61223.2024.10543612}
}

@InProceedings{ref_2025_t5absa,
    author="Ayaz, Bisma
    and Gao, Xiaoying
    and Xue, Bing",
    editor="Wu, Xintao
    and Spiliopoulou, Myra
    and Wang, Can
    and Kumar, Vipin
    and Cao, Longbing
    and Zhou, Xiangmin
    and Pang, Guansong
    and Gama, Joao",
    title="Advancing Comprehensive Aspect-Based Sentiment Analysis with Generative Models",
    booktitle="Data Science: Foundations and Applications",
    year="2025",
    publisher="Springer Nature Singapore",
    address="Singapore",
    pages="93--105",
    isbn="978-981-96-8298-0"
}

@ARTICLE{ref_2025_decoupleabsa,
    author={Wang, Zengzhi and Xia, Rui and Yu, Jianfei},
    journal={IEEE Transactions on Knowledge and Data Engineering}, 
    title={Unified ABSA via Annotation-Decoupled Multi-Task Instruction Tuning}, 
    year={2024},
    volume={36},
    number={11},
    pages={7242-7254},
    doi={10.1109/TKDE.2024.3392836}
}

@inproceedings{hu2022lora,
    title={Lo{RA}: Low-Rank Adaptation of Large Language Models},
    author={Edward J Hu and yelong shen and Phillip Wallis and Zeyuan Allen-Zhu and Yuanzhi Li and Shean Wang and Lu Wang and Weizhu Chen},
    booktitle={International Conference on Learning Representations},
    year={2022},
    url={https://openreview.net/forum?id=nZeVKeeFYf9}
}

@InProceedings{ref_2025_gpt_on_sa_1,
    author="Obinwanne, Tobechi
    and Brandtner, Patrick",
    editor="Nagar, Atulya K.
    and Jat, Dharm Singh
    and Mishra, Durgesh
    and Joshi, Amit",
    title="Enhancing Sentiment Analysis with GPT---A Comparison of Large Language Models and Traditional Machine Learning Techniques",
    booktitle="Intelligent Sustainable Systems",
    year="2024",
    publisher="Springer Nature Singapore",
    address="Singapore",
    pages="187--197",
    isbn="978-981-99-7569-3"
}

@misc{ref_2025_glms_on_sa_2,
    title={ChatGPT vs Gemini vs LLaMA on Multilingual Sentiment Analysis}, 
    author={Alessio Buscemi and Daniele Proverbio},
    year={2024},
    eprint={2402.01715},
    archivePrefix={arXiv},
    primaryClass={cs.CL},
    url={https://arxiv.org/abs/2402.01715}, 
}

@misc{mmlu,
    title={Measuring Massive Multitask Language Understanding}, 
    author={Dan Hendrycks and Collin Burns and Steven Basart and Andy Zou and Mantas Mazeika and Dawn Song and Jacob Steinhardt},
    year={2021},
    eprint={2009.03300},
    archivePrefix={arXiv},
    primaryClass={cs.CY},
    url={https://arxiv.org/abs/2009.03300}, 
}

@misc{modelsoup,
    title={Model soups: averaging weights of multiple fine-tuned models improves accuracy without increasing inference time}, 
    author={Mitchell Wortsman and Gabriel Ilharco and Samir Yitzhak Gadre and Rebecca Roelofs and Raphael Gontijo-Lopes and Ari S. Morcos and Hongseok Namkoong and Ali Farhadi and Yair Carmon and Simon Kornblith and Ludwig Schmidt},
    year={2022},
    eprint={2203.05482},
    archivePrefix={arXiv},
    primaryClass={cs.LG},
    url={https://arxiv.org/abs/2203.05482}, 
}

@inproceedings{mergekit,
    title = "Arcee{'}s {M}erge{K}it: A Toolkit for Merging Large Language Models",
    author = "Goddard, Charles  and
      Siriwardhana, Shamane  and
      Ehghaghi, Malikeh  and
      Meyers, Luke  and
      Karpukhin, Vladimir  and
      Benedict, Brian  and
      McQuade, Mark  and
      Solawetz, Jacob",
    editor = "Dernoncourt, Franck  and
      Preo{\c{t}}iuc-Pietro, Daniel  and
      Shimorina, Anastasia",
    booktitle = "Proceedings of the 2024 Conference on Empirical Methods in Natural Language Processing: Industry Track",
    month = nov,
    year = "2024",
    address = "Miami, Florida, US",
    publisher = "Association for Computational Linguistics",
    url = "https://aclanthology.org/2024.emnlp-industry.36",
    doi = "10.18653/v1/2024.emnlp-industry.36",
    pages = "477--485"
}

@article{llm_sft,
    author    = {Wei Lu and Rachel K. Luu and Markus J. Buehler},
    title     = {Fine-tuning large language models for domain adaptation: exploration of training strategies, scaling, model merging and synergistic capabilities},
    journal   = {npj Computational Materials},
    year      = {2025},
    volume    = {11},
    number    = {1},
    pages     = {84},
    doi       = {10.1038/s41524-025-01564-y},
    url       = {https://doi.org/10.1038/s41524-025-01564-y},
    issn      = {2057-3960}
}

@misc{rslora,
    title={A Rank Stabilization Scaling Factor for Fine-Tuning with LoRA}, 
    author={Damjan Kalajdzievski},
    year={2023},
    eprint={2312.03732},
    archivePrefix={arXiv},
    primaryClass={cs.CL},
    url={https://arxiv.org/abs/2312.03732}, 
}

@book{liu2012sentimentanalysisbook,
  title     = {Sentiment Analysis and Opinion Mining},
  author    = {Liu, Bing},
  year      = {2012},
  publisher = {Springer Cham},
  series    = {Synthesis Lectures on Human Language Technologies},
  isbn      = {978-3-031-01017-0},  
  isbn10    = {303101017X},
  isbn13    = {9783031010170},
  doi       = {10.1007/978-3-031-02145-9},
  pages     = {XIV + 167},
  url       = {https://doi.org/10.1007/978-3-031-02145-9}
}

@inproceedings{rouge_bleu,
    title = "Automatic Evaluation of Machine Translation Quality Using Longest Common Subsequence and Skip-Bigram Statistics",
    author = "Lin, Chin-Yew  and
      Och, Franz Josef",
    booktitle = "Proceedings of the 42nd Annual Meeting of the Association for Computational Linguistics ({ACL}-04)",
    month = jul,
    year = "2004",
    address = "Barcelona, Spain",
    url = "https://aclanthology.org/P04-1077/",
    doi = "10.3115/1218955.1219032",
    pages = "605--612"
}

@inproceedings{rouge,
    title = "{ROUGE}: A Package for Automatic Evaluation of Summaries",
    author = "Lin, Chin-Yew",
    booktitle = "Text Summarization Branches Out",
    month = jul,
    year = "2004",
    address = "Barcelona, Spain",
    publisher = "Association for Computational Linguistics",
    url = "https://aclanthology.org/W04-1013/",
    pages = "74--81"
}

@ARTICLE{linearsum,
  author={Crouse, David F.},
  journal={IEEE Transactions on Aerospace and Electronic Systems}, 
  title={On implementing 2D rectangular assignment algorithms}, 
  year={2016},
  volume={52},
  number={4},
  pages={1679-1696},
  doi={10.1109/TAES.2016.140952}
}

@inproceedings{bai_llm_2024,
    title = "Is Compound Aspect-Based Sentiment Analysis Addressed by {LLM}s?",
    author = "Bai, Yinhao  and
      Han, Zhixin  and
      Zhao, Yuhua  and
      Gao, Hang  and
      Zhang, Zhuowei  and
      Wang, Xunzhi  and
      Hu, Mengting",
    editor = "Al-Onaizan, Yaser  and
      Bansal, Mohit  and
      Chen, Yun-Nung",
    booktitle = "Findings of the Association for Computational Linguistics: EMNLP 2024",
    month = nov,
    year = "2024",
    address = "Miami, Florida, USA",
    publisher = "Association for Computational Linguistics",
    url = "https://aclanthology.org/2024.findings-emnlp.460/",
    doi = "10.18653/v1/2024.findings-emnlp.460",
    pages = "7836--7861"
}

@misc{zhou_llm_2024,
      title={A Comprehensive Evaluation of Large Language Models on Aspect-Based Sentiment Analysis}, 
      author={Changzhi Zhou and Dandan Song and Yuhang Tian and Zhijing Wu and Hao Wang and Xinyu Zhang and Jun Yang and Ziyi Yang and Shuhao Zhang},
      year={2024},
      eprint={2412.02279},
      archivePrefix={arXiv},
      primaryClass={cs.CL},
      url={https://arxiv.org/abs/2412.02279}, 
}

@inproceedings{llm_czechabsa_2025,
    author = {\v{S}m\'{\i}d, Jakub and P\v{r}ib\'{a}\v{n}, Pavel and Kr\'{a}l, Pavel},
    title = {Large Language Models for Czech Aspect-Based Sentiment Analysis},
    year = {2025},
    isbn = {978-3-032-02550-0},
    publisher = {Springer-Verlag},
    address = {Berlin, Heidelberg},
    url = {https://doi.org/10.1007/978-3-032-02551-7_3},
    doi = {10.1007/978-3-032-02551-7_3},
    booktitle = {Text, Speech, and Dialogue: 28th International Conference, TSD 2025, Erlangen, Germany, August 25–28, 2025, Proceedings, Part II},
    pages = {15–26},
    numpages = {12},
    location = {Erlangen, Germany}
}

@inproceedings{ding_llm_2024,
    title = "Boosting Large Language Models with Continual Learning for Aspect-based Sentiment Analysis",
    author = "Ding, Xuanwen  and
      Zhou, Jie  and
      Dou, Liang  and
      Chen, Qin  and
      Wu, Yuanbin  and
      Chen, Arlene  and
      He, Liang",
    editor = "Al-Onaizan, Yaser  and
      Bansal, Mohit  and
      Chen, Yun-Nung",
    booktitle = "Findings of the Association for Computational Linguistics: EMNLP 2024",
    month = nov,
    year = "2024",
    address = "Miami, Florida, USA",
    publisher = "Association for Computational Linguistics",
    url = "https://aclanthology.org/2024.findings-emnlp.252/",
    doi = "10.18653/v1/2024.findings-emnlp.252",
    pages = "4367--4377"
}

@INPROCEEDINGS{lee_llm_2024,
  author={Lee, Chaelyn and Lee, Hanyong and Kim, Kyumin and Kim, Sojeong and Lee, Jaesung},
  booktitle={2024 IEEE International Conference on Consumer Electronics (ICCE)}, 
  title={An Efficient Fine-tuning of Generative Language Model for Aspect-Based Sentiment Analysis}, 
  year={2024},
  volume={},
  number={},
  pages={1-4},
  doi={10.1109/ICCE59016.2024.10444216}
}

@ARTICLE{scipy,
  author  = {Virtanen, Pauli and Gommers, Ralf and Oliphant, Travis E. and
            Haberland, Matt and Reddy, Tyler and Cournapeau, David and
            Burovski, Evgeni and Peterson, Pearu and Weckesser, Warren and
            Bright, Jonathan and {van der Walt}, St{\'e}fan J. and
            Brett, Matthew and Wilson, Joshua and Millman, K. Jarrod and
            Mayorov, Nikolay and Nelson, Andrew R. J. and Jones, Eric and
            Kern, Robert and Larson, Eric and Carey, C J and
            Polat, {\.I}lhan and Feng, Yu and Moore, Eric W. and
            {VanderPlas}, Jake and Laxalde, Denis and Perktold, Josef and
            Cimrman, Robert and Henriksen, Ian and Quintero, E. A. and
            Harris, Charles R. and Archibald, Anne M. and
            Ribeiro, Ant{\^o}nio H. and Pedregosa, Fabian and
            {van Mulbregt}, Paul and {SciPy 1.0 Contributors}},
  title   = {{{SciPy} 1.0: Fundamental Algorithms for Scientific
            Computing in Python}},
  journal = {Nature Methods},
  year    = {2020},
  volume  = {17},
  pages   = {261--272},
  adsurl  = {https://rdcu.be/b08Wh},
  doi     = {10.1038/s41592-019-0686-2},
}

@inproceedings{zhang2021_asqp,
    title = "Aspect Sentiment Quad Prediction as Paraphrase Generation",
    author = "Zhang, Wenxuan  and
      Deng, Yang  and
      Li, Xin  and
      Yuan, Yifei  and
      Bing, Lidong  and
      Lam, Wai",
    booktitle = "Proceedings of the 2021 Conference on Empirical Methods in Natural Language Processing",
    month = nov,
    year = "2021",
    publisher = "Association for Computational Linguistics",
    url = "https://aclanthology.org/2021.emnlp-main.726",
    pages = "9209--9219",
}

@inproceedings{cai2021_acos,
  title={Aspect-Category-Opinion-Sentiment Quadruple Extraction with Implicit Aspects and Opinions},
  author={Cai, Hongjie and Xia, Rui and Yu, Jianfei},
  booktitle={Proceedings of the 59th Annual Meeting of the Association for Computational Linguistics and the 11th International Joint Conference on Natural Language Processing (Volume 1: Long Papers)},
  pages={340--350},
  year={2021}
}

@article{compound_task_survey_2025,
    title = {Methodologies and their comparison in complex compound aspect-based sentiment analysis: A survey},
    journal = {AI Open},
    volume = {6},
    pages = {53-69},
    year = {2025},
    issn = {2666-6510},
    doi = {https://doi.org/10.1016/j.aiopen.2025.02.002},
    url = {https://www.sciencedirect.com/science/article/pii/S2666651025000051},
    author = {Faiz Ghifari Haznitrama and Ho-Jin Choi and Chin-Wan Chung},
}

\end{document}